\documentclass[11pt,a4paper]{report}

\pdfoutput=1

\usepackage{amsmath}
\usepackage{amsfonts}

\usepackage{paralist}

\usepackage[pdftex]{graphicx}

\usepackage{multirow}

\usepackage{rotating}

\usepackage[left=40mm, right=20mm, top=20mm, bottom=25mm]{geometry}

\usepackage{subfigure}

\newcommand{\etal}{{et al.\ }}
\renewcommand{\figurename}{{Fig.\ }}
\renewcommand{\tablename}{{Table\ }}

\usepackage{setspace}
\usepackage{url}

\usepackage{wrapfig}

\begin{document}

\begin{titlepage}
\begin{center}
\vspace*{1.0in}
{\LARGE \textbf{Face Shape and Reflectance Acquisition using\\a Multispectral Light Stage}}
\par
\vspace{0.30in}
{\large Abhishek Dutta}
\par
\vfill
Submitted for the degree of Master of Science (M.Sc.)
\par
\vspace{0.25in}
Department of Computer Science
\par
\vspace{0.25in}
\textbf{THE UNIVERSITY of York}
\par
\vspace{0.25in}
October 2010
\vspace{1.25in}
\end{center}
\end{titlepage}

\begin{abstract}
In this thesis, we discuss the design and calibration (geometric and radiometric) of a novel shape and reflectance acquisition device called the ``Multispectral Light Stage''. This device can capture highly detailed facial geometry (down to the level of skin pores detail) and Multispectral reflectance map which can be used to estimate biophysical skin parameters such as the distribution of pigmentation and blood beneath the surface of the skin.

We extend the analysis of the original spherical gradient photometric stereo method to study the effects of deformed diffuse lobes on the quality of recovered surface normals. Based on our modified radiance equations, we develop a minimal image set method to recover high quality photometric normals using only four, instead of six, spherical gradient images. Using the same radiance equations, we explore a Quadratic Programming (QP) based algorithm for correction of surface normals obtained using spherical gradient photometric stereo.

Based on the proposed minimal image sets method, we present a performance capture sequence that significantly reduces the data capture requirement and post-processing computational cost of existing photometric stereo based performance geometry capture methods.

Furthermore, we explore the use of images captured in our Light Stage to generate stimuli images for a psychology experiment exploring the neural representation of 3D shape and texture of a human face.
\end{abstract}

\pagenumbering{roman}
\tableofcontents
\listoffigures
\listoftables

\chapter*{Acknowledgments}
First, I would like to thank Dr. William Smith, my supervisor, for always keeping his office doors open. This not only meant that he was accessible all the time but also ensured that I could discuss my problems as I encountered them while exploring my research terrain. Due to his friendly and cheerful nature, I never hesitated to discuss any of my ideas. He deserves to be ``Supervisor of the Year''. Thanks also to him for patiently standing in the Light Stage during long data capture sessions.

Thanks to my sister, Dr. Smita Dutta, for sending delicious food to me every week. My brother-in-law, Dr. Abhishek Kabra, tirelessly shuttled between York and Driffield to make sure that I was never left alone during weekends and holidays. I never felt like staying thousands of miles away from home.

Thanks also to Manoj mama(\textit{maithili} word for maternal uncle) for giving me a laptop at the age of 14 and introducing me to Computers. To be honest, I did not even deserve a pocket calculator from him as I had destroyed all his electronics kits and tools while exploring the wonderful world of hobby electronics. I am also truly grateful to Pawan Kumar Pandey (Computer Science teacher at DPS, Dharan) for being the architect of my C/C++ programming skills.

Finally, I would also like to acknowledge the EURECA project (a project funded by the Erasmus Mundus External Cooperation Window (EMECW) of the European Commission) for funding my MSc research.

\chapter*{}
\thispagestyle{empty} 
\addtocounter{page}{-1} 
\vspace{1.5in} 
\begin{center} 
  \textit{
  On your shoulders I stand, \\
  For a vantage point, very few have. \\
  Narrow minded thoughts fear to tread in my conscience, \\
  Thanks to your teachings that are always on guard. \\
  Wisdom made sense after you explained Russel's ``Knowledge and Wisdom'', \\
  And so did Tagore's ``Where the mind is without fear (Gitanjali)''. \\
  \vspace{0.5in}
  \hspace{2in}- \textbf{to Baba and my parents}
  }
\end{center}

\chapter*{Declaration}
I declare that all the work in this thesis is solely my own, except where attributed and cited to another author. Some of the material in this thesis has been previously published by the author: please refer to Section~\ref{ch:introduction:sec:contribution} for details.

\chapter{Introduction}
\pagenumbering{arabic}
\label{ch:introduction}
The 3 dimensional shape and skin texture (i.e. 2 dimensional skin reflectance) of a human face determines its visual appearance. Face shape and reflectance acquisition devices aim to capture these two information from real human faces, either separately or in a combined form. Extensive research has been pursued in the past three decades to develop devices that are capable of acquiring shape and reflectance information accurately and conveniently. Photorealistic renderings of human faces can be created using the highly detailed shape and texture information acquired from these capture devices. As a result, digital actors, who have natural looking faces, can be extensively used in movies and animation. Additionally, access to accurate shape and reflectance information is the key to developing a new generation of face recognition algorithms that can maintain accuracy even under arbitrary pose and illumination variation.

In this thesis, we discuss the design and calibration (geometric and radiometric) of a novel shape and reflectance acquisition device that is able to capture highly detailed facial geometry (down to the level of skin pores detail) and a Multispectral reflectance map. The Multispectral reflectance map can be used to estimate biophysical skin parameters such as the distribution of pigmentation and blood beneath the surface of the skin. This device, called the \textit{Multispectral Light Stage}, is an extension of the Light Stages developed at UC Berkeley and University of South California (USC ICT) \cite{lightstages2010}. We use a beam splitter based capture device to simultaneously acquire parallel and cross polarised images. This ensures that both the acquired images are in perfect registration and hence results in very accurate diffuse and specular reflectance separation. Previous Light Stages \cite{lightstages2010} relied on a servo motor to flip the plane of polarisation of a polarising filter.
As a result, the capture time increases. This increase in capture time does not affect the shape and reflectance recovery of a static object. However, for a non-static objects like a human face, it is extremely difficult to remain in the same position and maintain a facial expression for the capture duration. Hence, increased capture time compromises the quality of recovered surface geometry and reflectance information because the captured images are no longer perfectly aligned.

Capturing the facial geometry of human actors during a dynamic performance is the first step towards creating digital actors that can produce realistic and natural facial expressions. Such digital actors are in high demand for movies and animation. The human face is capable of producing a large number of facial expressions with small non-rigid motion of facial muscles. Hence, to reproduce such expressions in a digital actor, it is essential that, in addition to capturing overall facial expression, the motion of fine scale skin features (like wrinkles, pores, scars, etc) are also captured. Such fine details are the key ingredient to reproducing natural facial expressions.

Photometric stereo based methods can capture all the fine scale details of a dynamic performance. However, they require expensive high speed photography equipment and are data intensive. In this thesis, we have proposed a novel real time performance capture sequence by exploiting the fact that high quality photometric normals can be recovered using just $4$ images. This new capture sequence not only reduces the data capture requirements for realtime performance geometry capture, but reduces the need for expensive high speed photography equipment for capture of highly detailed performance geometry.

Understanding the way the human brain represents and processes visual information is the key to creating machine vision algorithms that can match the capabilities of the human visual cortex. Unfortunately, non-invasive reverse engineering methods are the only practical tools available for such study. One of the popular choices for such reverse engineering approach is to study the brain activity of a human observer, usually in a controlled environment, when they are exposed to various types of visual stimuli. Brain activity during such experiments is mostly monitored using functional Magnetic Resonance Imaging (fMRI) and Electroencephalography (EEG). In addition to the capability of monitoring devices, the effectiveness of these experiments also depend on the ability to control various aspects of the visual stimulus. For example: the neural representation of 3D shape and 2D skin reflectance function (i.e. texture) can be effectively studied if we can create stimuli image that only contain the 3D shape or the 2D skin reflectance information. The photographs that we capture using a standard camera contain a mix of these two information sources. Availability of such stimulus data is the key to unravelling the psychology and neuropsychology of face perception.

In this thesis, we have also explored the use of our Light Stage for generating such a stimulus dataset. This dataset was used by Jones \etal \cite{jones2010how} in their study of the neural representation of 3D shape and 2D skin reflectance of human faces. To our knowledge, this is the first research to use Light Stage data (the high quality photometric normals and image of a face under spherical illumination) for generating a psychological stimulus dataset. We envisage that this will lead to further exploration of the use of a Light Stage in psychology experiments.

\section{Contributions}
\label{ch:introduction:sec:contribution}
The major contributions of this thesis are:
\begin{description}
\item[Design and Calibration of Multispectral Light Stage]
We have described the design and calibration (geometric and radiometric) of an extended version of the original Light Stage. Our Light Stage design consists of a ``beam splitter'' based setup that allows simultaneous capture of parallel and cross polarised images. Furthermore, our capture device uses a filter wheel containing narrowband optical filters, to separately record reflectance in different bands of the visible spectrum.

\item[Minimal Image Sets for Robust Spherical Gradient Photometric Stereo]
We extend the analysis of original spherical gradient photometric stereo developed by Ma \etal \cite{ma2007rapid} to consider the effect of deformed diffuse lobes on the quality of recovered surface normals. Based on our modified radiance equations, we explore a Quadratic Programming (QP) approach to correction of surface normals recovered using existing spherical gradient photometric stereo methods. Using the same set of equations, we propose that a minimal set of $4$ images can recover surface normals of the quality provided by the existing $6$ image method. This minimal image set method has also been described in the following publication:
\begin{itemize}
  \item Abhishek Dutta and William A. P. Smith. Minimal image sets for robust spherical gradient photometric stereo. In ACM SIGGRAPH ASIA 2010 Sketches, SA ’10, pages 22:1–22:2, New York, NY, USA, 2010. ACM.
\end{itemize}
It is important to realise that our analysis is based on the following simplifying assumption described in section \ref{ch:mls_data_processing:sec:quad_prog_for_norm_improvement:subsec:modified_radiance_eq_grad_illum} and \ref{ch:mls_data_processing:sec:quad_prog_for_norm_improvement:subsec:complement_constraints}: overall deformation in the diffuse reflectance lobe for gradient and complement gradient illumination environment can be quantified using a single scalar parameter $\delta_{\{x,y,z\}}$ and $\delta_{\{x,y,z\}} + \delta_{\{\bar{x},\bar{y},\bar{z}\}}$ respectively.

\item[Novel Capture Sequence for Real Time Performance Capture]
Based on our ``Minimal Image Sets'' analysis and building on the work of Wilson \etal \cite{wilson2010temporal}, we propose a new image capture sequence for facial performance geometry capture during dynamic performance. In addition to reducing the data capture requirement of Wilson \etal performance capture framework, the proposed capture sequence also reduces its computational cost by requiring alignment of only one, instead of three, pair of gradient and complement gradient images.

\item[Stimulus Image Dataset for Psychology Experiment]
We explore, for the first time, the use of images captured in a Light Stage to generate stimulus images for a psychology experiment. For a given face, we generate three stimulus images: the first contains only the 3D shape information, the second contains only 2D skin reflectance (texture) information and the third contains both shape and texture information. This image dataset has been used for studying the neural representation of 3D shape and texture of a human face.

\end{description}

\chapter{Related Work}
\label{ch:related_work}
In this chapter, we will discuss the previous work done in shape and reflectance acquisition and the method used to align captured data affected by subject motion during the capture process. In the later part of this thesis, we discuss how the data captured in our Light Stage can be used to produce better stimulus images for psychology experiments. Moreover, we also discuss a new performance capture strategy to reduce the data capture requirement of existing methods used to capture facial geometry during dynamic performance. Therefore, in this section, we also review the previous work done in these two areas.

\section{Shape Acquisition}
Two of the most popular methods for 3D shape acquisition are: Depth from triangulation and Photometric Stereo. As a result, two types of representation exist for conveying the 3D shape information: 3D meshes (vertices and their connectivity) and normal map (surface normal at each image point).

In the depth from triangulation method, a surface point is viewed from two (or more) viewpoints using a calibrated camera and the corresponding image points are recorded. In the ideal case (i.e. with no imaging noise), the rays through these two (or more) corresponding images points will intersect at a point in 3D space. This 3D location represents the original surface point represented by the corresponding image points observed using calibrated camera. There exist several methods to determine the corresponding image points in multiple views of a 3D object recorded using a calibrated camera. Nehab \cite{nehab2007advances} developed the ``spacetime stereo'' framework to classify all the existing depth from triangulation methods. This classification was based on the domain (spatial or temporal) in which the corresponding image points are located. The class of methods which determine corresponding image points by the analysis of similar pixels in the image plane are classified as ``spatial domain'' methods. On the other hand, methods which determine image point correspondance by analysis of pixel intensity variation over time are classified as ``temporal domain'' methods. This classfication not only provides a unified view of all the existing depth from triangulation methods, but also provided valuable insight for development of two new methods in \cite{nehab2007advances} that exploited both ``spatial'' and ``temporal'' domain constraints of corresponding image points.

Woodham \cite{woodham1980photometric} proposed the photometric stereo method to determine the surface geometry of each image point using diffuse images captured by varying the direction of incident illumination while keeping the view direction constant. The basis for this technique is the observation that each pixel intensity of a Lambertian surface image illuminated by a point source results in a linear photometric equation. If the direction of point source is known, then this system of linear equations can be inverted to recover unknown diffuse albedo and surface orientation using at least $3$ images. The unit surface normal constraint allows separation of these two quantities from a system of 3 linear photometric equations. This early version of photometric stereo method developed in \cite{woodham1980photometric} did not consider the effect of specular highlight, shadow and inter-reflection in the captured images.

It is convenient to invert a linear system resulting from photometric equations of a Lambertian reflection. However, no such linear system exists for non-Lambertian surface reflection. Hence, several previous research in photometric stereo has focused on developing methods to detect image points affected by specular highlight and shadow. Colenman and Jain \cite{coleman1982obtaining} proposed the use of $4$ point light sources, instead of just $3$, to detect and exclude pixels affected by specular highlights and shadow. Three surface normals corresponding to a single surface patch were available from these $4$ images captured using point light sources. They predicted that a large amount of deviation in both direction and magnitude of these three surface normals would occur for a pixel affected by specular highlight. This allowed them to tag and remove the specular source. Their method was based on the assumption that only one of the four light sources can cause specular highlight at any given image point.

Following the $4$ source strategy, Barsky and Petrou \cite{barsky2003four} used four spectrally distinct light sources to exploit the linearly independent photometric equations resulting from different color channels of a color image. They used spectral or directional cues to detect shadows and highlights in the input images. However, the choice of threshold parameter proved pivotal to the detection accuracy. \cite{barsky2003four} observed that with increase in imaging noise, a single threshold value cannot detect all the specular highlights and shadows present in the captured images.

A completely different approach to photometric stereo was pursued by Basri \etal \cite{basri2007photometric}. They felt the need for photometric stereo technique to work under general illumination condition. They argued that it was not always possible to control illumination for large outdoor structures or have knowledge of light source direction and strength for photographs taken under everyday lighting condition. The fact that any image of a convex Lambertian object under complex illumination can be approximated as a linear combination of $4$ (first order) or $9$ (second order) harmonic images\footnote{harmonic images represent the image of an object in low frequency lighting condition} forms the basis of their proposed photometric stereo algorithm for general illumination. Harmonic images can be expressed in terms of surface albedo and normal components and hence such decomposition allowed them to estimate these two quantities. They propose 9D (which requires at least 9 images) method and 4D (which requires at least 4 images) method of photometric stereo under general lighting condition. The 9D method produces slightly better results at the expense of higher computational cost of decomposing images captured under general lighting condition into $9$ harmonic images. The authors illustrate the quality of surface geometry reconstruction by using more images $(64, 32, 11, 10)$ than the required minimum. For example: the fine scale surface details of a volleyball was recovered by using $64$ images of the ball lit by point light sources (strength and direction unknown). Hence, at the expense of large computational cost and comparitively larger number of images, they were able to estimate good quality surface geometry under general illumination conditions. It is important to realise that this method is not applicable to images containing specular highlight or shadows.

All the previous photometric stereo methods (\cite{woodham1980photometric}, \cite{coleman1982obtaining}, \cite{barsky2003four}, \cite{basri2007photometric}) treated specular highlight in an image as a undersirable effect which restricted the application domain of photometric stereo. Extensive research has been done to develop methods for tagging and removal of specular highlights. However, Ma \etal \cite{ma2007rapid} used specularity to their advantage and acquired specular normal maps containing fine surface details of a human face never recovered by previous methods. They have shown how high resolution shape and reflectance information can be measured using an extended version of photometric stereo called the spherical gradient photometric stereo. An object is placed at the centre of a ``light stage'' which uses polarised spherical gradient illumination arranged such that the plane of polarisation 
after reflection, from the object towards the camera, are all the same. This allows the setup to separate diffuse and specular reflectance components by acquiring parallel and cross polarised images. The key observation underpinning this approach is that the centroid of the diffuse or specular reflectance lobe coincides with the surface normal or reflection vector respectively. The insight of Ma \etal was to show how to estimate the reflectance centroids using spherical gradient illumination conditions. When integrated with an illumination gradient in X, Y, or Z direction, the corresponding components of the reflectance centroid, and hence surface normal, can be recovered. This extended version of photometric stereo was capable of recovering fine scale surface details that was unmatched by the existing photometric stereo methods in terms of quality and level of detail.

The quality of surface geometry recovered using Ma \etal \cite{ma2007rapid} method is affected by the extent to which the following assumptions are satisfied: 
\begin{inparaenum}[\itshape a\upshape)]
\item no shadowing of light sources , i.e. object is convex;
\item no inter-reflections;
\item Fresnel term\footnote{The proportion of light transmitted into the surface and subsequently diffusely reflected varies with incidence angle according to Fresnel's equations. The same effect will occur when the diffused light exits the surface again.}; and
\item light sources closely approximate a continuous illumination environment.
\end{inparaenum}
The last assumption can be addressed by maximising the number of light sources in the light stage: Ma \etal used 156 LEDs attached to vertices and edges of a twice subdivided icosahedron. This method also ignores light source attenuation effects, which is equivalent to assuming all the points on the object lie exactly at the centre of the light stage.

Wilson \etal \cite{wilson2010temporal} proposed using gradient and complement gradient images to reduce the effect of shadowing. Instead of using the ``ratio'' method of Ma \etal to compute the surface normal components, they used the difference of gradient and complement gradient images to estimate more accurate surface geometry. They argued, ``since the pixels that are dark under one gradient illumination condition are most likely well exposed under the complement gradient illumination condition'' \cite{wilson2010temporal}. Recently, Dutta and Smith \cite{dutta2010minimal} have proved the validity of this claim by showing that the difference image method of Wilson \etal result in cancellation of symmetric deformation in diffuse lobes. This deformation cancellation property is not present in the method of Ma \etal because it involves estimation of surface normal components from the ratio images which preserves the term quantifying deformation in diffuse reflectance lobe. Dutta and Smith \cite{dutta2010minimal} have also shown that a minimal four image set can achieve the ``improved robustness'' quality of \cite{wilson2010temporal} while preserving the ``reduced data capture'' benifit of \cite{ma2008framework}. They used a ``light stage'' with only $41$ LEDs (attached only to vertices of a twice subdivided icosahedron) to study the degradation in the quality of recovered surface geometry with increase in ``light discretization'' i.e. coarse approximation of continuous spherical gradient illumination. Their minimal image set method was able to recover high quality normal map using a spherical illumination created with only $41$ LEDs.

\section{Reflectance Acquisition}
Reflectance models are an attempt to mathematically capture the interaction of light with a given material or class of materials. In Computer Graphics and Computer Vision, the reflectance properties of human skin have been investigated extensively in the past two decades. Reflectance models allow the creation of photo-realistic renderings of human faces in arbitrary pose and under complex illumination. It helps with development of natural looking cosmetics because reflectance models provide insight into the way light interacts with human skin \cite{tsumura2003image}. Photo therapy (or Laser based treatment) of skin disease requires good understanding of the interaction between light and human skin. Skin reflectance models help improve the precision of such treatment methods by allowing designers to simulate the effect of light based skin treatment methods \cite{igarashi2007appearance}.

Reflectance models mostly rely on measured reflectance data for estimation of their model parameters. The practicability of reflectance models depend on the ease with which reflectance properties of real world objects can be acquired. Most reflectance models discuss the related capture device that can acquire reflectance measurements required for estimation of the model parameters. Often, new capture devices trigger the development of reflectance models that can make full use of the available reflectance data. Hence, in addition to reviewing existing skin reflectance models, we will also discuss about the corresponding reflectance measurement device. In this section, we will discuss previous work done in the reflectance models related to human skin. These models can easily be modified to simulate light interaction in other types of materials like milk, marble, etc.

Marschner \etal \cite{marschner1999image} developed a reflectance capture device that, for the first time, measured the \textit{in vivo} surface refletance of human skin. A set of three machine readable targets were used for geometric calibration. This allowed automatic estimation of the relative position of the light source, sample material and the camera. For radiometric calibration, they used a calibrated reference source to determine the spectral characteristic of the camera. The radiometric calibration allowed them to relate the recorded pixel values with radiance reflected from the sample under study. A section of forehead was imaged under several incident illumination directions (capture time $\sim 30$min ). This region of the face was selected because it was relatively smooth, convex and involved least amount of deformation during long capture session. Using the machine readable targets, the geometric arrangement of sample, camera, light source and reference white target was automatically determined for each captured image. All these information was supplied to a ``derenderer'' which computed the BRDF value at each pixel positon by dividing the measured pixel radiance with the source irradiance. The scene geometry required by the ``derenderer'' was captured using a 3D range scanner. The authors produced renderings of human head using the measured BRDF of the skin sample. This rendering had a hard look and lacked the features of actual human skin because the proposed skin reflectance model only considered the surface reflectance component of the overall skin reflectance.

Overall skin reflectance from human skin can be decomposed into two components: surface reflectance (modelled using BRDF) and subsurface reflection (modelled using BSSRDF). In facial skin, the subsurface reflection component dominates the overall reflection \cite{donner2008layered}\cite{igarashi2007appearance}. Hence, a skin reflectance model involving only the surface reflectance component cannot achieve photorealistic rendering of human skin. Debevec \etal \cite{debevec2000acquiring} developed a novel capture device called the ``light stage'' which can illuminate a face from a dense set of spherical positions while recording the appearance from multiple viewpoints. Using the images captured in this device, they propose a method to recreate facial appearance under novel illumination and viewpoint. Their method exploits the fact that a given facial appearance under general lighting condition can be represented as linear combination of facial appearance under illumination by point light sources densly distributed over a sphere surface. In other words, if all the possible appearance of a human face lie in a \textbf{N} dimensional space, then the face images captured under illumination by a dense sampling of incident illuminaton direction forms the basis of this vector space. To recreate facial appearance from novel viewpoint, they create a geometric model of the face using structured lighting. The facial appearance from original viewpoint is projected onto this geometric model and appearance from novel viewpoint is computed based on this projected appearance.

The facial appearance projected from the original viewpoint cannot reproduce the shifting and scaling in the measured reflectance function caused by change in viewpoint. Hence, the viewpoint specific changes to diffuse and specular reflectance components from a region in forehead --- a region also selected by Marschner \etal \cite{marschner1999image} --- is used to extrapolate the corresponding deviation in other regions of the face. Specular and diffuse reflectance components are separated using the difference of parallel and cross polarized images. Colorspace analysis is used to separate the diffuse and specular reflectance components in other parts of the face. These separated components undergo shifting and scaling according to novel viewpoint specific scaling and shift observed for a $2\times5$ pixel in forehead region. The specular reflectance component is fitted to the microfacet based rough surface model of Torrance and Sparrow \cite{torrance1967theory}.

The authors aimed to produce realistic rendering of subsurface reflectance phenomena. Hence, the subsurface scattering data was not fitted to any skin reflectance model and instead was only used to determine viewpoint specific changes to subsurface reflectance component for a given illumination environment. Hence, the renderings produced using this method cannot reproduce correct subsurface scattering effect due to heterogeneous illumination environment. Also, it is a data driven technique and hence requires capture of a large number of images ($64\times32=2048$ photographs) resulting in long capture procedure ($1$ min). Moreover, the data driven nature of this method prevents its use for editing or transfer of facial appearance characteristics among the captured subjects i.e. we are locked in the facial appearance space spanned by the captured data.

Hanrahan and Krueger \cite{hanrahan1993reflection} developed a reflectance model which, for the first time, related the physical properties (like refractive index, thickness, absorption and scattering coefficients) of a layered material to the subsurface reflectance properties of that material. They presented a model --- suitable for Computer Graphics --- for reflection of light due to subsurface scattering in a layered material. This model treated a physical material as a layered homogeneous scattering medium. The authors suggested modeling heterogenity in a material using random noise or a texture map. Reflection from outer surface of the material was modeled using the Torrance and Sparrow \cite{torrance1967theory} microfacet model and subsurface scattering was modeled using the proposed reflectance model based on 1D linear transport theory. Rendering of a human face, whose 3D geometry was acquired using a medical MRI scanner, was generated using a two layered model which correspond to the epidemis and demis layers of a human skin. The model parameters for each layer were chosen manually to generate renderings that were close in appearance to real human skin. The authors did not present a method to capture reflectance data required for estimation of the model parameters. Hence, although the model was anatomically motivated and produced acceptable face renderings, the layer parameters used for generating these renderings were not based on measurements from actual human skin.

Jensen \etal \cite{jensen2001practical} took a different approach to modeling subsurface reflectance by introducing the dipole model based on ``diffusion theory'' to the graphics community. The diffusion theory existed in the optics community prior to this but had never been used for subsurface skin reflectance modeling. They modeled subsurface scattering of light using diffusion approximation which is based on the observation that light distribution in a highly scattering media is isotropic. The authors acknowledge insipiration for this model from the use of diffusion theory used in describing the scattering of laser light in human tissue in medical physics research. Unlike \cite{hanrahan1993reflection}, they described a capture device setup to capture reflectance data of real world objects which can be used to estimate all the model parameters. This device focused a beam of white light on the sample material and recorded a High Dynamic Range (HDR) image corresponding to radiance fall off from the point of incident beam i.e. the radially symmetric diffusion profile. Using this setup, they measured the diffusion profile $R_{d}(r)$ of a wide variety of real world objects like milk, human skin, marble, etc. The model parameters, absorption $\sigma_{a}$ and reduced scattering coefficients $\sigma_{s}'$, were estimated from these measured diffusion profiles. Hence, for the first time the measured subsurface scattering characteristics of real world material was plugged into a reflectance model.

The capture device proposed by Jensen \etal is not suitable for facial skin as focused beam of white light may harm the skin during the capture process. Hence, they measured the diffusion profile of skin in the arm region for their experiment and extrapolated the subsurface reflectance properties to skin in other body parts. This reflectance model assumed the scattering medium to be semi-infinite i.e. only one side of the medium had well defined boundary. In other words, this model assumed that every component of the incident light will eventually be reflected back. Hence, the model failed to account for incident light that gets transmitted into the material. Also, this model can only be used with highly scattering media because the diffusion approximation, used by this model, is only applicable to highly scattering medium i.e. $\sigma_{s}' >> \sigma_{a}$.

Weyrich \etal \cite{weyrich2006analysis} overcame this practical limitation of the diffusion profile capture device by building a contact probe consisting of a linear array of optical fibres: one of them being the source fibre and the remaining are detectors. Using this contact probe, all the parameters of the dipole model can be robustly estimated. They modeled skin as a single layered homogeneous scattering medium. They added a spatially varying absorptive film of zero thickness --- called the modulation texture layer --- to simulate inhomogeneous scattering in human skin. Use of texture map (or random noise) was also suggested by \cite{hanrahan1993reflection} to simulate the effects due to hetereogenity in a scattering medium. The parameters of this layer were estimated from albedo map and the dipole model parameters $\sigma_{s}'$, $\sigma_{a}$ were obtained from the diffusion profile captured using their contact probe. They measured 3D face geometry, skin reflectance and subsurface scattering using custom built devices for $149$ subjects of varying age, gender and race. This allowed them to study the variation of subsurface scattering parameters for a large population of skin types. Moreover, flexibility in their reflectance model allowed intutive editing of facial appearance. For example: they presented the results of face renderings obtained by transfer of skin features like freckle and skin type (BRDF and albedo).

Weyrich \etal added a suction pump to the contact probe in order to maintain the contact and position during the capture of diffusion profile. The total capture time of $88$ sec. necessitated addition of the suction feature to the contact probe. It is known that physical pressure alters the normal blood flow mechanism in a human skin. Hence, the scattering and absorption coefficients obtained from the diffusion profile captured using such contact probe may be biased to some extent. Moreover, the design of contact probe limits its use to flat areas of a human face. Hence, they extrapolate the reflectance measurements from forehead, cheek and below the chin to other parts of a human face. Although, the addition of a modulation texture closely reproduces effects due to heterogenity in skin, its makes the reflectance model anatomically implausible.

Ghosh \etal \cite{ghosh2008practical} have described a skin reflectance model which treats overall skin reflectance as the linear sum of four reflectance components: specular, single scattering, shallow scattering and deep scattering. These components are classified according to the depth of skin from which they get reflected. They are able to estimate all the model parameters from just $20$ photographs of human face captured under spherical illumination (developed by \cite{ma2008framework}) and projected lighting condition. The spectral difference between these two sources is compensated by computing a colour transformation matrix which transforms both photographs to a common colorspace. A 24 ColorChecker square and 10 skin patches are imaged under these two illumination condition to compute this colour transformation matrix. First, the specular and single scattering components are separated from overall reflection by exploiting the fact that these components preserve the polarization of incident light. This is also true for single scattering reflectance because the probability of depolarization of light increases exponentially with each additional scattering event. Furthermore, these two components are separated from each other by exploiting another interesting difference between these two components: any non-specular reflectance component that preserves polarization of incident light is treated as single scattering term. The specular reflectance component is modeled using Torrance and Sparrow \cite{torrance1967theory} microfacet model and Hanrahan and Krueger \cite{hanrahan1993reflection} first order single scattering BRDF is used to model the single scattering term.

Multiple scattering is composed of shallow and deep scattering reflectance. The diffuse only image obtained from polarization difference image of parallel and cross polarized images contain the multiple scattering reflectance component. Using the method of Nayar \etal \cite{nayar2006fast}, they separate the multiple scattering reflectance component into direct and indirect reflectance components. They key observation underpinning this separation is that when the frequency of illumination pattern is in the order of thickness of epidermis, then the direct component relates to shallow scattering and indirect component corresponds to the deep scattering reflectance component.

Ghosh \etal modelled subsurface scattering (or multiple scattering) using reflectance in a two layered medium. Although, they do not explicitly model the epidermis and dermis layers in a human skin, they use the notion of deep and shallow scattering to roughly model the light interactions occurring in these two layers. Deep scattering, caused by the bottom layer, was modelled using the dipole model of Jensen \etal \cite{jensen2001practical} which treats the scattering layer to be semi-infinite. The semi-infinite assumption is practical for the bottom layer but not for the top layer. Were the top layer semi-infinite, it would not have a tranmission profile and hence the bottom layer would not receive any portion of the incident light. Hence, the shallow scattering, caused by the top layer, is modeled using the multipole diffuse model of Donner and Jensen \cite{donner2005light}. The transmission profile of the top layer obtained using the multipole model becomes the incident profile for the bottom layer which is modeled using a dipole model.

Ghosh \etal model the overall skin reflectance as the linear sum of four reflectance components: specular, single scattering, shallow scattering and deep scattering. The purely additive nature of these reflectance components prohibits the modelling of phenomenon involving interaction between skin layers; for instance, the epidermal effects on dermal scattering.

Donner \etal \cite{donner2008layered} proposed the ``physiologically most advanced skin reflectance model that is still practical for rendering''\cite{weyrich2009principles}. They described a two layered skin reflectance model which used, for the first time, spatially varying model parameters for each layer to account for heterogeneous light transport in human skin. Using diffuse images captured in $9$ different bands of the visible region, they were able to model spectral dependence of subsurface scattering characteristics. The proposed two layered skin reflectance model has $6$ spatially varying model parameters which relate to physiological skin parameters and are represented as 2D chromophore \footnote{skin constituent that selectively absorb some spectral bands of the incident light} map. The two layers in this model correspond to epidermis (top layer) and dermis (bottom) layers of a human skin. A thin absorbing layer was added between these two scattering layers which corresponds to pigmentation concenterated in a narrow region between epidermis and dermis of actual human skin.

They demonstrated the strength of anatomically motivated reflectance model by generating photo-realistic renderings of human hand from just the user painted 2D chromophore maps corresponding to the $6$ model parameters. In addition to the user painted chromophore map method, they also devised a inverse rendering based approach to estimate the $6$ model parameters from multispectral images of a flat skin sample. They developed a filterwheel based multispectral capture device to capture the multispectral reflectance map of a flat skin sample in the arm region. Scattering in each layer was modeled using the multipole diffusion model of \cite{donner2005light}.

The proposed inverse rendering approach is not scalable to the Multispectral images of the full face. Inverse rendering resembles a ``brute force'' approach in which estimation of model parameters from Multispectral images involves searching a 6D space for model parameter values that minimised the difference between rendered Multispectral images and the captured Multispectral photographs. This strategy is not applicable to estimation of model parameters from Multispectral images of full face because the complex geometry of human face makes the process of inverse rendering computationally intractable. Also, the capture process required the skin surface to be coated with Ultrasound gel. Ultrasound gel has same refractive index as a human skin and hence created a smooth surface over the skin sample under observation. This allowed the use of Fresnel transmission term for estimation of radiance transmitted into the skin. It is impractical to apply the Ultrasound gel to full face for acquisition of Multispectral images.

Ghosh \etal \cite{ghosh2008practical} and Donner \etal \cite{donner2008layered} have proposed the current state-of-the-art skin reflectance models. While the data driven model of Ghosh \etal can estimate model parameters of complete face in natural expressions using just $20$ photographs captured in $5$ sec, the anatomically plausible skin reflectance model of Donner \etal can produce realistic renderings of human hand with just a user painted 2D chromophore map representing the model parameters. On the other hand, the data acquisition procedure of Donner \etal is not scalable to complete human face whereas the reflectance model proposed by Ghosh \etal lacked biophysically meaningful parameters.

\section{Alignment}
Almost all shape and reflectance acquisition system has to deal with motion of non-static objects, like a human face, during the capture process. Marschner \etal \cite{marschner1999image} used a set of three machine readable targets for automatic estimation of the relative position of the light source, sample material and the camera during the $30$ minute capture process. Debevec \etal \cite{debevec2000acquiring} proposed using a head rest to reduce motion during the capture process which lasted for $1$ minute. In reality, it would be extremly difficult to maintain facial expression and position for $1$ minute in spite of a head rest. Weyrich \etal \cite{weyrich2006analysis} used a contact probe with suction to maintain the position of the subsurface reflectance capture device during the $90$ seconds of capture time. Ghosh \etal \cite{ghosh2008practical} have not mentioned how they corrected for subject motion during the capture of $20$ photographs in $5$ seconds. Donner \etal \cite{donner2008layered} use a filterwheel based multispectral capture device to capture $9$ multispectral photographs of a skin sample in the arm region. As compared to human face, it is relatively easy to maintain position of arm during the capture process. They marked a rectangular region in the skin sample which allowed them apply rigid alignment methods.

Recently, Wilson \etal \cite{wilson2010temporal} have developed the Joint Photometric Alignment technique for the registration of gradient images captured in a Light Stage. Traditional optical flow based alignment techniques were not applicable to the alignment of gradient images as the ``brightness constancy'' assumption is violated in each of these images. Wilson \etal exploited the complement image constraint to devise an iterative algorithm for alignment of gradient images. Photometric normals computed from aligned gradient images can recover the fine surface details like wrinkles, scar, etc present in a human face.

\section{Real Time Performance Capture of Human Face}
Marker based facial motion capture is widely used for the capture of geometric deformations in human face during a dynamic performance. The 3D position and velocity of these reflective markers are used as cues to the 3D motion of body structure to which these markers are attached. A limitation of this method is that it requires placement of a very large number of markers on the target face in order to accurately model the 3D motion of each facial muscle. In addition to inconvenience caused by these markers during facial performance, there exists a limit to which these markers can be attached to a face. This limitation prevents from acquiring fine scale geometric details of face muscles during a dynamic performance. Human observers have a mastery in detecting unnatural facial motion caused by sparse distribution of these markers. Hence, marker based motion capture techniques are not used for close up shots of the human faces.

Furukawa and Ponce \cite{furukawa2009dense} have developed a markerless 3D motion capture method for human faces. They track the nonrigid motion of vertices in the 3D mesh of the face obtained from multiview stereo technique. Their method is capable of dealing with unreliable texture information due to fast motion, self occlusion, etc. However, this method involves a data intensive capture process and is affected by specular highlights on a face.

Wilson \etal \cite{wilson2010temporal} have developed facial performance geometry capture method which is not data intensive and can capture highly detailed facial geometry without requiring expensive and complex setup of high speed photography. They capture a set of gradient and complement gradient spherical illumination images which flank the constant illlumination image, also called the tracking frame. Using their Joint Photometric Alignment method, the gradient images are aligned to the tracking frame which allowed computation of photometric normal at each tracking frame. These photometric normals are warped to the intermediate gradient frames according to the flow fields computed by the alignment stage. This process is called ``Temporal Upsampling'' because it increases the effective performance capture frame rate by adding warped photometric normals at the temporal position of intermediate gradient frames. They use the spherical gradient photometric stereo technique to recovery very high resolution photometric normal at each tracking frame. Hence, their dynamic performance capture algorithm is able to capture the motion of very fine facial features like wrinkles, pores visible during skin deformation, etc. Moreover, the absence of makers on face allows capture of natural facial expressions.

The proposed Joint Photometric Alignment method is an iterative process requiring two optical flow computation in each iteration. Because this alignment procedure has to be applied to each of the gradient and complement gradient image pairs separately, the computational cost of performance capture using this method is very high. This does not affect the practicability of this performance capture method because the alignment is a post-processing operation which can be carried out offline.

\section{Stimuli for Psychology Experiments}
Research in Psychology and Neuropsychology of face perception has always relied on Computer Vision and Computer Graphics community for stimuli image dataset required for their experiments. Ability to control different aspects of facial appearance is the key to success of these experimental procedures designed to unravel the face representation and processing mechanisms of the visual cortex in human brain.

Caharel \etal \cite{caharel2009recognizing} used a 3D Morphable Model to generate stimuli images for studying the time course (i.e. temporal sequence) for processing of 3D shape and 2D skin reflectance information of a human face. Their stimuli image contained face images in which texture and shape information of the test subjects were controlled. Although, the 3D Morphable Model produces facial rendering close to natural human faces, it does not include the high frequency skin texture detail. Lack of detailed skin texture, which is known to contribute to face perception, can bias the results of such psychology experiments.

Recently, we explored the application of Light Stage in generating stimuli image dataset for the psychology experiment conducted by Jones \etal \cite{jones2010how}. This experiment investigated the neural representation of 3D shape and 2D skin reflectance information of a human face. Using the image data captured in our Light Stage, we were able to separate 3D shape and 2D skin reflectance information for a given face. Spherical illumination of a Light Stage ensured that the texture images were not affected by shadows. Also, these ``texture only'' face images included all the high frequency facial skin details like wrinkles, mole, frackles, etc.

\chapter{Design and Calibration of the Multispectral Light Stage}
\label{ch:design_and_calib_of_mls}
The Photometric stereo technique was developed by Woodham \cite{woodham1980photometric} to determine the surface geometry of each image point using images captured by varying the direction of incident illumination while keeping the view direction constant. Ma \etal \cite{ma2007rapid} have proposed spherical gradient photometric stereo --- an extended version of original photomeric stereo --- for acquisition of high resolution shape and reflectance information. A spherical illumination environment is pivotal to this state-of-the-art shape and reflectance acquisition technique because it requires images of an object captured under spherical gradient and constant illumination environment. In this chapter, we discuss the design and calibration of a device that can be used to create a spherical gradient illumination environment. We also propose an extended version of the required acquisition device setup to allow capture of multispectral images in a spherical illumination environment.

\begin{figure}[htbp]
  \centering
  \subfigure[Ideal diffuse reflectance lobe]{
    \label{fig:design_and_calib_of_mls_diff_refl_lobe}
    \includegraphics{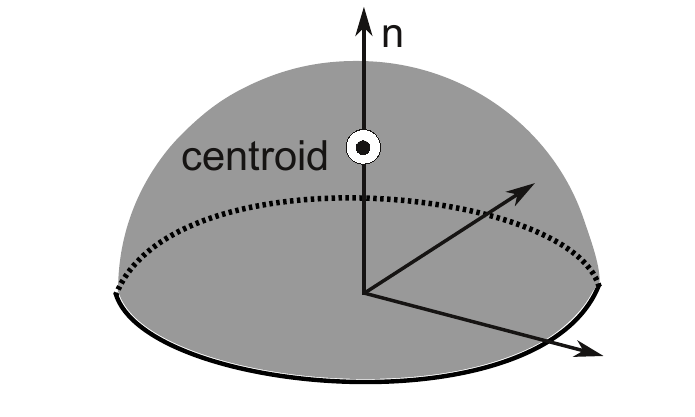}
  }
  \subfigure[Ideal specular reflectance lobe]{
    \label{fig:design_and_calib_of_mls_spec_refl_lobe}
    \includegraphics{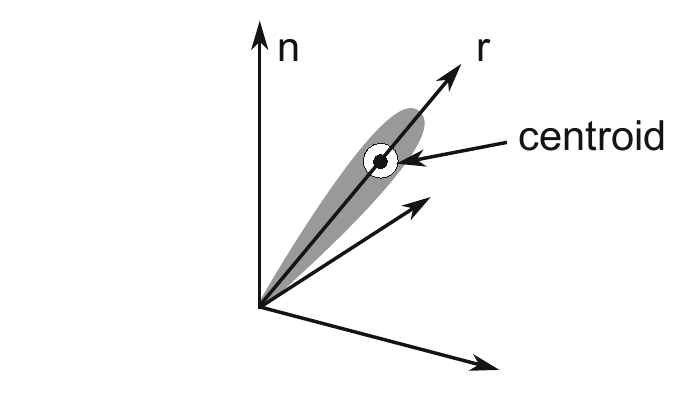}
  }
  \caption{Centroid of ideal specular and diffuse reflectance lobe}
  \label{fig:design_and_calib_of_mls_spec_diff_refl_lobe}
\end{figure}

The mass centroid of an ideal diffuse reflectance lobe coincides with the surface normal $\vec{n}$. For the specular reflectance lobe, the centroid coincides with the reflection vector $\vec{r}$ as shown in \figurename\ref{fig:design_and_calib_of_mls_spec_diff_refl_lobe}. The centroid $(x_0, y_0, z_0)$ of a diffuse or specular reflectance function $f(x,y,z)$ can be computed by integrating it with a linear gradient. Mathematically,
\begin{equation}
    (x_0, y_0, z_0) = \frac{1}{\int_{-1}^{1}f(x,y,z) \textrm{dx}} \left( \int_{-1}^{1}xf(x,y,z) \textrm{dx}, \int_{-1}^{1}yf(x,y,z) \textrm{dy} ,\int_{0}^{1}zf(x,y,z) \textrm{dz}.
\right)
\label{eq:design_and_calib_of_mls_centroid_formula}
\end{equation}

The insight of Ma \etal \cite{ma2007rapid} was to show how to estimate the diffuse and specular reflectance centroids using spherical gradient illumination. They proposed the spherical gradient photometric stereo technique which suggests that when integrated with a linear illumination gradient in the X, Y or Z direction, the corresponding component of the reflectance centroid, and hence surface normal, can be recovered. The key observation underpinning this approach is evident when we look at the radiance equation for diffuse and specular reflection:

\begin{equation}
 r = \int_{\Omega}P(\omega)R(\omega,n) \textrm{d}\omega,
 \label{eq:design_and_calib_of_mls_radiance_eq}
\end{equation}
where $P(\omega)$ is the intensity of light incident from direction $\omega$ and $R(\omega,n)$ is the Lambertian or specular Bidirectional Reflectance Distribution Function (BRDF). According to (\ref{eq:design_and_calib_of_mls_radiance_eq}), if we replace the illumination environment $P(\omega)$ with a spherical gradient illumination in X,Y or Z, the radiance value recorded by an imaging device is related linearly to the centroid of Lambertian or specular BRDF $R(\omega,n)$. In the next section, we will discuss design and calibration of a ``light stage'' : a device proposed by Ma \etal \cite{ma2007rapid} to create a spherical illumination environment.

\section{Creating the Spherical Illumination Environment}
\label{ch:design_and_calib_of_mls:sec:sph_grad_illum}
\begin{figure}[htbp]
  \centering
  \includegraphics{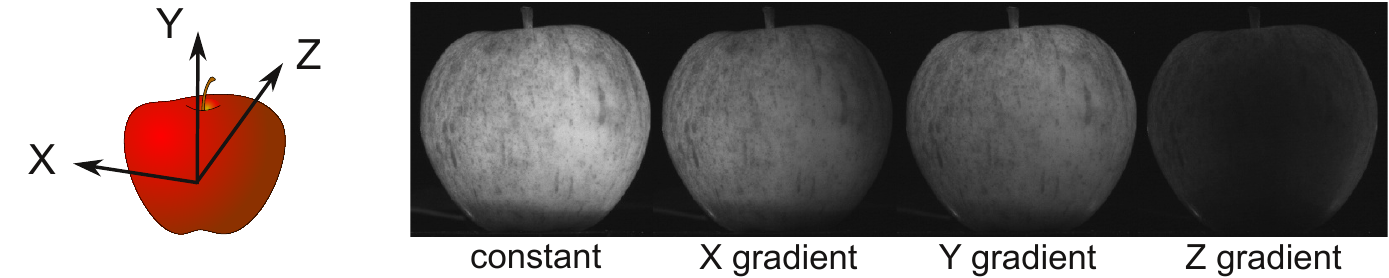}
  \caption{Images of an apple captured under spherical gradient and constant illumination}
  \label{fig:design_and_calib_of_mls_apple_gradient_img}
\end{figure}

\begin{figure}[htbp]
  \centering
  \includegraphics{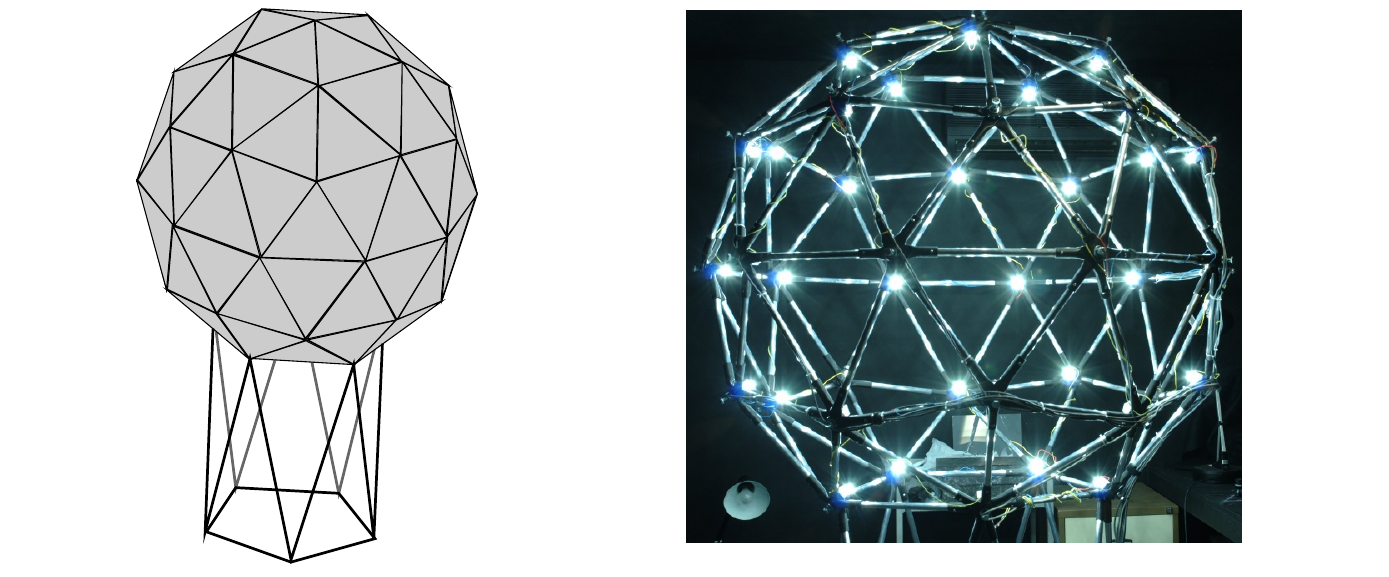}
  \caption{Our Light Stage}
  \label{fig:design_and_calib_of_mls_light_stage}
\end{figure}

Spherical illumination refers to an illumination environment in which every surface patch of an object receives illumination incident from every direction of its visible hemisphere. \figurename\ref{fig:design_and_calib_of_mls_apple_gradient_img} shows the images of an apple illuminated by spherical gradient and constant illumination.
An object placed at the centre of a sphere can be illuminated by spherical illumination by using light sources distributed evenly and finely over the surface of that sphere. Ma \etal \cite{ma2007rapid} argued that as the position of edges and vertices of a twice subdivided icosahedron closely approximates the surface of a sphere, LEDs attached to these positions can create spherical illumination. They proposed a device called ``light stage'' (or led sphere) that consists of 156 LEDs attached to the edges and vertices of a twice sub-divided icosahedron. \figurename\ref{fig:design_and_calib_of_mls_light_stage} shows an image of our light stage of diameter $1.58$ meter consisting of 41 LEDs attached only to the vertices of twice sub-divided icosahedron. In \ref{ch:design_and_calib_of_mls:sec:light_src_sel}, we discuss the reason behind using only 41 LED in our Light Stage.

\begin{figure}[htbp]
  \centering
  \subfigure[Constant illumination]{
    \label{fig:design_and_calib_of_mls_const_inten_plot}
    \includegraphics{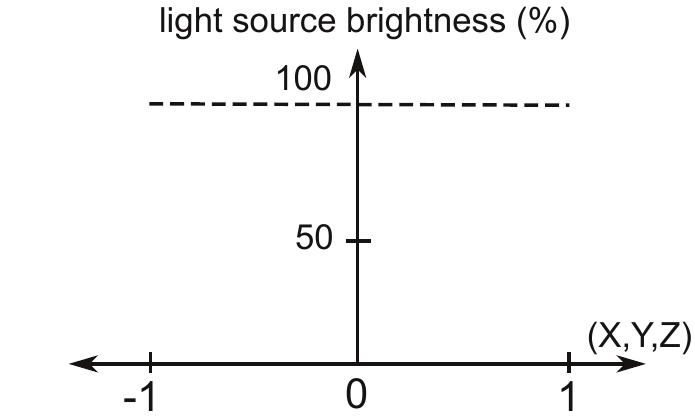}
  }
  \subfigure[Gradient illumination]{
    \label{fig:design_and_calib_of_mls_grad_inten_plot}
    \includegraphics{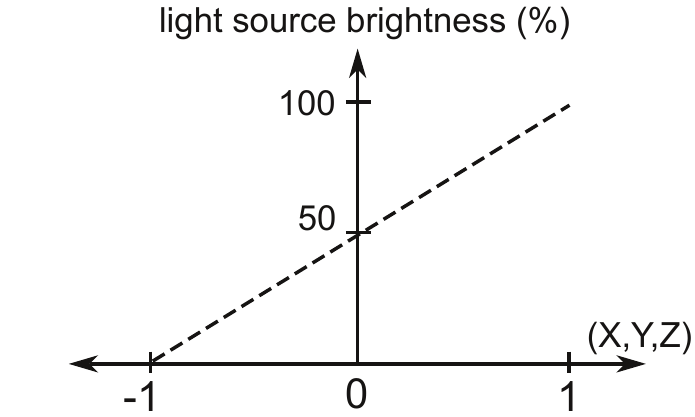}
  }
  \caption{Light source intensity for constant and gradient illumination environment}
  \label{fig:design_and_calib_of_mls_const_grad_inten_plot}
\end{figure}

Constant illumination is created by switching all the LEDs to their maximum brightness level as shown in \figurename\ref{fig:design_and_calib_of_mls_const_inten_plot}. For the X, Y or Z
gradient illumination environment, the intensity of each LED is proportional to the X, Y or Z coordinate of their 3D position respectively. If the 3D position coordinate of each LED is normalized i.e.: $||(x,y,z)|| = 1$ then \figurename\ref{fig:design_and_calib_of_mls_grad_inten_plot} depicts the plot of LED intensity for gradient illumination environment. We can setup a gradient illumination environment by assigning each LED an intensity level that is proportional to their 3D position with respect to the center of the light stage. Hence, the knowledge of light source 3D position is essential to setup a gradient illumination environment.

\subsection{Selection of the Light Source}
\label{ch:design_and_calib_of_mls:sec:light_src_sel}
We have extended the basic light stage design of Ma \etal \cite{ma2007rapid} to achieve the following additional functionalities
\begin{itemize}
  \item simultaneous capture of cross polarized images using a polarizing beam splitter
  \item Multispectral capture capability using a set of narrow band optical filters
\end{itemize}
We have selected the VIO (Vio/3.6W/741) High–Power White LED (manufactured by General Electric Illumination) as the light source for our light stage because :
\begin{itemize}
  \item the light reaching the camera sensor is attenuated by the light source polariser($<50\%$ transmission), optical filter($<90\%$ transmission) and the polarizing beam splitter($<50\%$ transmission). Hence the camera sensor receives only $~22\%$ of the total emitted light even if we image a perfect reflector. The VIO LED has the brightness of 196 lumens \cite{ge2009viodatasheet}. This level of brightness is adequate to image human skin when the attenuation factor of the capture device is $0.78$. 
  \item The $180^{\circ}$ beam angle of these LED provide complete coverage of large objects like human face in a small light stage of diameter $1.58$ meter.
\end{itemize}
\begin{figure}[htbp]
  \centering
  \includegraphics{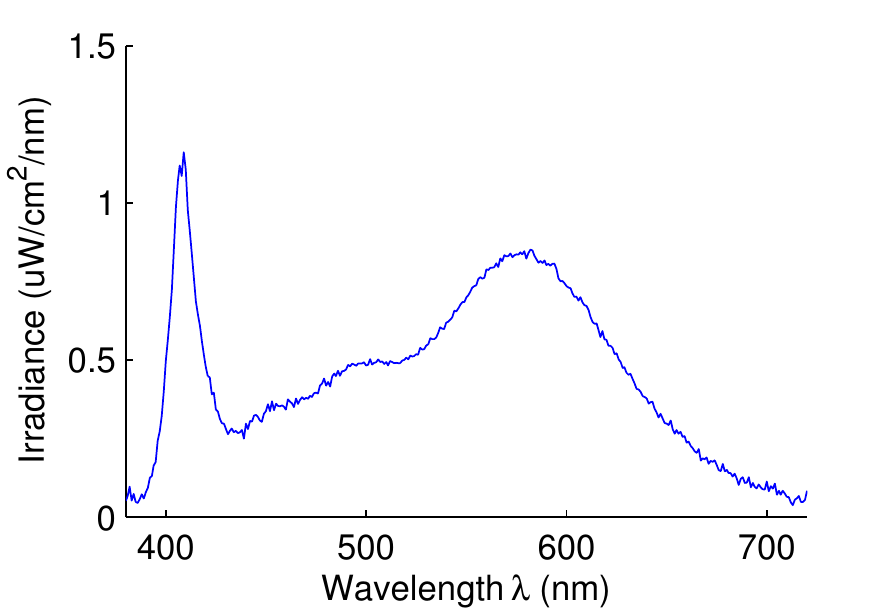}
  \caption{Emission spectrum of VIO (Vio/3.6W/741) LED measured using our CCD spectrometer}
  \label{fig:design_and_calib_of_mls_vio_emis_spectrum}
\end{figure}
The data obtained from our Multispectral light stage can be used to obtain parameters of skin reflectance models like Donner \etal \cite{donner2008layered}. It is known that there occurs peak absorption by human skin chromophores in the $400-450$nm visible range band \cite[Fig. 7]{donner2008layered}. This fact can be exploited to obtain accurate model parameters for parameteric skin reflectance models. \figurename\ref{fig:design_and_calib_of_mls_vio_emis_spectrum} shows that apart from peak emission in $550-600$nm, the VIO LED also has peak emission in the 380nm to 420nm range: a common feature of most LEDs. This behaviour is ideal for capturing the multispectral reflectance map of human skin.

Because of the high cost of VIO LED, we decided to attach these light sources only to the vertices (and not to the edges) of a twice subdivided icosahedron. Although, this design decision causes extreme ``light discretization'', we have developed an algorithm in section \ref{ch:mls_data_processing:sec:minimal_image_sets} to expliot the complement gradient constraint in order to reduce the effects of inter-reflection, ambient occlusion and ``light dicretization''.

\subsection{Estimation of Light Source's 3D Position}
\label{ch:design_and_calib_of_mls:sec:led_3d_pos_est}
From the discussion in \ref{ch:design_and_calib_of_mls:sec:sph_grad_illum}, it is evident that the knowledge of 3D position of each light source is essential for the setup of spherical gradient illumination environment. These 3D positions should be represented with reference to the center of the light stage as the object placed at the center of the light stage needs to be illuminated by gradient illumination. The 3D positions of each light source can be estimated by manual measurement or by using Computer Aided Design (CAD) drawing of the light stage. However, this method does not provide accurate measurement of 3D position. Moreover, due to limitations related to manufacture of geodesic domes, there is always some asymmetry and deformation introduced during assembly of the light stage.

We estimate the 3D position of each light source in a viewer centred coordinate system by exploiting the relationship between light source position and the position of its specularity on a mirror ball placed at the centre of the light stage.

\subsubsection{Determining the Location of Specular Highlight}
\label{ch:design_and_calib_of_mls:sec:led_3d_pos_est:subsec:spec_hglt_loc}
\begin{figure}[htbp]
  \centering
  \includegraphics{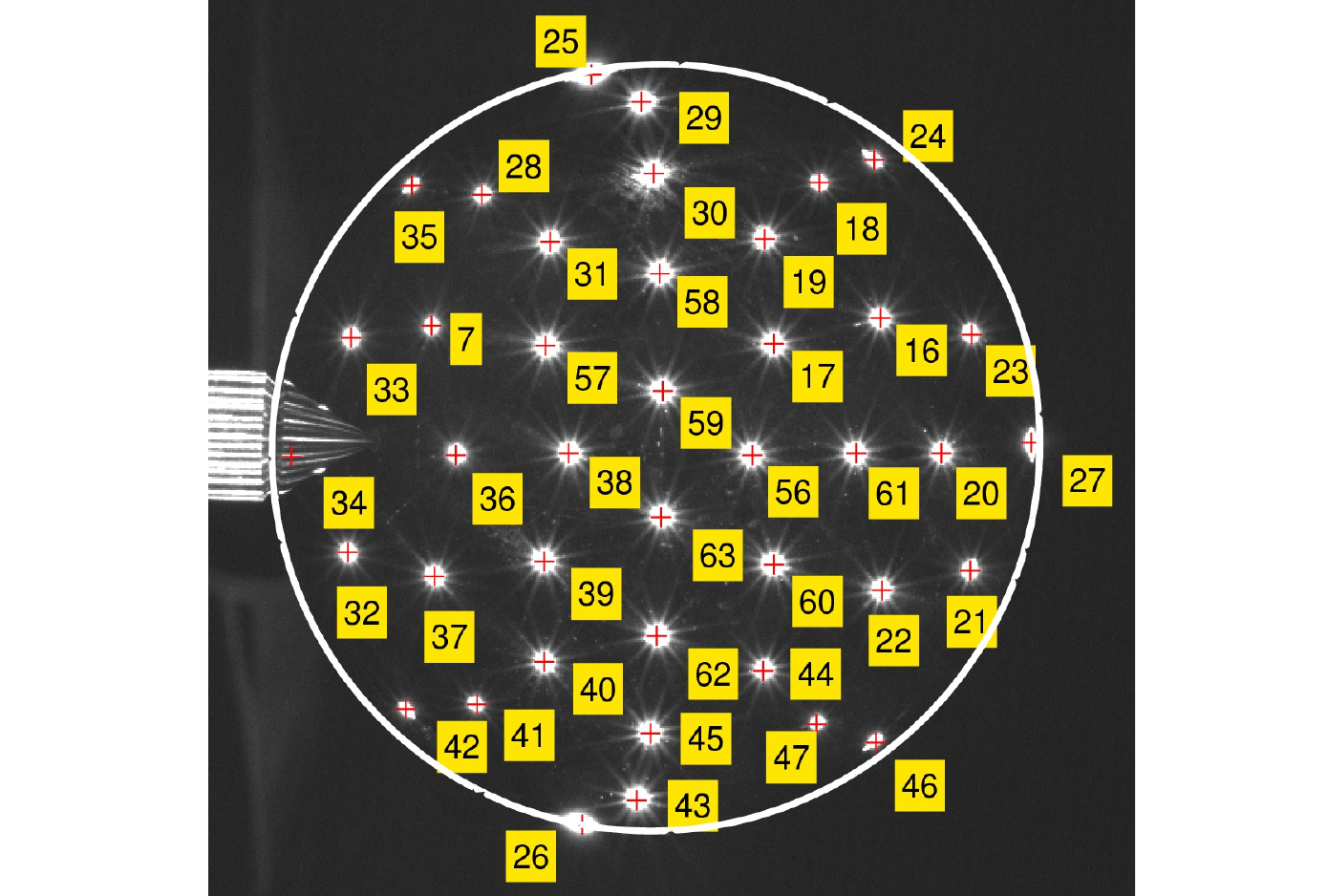}
  \caption{Centroid $h(x,y)$ of specular highlight depicted in a full illumination mirror ball image (numbers correspond to light source unique identifier)}
  \label{fig:design_and_calib_of_mls_spec_highlt_loc}
\end{figure}

We place a $76.2$mm hardened chrome steel ball bearing (mirror ball whose boundary\footnote{The edge of cylindrical rod supporting the mirror ball tapers to a smaller radius to provide threading for screws and this forms the contact point for mirror ball. Therefore, the white boundary fit on the far left of \figurename\ref{fig:design_and_calib_of_mls_spec_highlt_loc} is correct.} is shown in white in \figurename\ref{fig:design_and_calib_of_mls_spec_highlt_loc}) at the center of the light stage and capture its photograph when each of the light sources are switched on individually. These images contains the specular highlight corresponding to each light source. The captured images are preprocessed with morphological erosion and dialation operations to remove any stray bright spots and make the specular highlight more symmetric. The centroid $h(x,y)$ (as shown in \figurename\ref{fig:design_and_calib_of_mls_spec_highlt_loc} by red cross hair) of bright spot in each image forms the location of specular highlight caused by a given light source.

In \figurename\ref{fig:design_and_calib_of_mls_spec_highlt_loc}, notice that there is no specular highlight for LED $34$. The reason being that the specular highlight caused by this LED falls in the blind spot region of the sphere surface visible in the captured images. This blind spot is caused by the stand that supports the metallic sphere. These specular highlight locations can be determined by interpolation of symmetric position of neighbouring light sources in the light stage.

\subsubsection{Mirror Ball Sphere Centre}
\label{ch:design_and_calib_of_mls:sec:led_3d_pos_est:subsec:sphere_center}

The 3D coordinate of centre of the mirror ball is essential for the computation of each light source's direction. We apply the method proposed by Wong \etal \cite{wong2008recovering} to recover the sphere center. First, we manually select at least 6 points on the conic $C$ formed by the boundary of the mirror ball in its image. Using the direct least square fitting method of Fitzgibbon \etal \cite{fitzgibbon1999direct}, we obtain the parameters $(a,b,c,d,e,f)$ that define the conic $C$ such that any point $x$ lying on $C$ satisfies the equation
\[
 \tilde{x}^{T} C \tilde{x} = 0 \qquad \textnormal{where}, \quad C = \left[
 \begin{matrix}
  a & b/2 & d/2 \\
  b/2 & c & e/2 \\
  d/2 & e/2 & f
 \end{matrix}
\right],
\]
and, $\tilde{x}$ is the homogeneous coordinate representation of $x$. The result of this fitting process is depicted by the white conic shown in \figurename\ref{fig:design_and_calib_of_mls_spec_highlt_loc}.

The calibration matrix $K$ was computed using the Matlab camera calibration toolbox \cite{bouguet2004camera}. To remove the effect of camera calibration matrix $K$, we normalize the image with $K^{-1}$. This normalization transforms conic $C$ to a normalized conic $\hat{C} = K^{T}CK$. Using singular value decomposition, we diagonalize conic $\hat{C}$ into
\begin{equation}
  \hat{C} = MDM^{T} = M \left[
 \begin{matrix}
  a & 0 & 0 \\
  0 & a & 0 \\
  0 & 0 & b
 \end{matrix}
  \right] M^{T}.
\end{equation}
Finally, the sphere centre can be computed using
\begin{equation}
  \textrm{S}_{c} = \textrm{M} \left[
 \begin{matrix}
  0 & 0 & d
 \end{matrix}
\right]^{T} \qquad where \quad d = R \sqrt{\frac{a+b}{b}}.
\end{equation}
Here, $R$ is the radius of mirror ball and $d$ is the distance between camera center and sphere center. Wong \etal \cite{wong2008recovering} have also proved that the light direction estimated from an observed specular highlight in an image of a sphere will be independent of the radius used in recovering the location of the sphere center. It is important to recognise that this observation is valid only when the light sources are placed at infinity. In this experiment, the light sources are present close to the mirror ball and therefore requires accurate measurment of mirror ball radius to recover correct values of $d$ and $S_{c}$ using \cite{wong2008recovering}. Very small deviation between manually measured values for $d$ and $S_{c}$ and that measured using \cite{wong2008recovering} (as shown in \tablename\ref{tbl:design_and_calib_of_mls_man_auto_msr_d_sc}) support the fact that manual measurement of mirror ball radius was quite accurate.
\begin{table}
  \centering
  \caption{Manual and automatic measurement of $d$ and $S_{c}$}
  \begin{tabular}{ |l |c |c| }
  \hline
        & Manual (mm) & Using Wong \etal \cite{wong2008recovering} (mm)\\ \hline
    $d$ & $890$ & $898.35$ \\
    $S_{c}$ & $(0, 0, 890)$ & $(21.33, -16.56, 897.94)$ \\  \hline
  \end{tabular}
  \label{tbl:design_and_calib_of_mls_man_auto_msr_d_sc}
\end{table}

\subsubsection{Light Source Position Estimation}
\begin{figure}[htbp]
  \centering
  \includegraphics{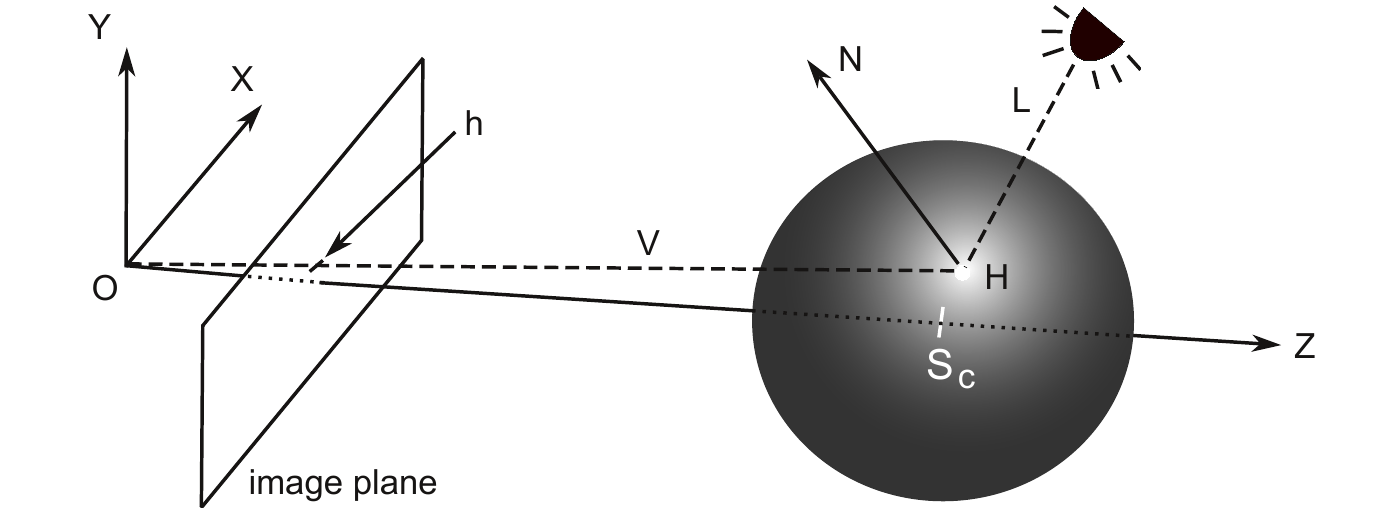}
  \caption{Estimation of light source 3D position using position of its specularity in a mirror ball}
  \label{fig:design_and_calib_of_mls_light_src_pos_est_illus}
\end{figure}
In \figurename\ref{fig:design_and_calib_of_mls_light_src_pos_est_illus}, $H(X,Y,Z)$ and $h(x,y)$ represent the location of specular highlight on the surface of sphere and the image plane respectively. The ray from the camera center $O$ to the location of specular highlight $H$ on the sphere surface forms the view vector $V$. $L$ represents the light source direction and $N$ is the surface normal at point $H$.

From the discussions in previous two sections (\ref{ch:design_and_calib_of_mls:sec:led_3d_pos_est:subsec:spec_hglt_loc} and \ref{ch:design_and_calib_of_mls:sec:led_3d_pos_est:subsec:sphere_center}), we have the values for sphere center $S_{c}$ and image plane location of specular highlight $h(x,y)$. However, to estimate the light source direction $L$, we need the value of one more quantity $H$ : the location of specular highlight on the surface of mirror ball.

To determine the values of $H$, we first construct a ray $\vec{l}$ originating at the camera center $O$ through the pixel coordinate of the specularity in the image plane $h(x,y)$. If $K$ is the camera calibration matrix and $\tilde{h}=[x \; y \; 1]$ represents $h(x,y)$ in homogeneous coordinates, then
\begin{equation*}
  \vec{l} = K^{-1} \tilde{h} + O.
\end{equation*}
The location $H(X,Y,Z)$ of specular highlight on the mirror ball is the point of intersection of ray $\vec{l}$ and a sphere centered at $S_{c}$ with radius $R$. \cite[p116]{pharr2010physically} discusses the method to compute the point of intersection of a ray and a sphere.

\begin{figure}[htbp]
  \centering
  \includegraphics{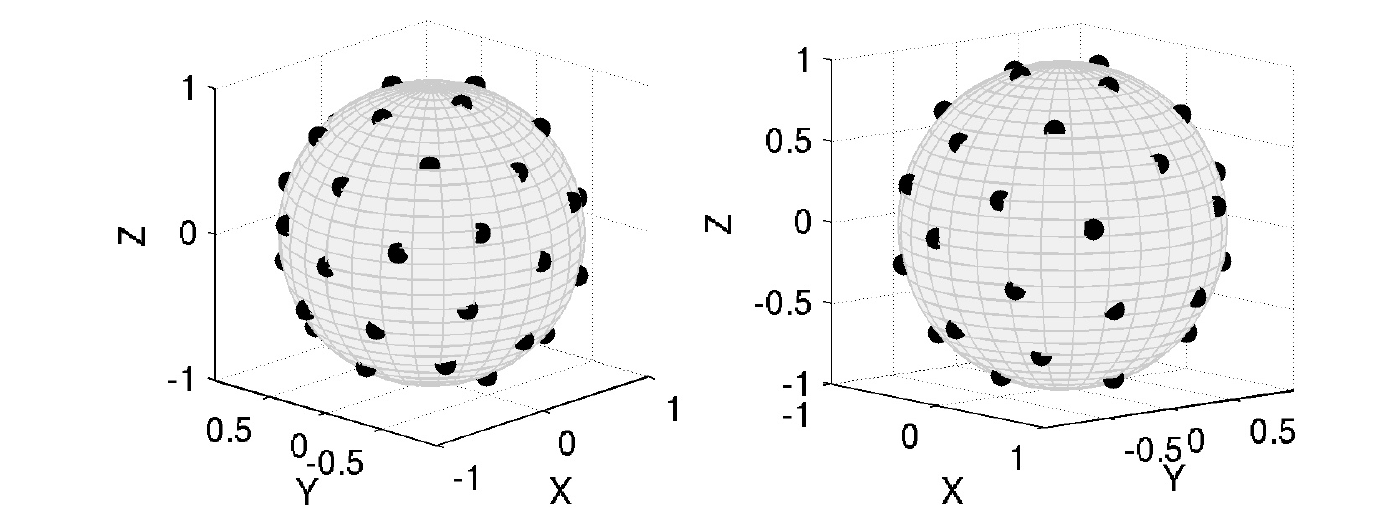}
  \caption{Position of light sources depicted as black spots on a unit sphere}
  \label{fig:design_and_calib_of_mls_light_src_pos_on_unit_sphere}
\end{figure}

With all these measurements to hand, we can now compute the light source direction vector $L$ using
\begin{equation}
  L = (2N.V)N - V
\end{equation}
where, $V = \frac{H-O}{|H-O|}$ and $N = \frac{H-S_{c}}{|H-S_{c}|}$. The positions of light sources recovered using this method are depicted in \figurename\ref{fig:design_and_calib_of_mls_light_src_pos_on_unit_sphere}

\subsection{Light Source Intensity and Camera Shutter Controller}

The LED controller used in this project was designed and built by Cooper \etal \cite{cooper2010geodesic}. This controller is based on an MBED\footnote{\url{http://www.mbed.org}} board and PCA9685\footnote{\url{www.nxp.com/documents/data_sheet/PCA9685.pdf}} I2C LED controller. The MBED (LPC1768) board acts as the control hardware for PCA9685 and camera shutter. PCA9685 is 16 channel I2C LED controller that uses 12 bit (4096 brightness levels) Pulse Width Modulation (PWM) to control LED brightness. The LED controller \cite{cooper2010geodesic} uses four PCA9685 to provide control interface for 41 LEDs in our Light Stage. As PCA9685 is controlled using I2C bus, this design can be easily extended to provide control interface for
even larger number of LEDs.

We have used the ``Geodesic Light Dome'' designed by Cooper \etal \cite{cooper2010geodesic} to control all the light sources and the camera shutter in our Light Stage. An MBED (LPC1768) board acts as the control hardware for the camera shutter and PCA9685 LED controller. PCA9685 is  16 channel I2C LED controller that uses 12 bit (4096 brightness levels) Pulse Width Modulation (PWM) to control LED brightness. We have used four PCA9685 to provide the control interface for 41 LEDs in our Light Stage. As PCA9685 is controlled using I2C bus, this design can be easily extended to provide control interface for even larger number of LEDs. The MBED board provides a ``C'' like programming environment for the control of LED intensity and camera shutter.

\begin{figure}[htbp]
  \centering
  \includegraphics{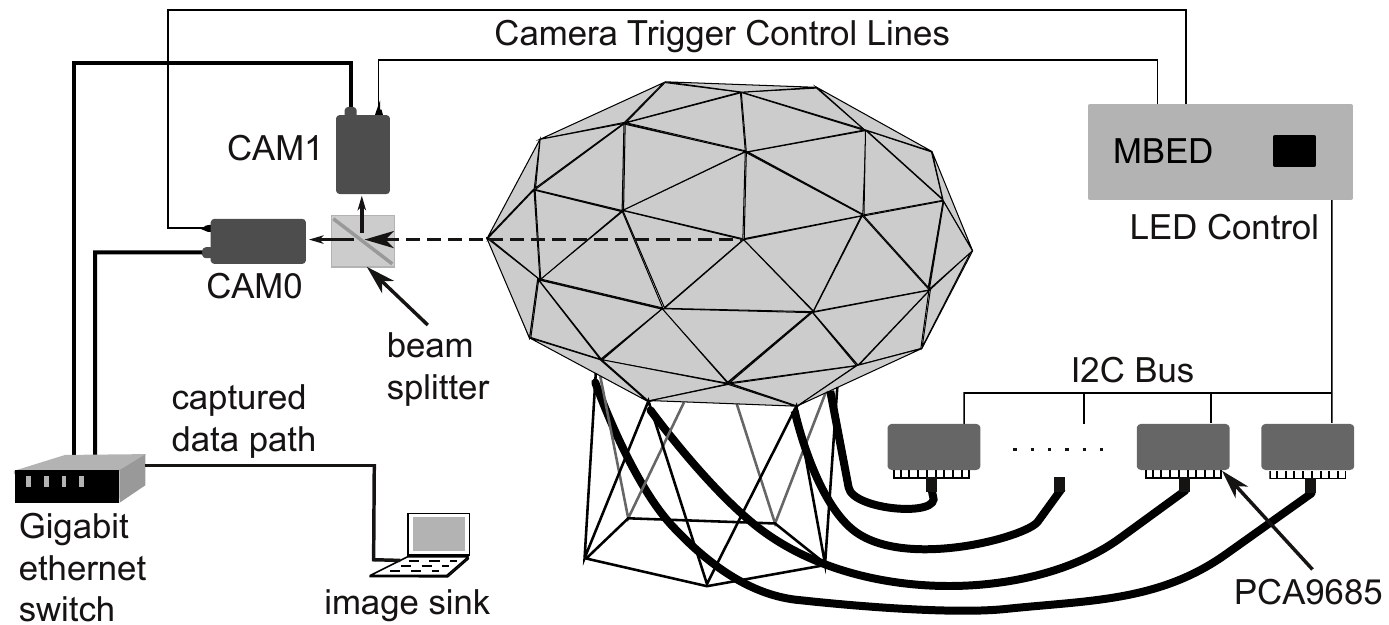}
  \caption{Led intensity and camera trigger control diagram}
  \label{fig:design_and_calib_of_mls_led_and_cam_control_setup}
\end{figure}

We use two JAICM200GE machine vision camera along with a polarizing beam splitter to capture cross polarized images (refer to \ref{ch:design_and_calib_of_mls:sec:simul_cap_cross_pol_img} for details). The connection diagram of MBED, PCA9685, two cameras and a computer is shown in \figurename\ref{fig:design_and_calib_of_mls_led_and_cam_control_setup}.

\subsubsection{LED Intensity Control}
The LED controller \cite{cooper2010geodesic} uses MBED board and PCA9685 to provide a ``C'' like programming environment for LED intensity control. We program the MBED board such that it writes the LED channel identifier (explained in the next paragraph) and corresponding brightness level ($0$ to $4095$) to its I2C bus pins which in turn is connected to the PCA9685. The PCA9685 chip provides a very simple interface (in the form of I2C commands) for LED brightness control and therefore allows us to avoid the intricacies of Pulse Width Modulation (PWM) based LED intensity control.

The LED identifier assigned to specular highlight shown in \figurename\ref{fig:design_and_calib_of_mls_spec_highlt_loc} represents the channel identifier of the corresponding LED. The 3D position of each LED (obtained using the method discussed in \ref{ch:design_and_calib_of_mls:sec:led_3d_pos_est}) is used to create an Intensity Lookup Table (ILT). This lookup table contains the intensity level (0 to 4095) of all the LED for X, Y and Z gradient illumination environment and we store the ILT in the flash memory of MBED. This allows us to setup X, Y or Z gradient illumination in just $23161 \mu s$.

\subsubsection{Camera Trigger Control}
The two JAICM200GE camera are connected to a computer (henceforth referred to as ``image sink'' because it receives all the captured images) via a Gigabit ethernet switch. All the camera functions, including the camera shutter, can be controlled via the GigE vision interface. However, to synchronise the illumination environment setup with the image capture, we use the MBED board (present in the LED controller \cite{cooper2010geodesic}) to control the shutter of both camera. We built a cable to use the digital I/O lines available in the
MBED board of the LED controller \cite{cooper2010geodesic} for control of the two cameras via their General Purpose Input Output (GPIO) interface. The camera is configured to use Pulse Width Modulation (PWM) based shutter control in which the rising edge and falling edge represent shutter open and close events respectively. After configuring the camera shutter control mode to ``Pulse Width Trigger Mode'' (PWC), the ``image sink'' then waits for ethernet packets containing the image data. Now all the capture sequence is handled by the MBED board which is programmed to control the camera shutter using PWM.

MBED board performs the following two operations in a sequence to allow capture
of spherical gradient images in X, Y, Z and constant illumination environment:
\begin{itemize}
  \item Setup the brightness of each LED according to the data in Intensity Lookup Table (ILT) corresponding to required illumination environment
  \item Send a pulse to GPIO pins of both camera such that the rising and falling edge indicate the shutter open and close events respectively.
\end{itemize}
This process is repeated to setup X, Y, Z or constant illumination environment.

The two cameras are triggered simultaneously and hence they start sending ethernet packets, containing the captured image, at the same time. The network switch has sufficient memory to avoid congestion when a single set of gradient images is captured. However, for real time performance capture, a huge amount of data is generated every second: two camera capturing $1200\times1000$ image at 10 bit/pixel ($\sim 2$ byte/pixel) generate $2\times93$ grayscale images per second for 30 fps tracking frame rate. This causes congestion in the network
switch resulting in loss of ethernet packets due to limited buffer memory of the camera and network switch.

\begin{figure}[htbp]
  \centering
  \includegraphics{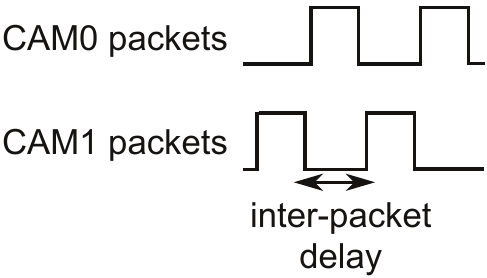}
  \caption{Inter-packet delay parameter introduces controlled amount of delay between ethernet packets generated by a camera}
  \label{fig:design_and_calib_of_mls_camera_inter_pkt_delay}
\end{figure}

The camera manufacturer recommends using the inter-packet delay feature available in the camera to avoid congestion in the network switch\cite[p22, p26]{jai2009cm200gedatasheet}. They provide a tool to compute optimal inter-packet delay in order to make best use of the available video bandwidth. The inter-packet delay parameter of a camera determines the time
interval delay between two adjacent packets transmitted by the camera to the receiving computer. If this delay time is larger than the packet size of other camera, the ``image sink'' will receive ethernet stream in which the packets from both camera are interleaved as shown in \figurename\ref{fig:design_and_calib_of_mls_camera_inter_pkt_delay}. This allows for optimal use of available video bandwidth. For details on computing the inter-packet delay, refer to \cite[p26]{jai2009cm200gedatasheet}.

\subsubsection{Verification of Camera and Illumination Synchronisation}
\begin{figure}[htbp]
  \centering
  \includegraphics{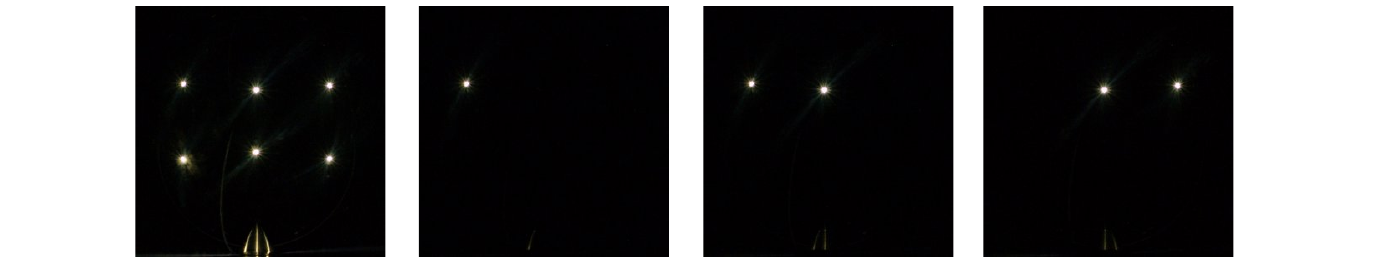}
  \caption{Images captured from four test illumination environments}
  \label{fig:design_and_calib_of_mls_sync_test_pattern}
\end{figure}

The MBED code is executed sequentially. Hence, if we trigger the camera only after setting up intensity of all the LEDs in the Light Stage, the illumination and capture process would always be synchronised. However, to verify correct synchronisation of our setup, we created four test illumination environment. The test illumination environment have the same setup time as the original gradient illumination environment. Moreover, as these test illumination have simple patterns of light, it allows us to verify if there is any ``illumination leak'' from neighbouring illumination environments. \figurename\ref{fig:design_and_calib_of_mls_sync_test_pattern} shows the mirror ball captured under the four test illumination environment. These images support our assumption of correct syncronization.

\section{Diffuse and Specular Reflectance Separation}
\label{ch:design_and_calib_of_mls:sec:diff_spec_separation}
Reflection from a surface consists of diffuse and specular reflectance components. The specular component is caused by light reflected directly from the surface and hence is also called a surface phenomena. The diffuse component results from light rays penetrating the surface, undergoing multiple reflections and refractions, and re-emerging at the surface \cite{nayar1997separaion}. For linearly polarized incident light, specular reflection has polarization oriented perpendicular to the plane of incidence\footnote{plane of incidence at a given surface point is defined as the plane containing view vector $\vec{V}$ and the surface normal $\vec{n}$ at that point} and the diffuse component is essentially unpolarized \cite[p84]{wolff1997polarization}. The fact that the specular component has the same polarization as the incident light is the basis of the ``cross polarization'' technique for separation of the diffuse and specular reflection components.
\begin{figure}[htbp]
  \centering
  \includegraphics{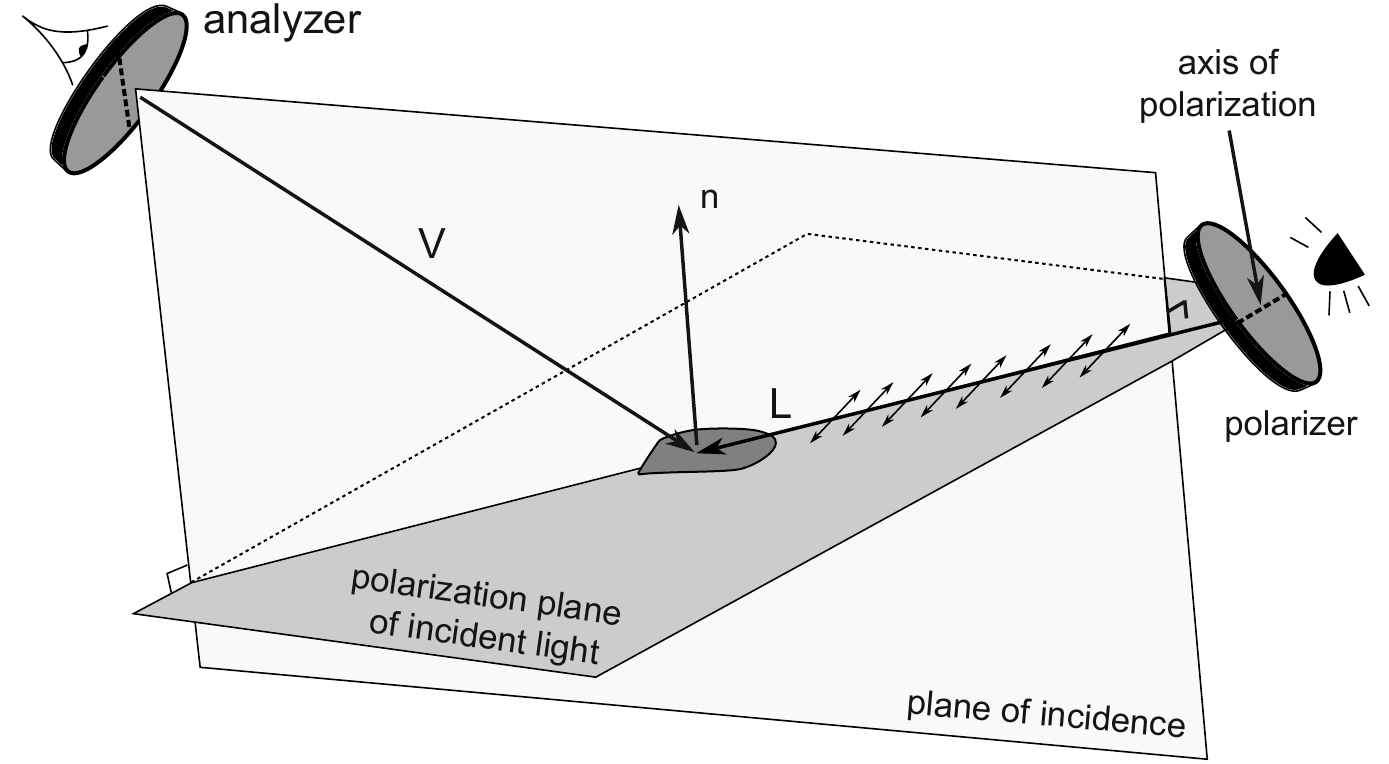}
  \caption{Cross polarization}
  \label{fig:design_and_calib_of_mls_cross_pol_illus}
\end{figure}
The axis of polarization of the linear polariser is such that the plane of polarization of the incident light is orthogonal to the plane of incidence as shown in \figurename\ref{fig:design_and_calib_of_mls_cross_pol_illus}. When the axis of polarization of the analyzer\footnote{the linear polarizer placed in front of the camera} is aligned with the plane of incidence(as shown in \figurename\ref{fig:design_and_calib_of_mls_cross_pol_illus}), only the diffuse component of the reflected light can be observed ($I_{1}$ - diffuse only image). This is beacause the specular component of the reflected light has polarization perpendicular to the plane of incidence. To record both the unpolarized diffuse and the polarized specular reflection ($I_{0}$ - specular and diffuse image), the axis of polarization of the analyzer is oriented orthogonal to the plane of incidence. Hence, we can write the following expression for the two cross polarized images \cite[p40                                                                                                                                                                                                            ]{ma2008framework}
\[
I_{0} = \frac{1}{2}I_{D} + I_{S} , \qquad I_{1} = \frac{1}{2}I_{D} .
\]
Finally, the images containing only the diffuse and specular reflectance components can be recovered using : $I_{S} = I_{0} - I_{1}$ and $I_{D} = 2I_{1}$.

\subsection{Light Source Polariser Orientation}
Spherical gradient illumination requires all the light sources to be distributed uniformly on the surface of a sphere. Hence, we require a spherical field of linear polarization for all the light sources in which all of them have the same plane of polarization. The optimal orientation of each light source polariser is achieved when the ``diffuse only'' image $I_{1}$ of the two cross polarized images contain no specular highlights. In addition to the numerical optimization method, Ma \etal \cite{ma2008framework} also describe a simpler method to obtain such optimal orientation by manually tuning the orientation of each light source polarizer until all the specularity from a mirror ball gets cancelled in one of the cross polarized images. The ``live view'' feature\footnote{real time view of both the cross polarized images} of our acquisition device allowed us to quicky find the optimal orientation of each polarizer.
\begin{figure}[htbp]
  \centering
  \includegraphics{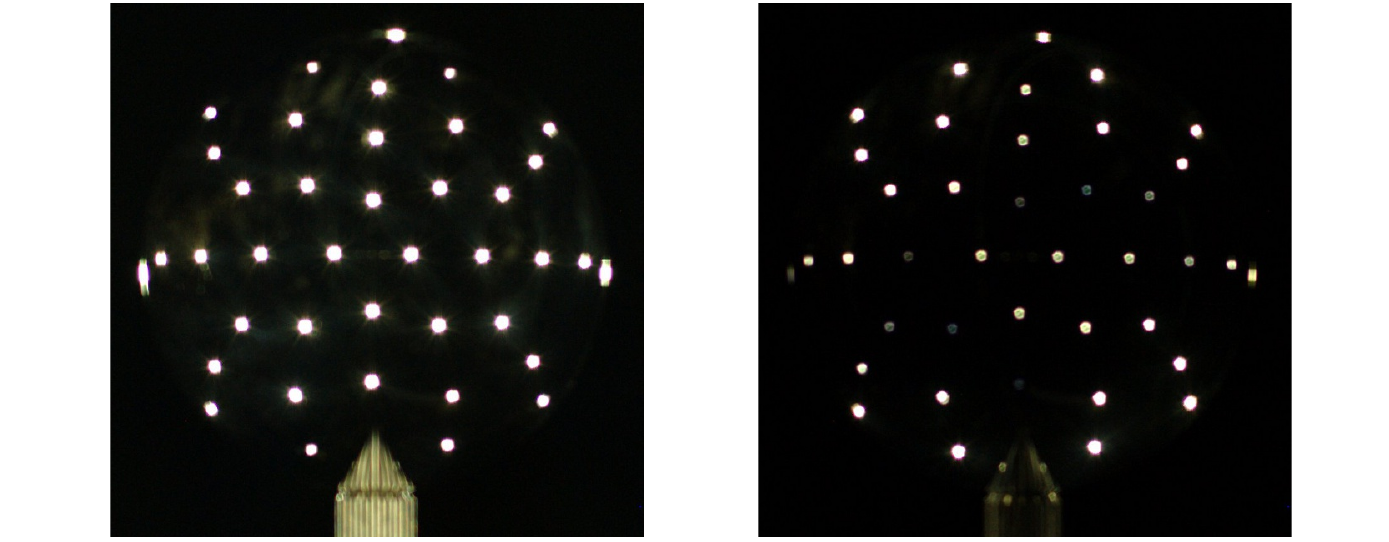}
  \caption{Cross polarized images of a hardened chrome steel ball bearing (mirror ball). (left) Specular and diffuse $I_{0}$ and (right) diffuse only $I_{1}$}
  \label{fig:design_and_calib_of_mls_mirror_ball_cross_pol_images}
\end{figure}

\begin{figure}[htbp]
  \centering
  \includegraphics{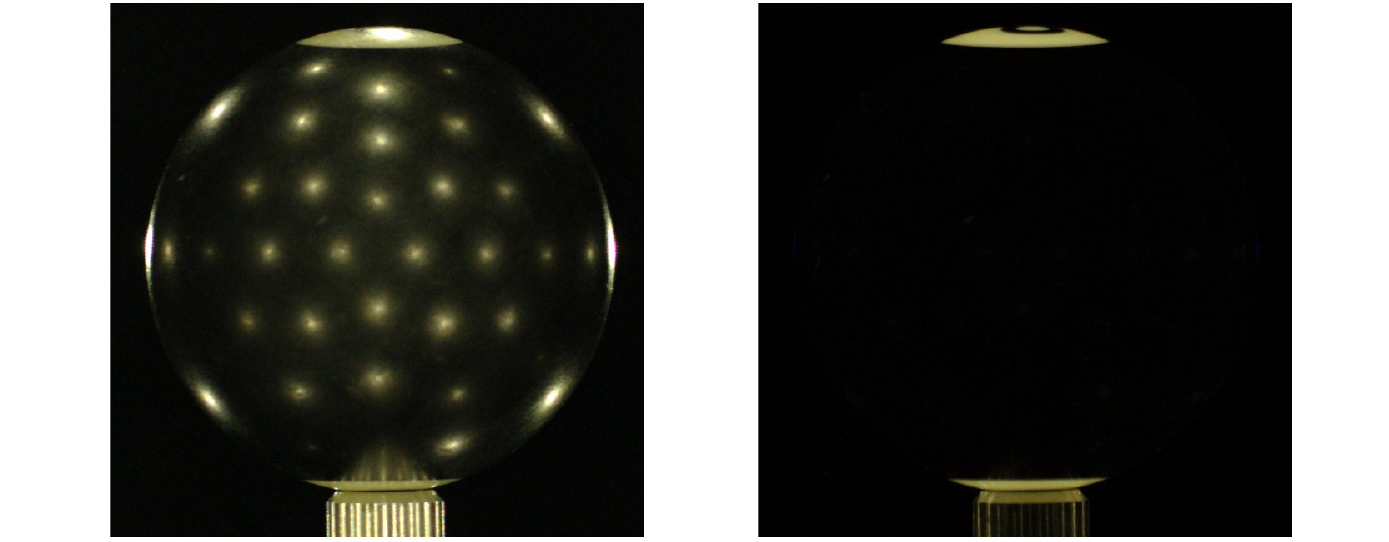}
  \caption{Cross polarized images of a snooker ball. (left) Specular and diffuse $I_{0}$ and (right) diffuse only $I_{1}$}
  \label{fig:design_and_calib_of_mls_snooker_ball_cross_pol_images}
\end{figure}

We came across a peculiar behaviour while looking for optimal polariser orientation using a hardened chrome steel ball bearing (mirror ball) placed at the centre of the light stage. It was not possible to completely cancel specular highlight in the ``diffuse only'' image. This effect was more pronounced for the specular highlights corresponding to the light sources for which the angle of incidence was close to the $90^{\circ}$ as shown in \figurename\ref{fig:design_and_calib_of_mls_mirror_ball_cross_pol_images}. Suspecting the way metallic surface interact with polarized light, we tried using a snooker ball (made of PVC - a dielectric). We were able to quickly find the optimal polarizer orientation using a snooker ball as shown in \figurename\ref{fig:design_and_calib_of_mls_snooker_ball_cross_pol_images}. Ghosh \etal \cite{ghosh2008practical} have also emphasised the use of a dielectric spherical reflector (i.e plastic ball) for the polariser orientation calibration. As we intend to only capture gradient images of dielectric materials (like human face, ceramic and plastic objects, etc), we did not further investigate the peculiar behaviour of metallic surfaces.

\subsection{Simultaneous Capture of Cross Polarized Images Using a Beam Splitter}
\label{ch:design_and_calib_of_mls:sec:simul_cap_cross_pol_img}
\begin{figure}[htbp]
  \centering
  \includegraphics{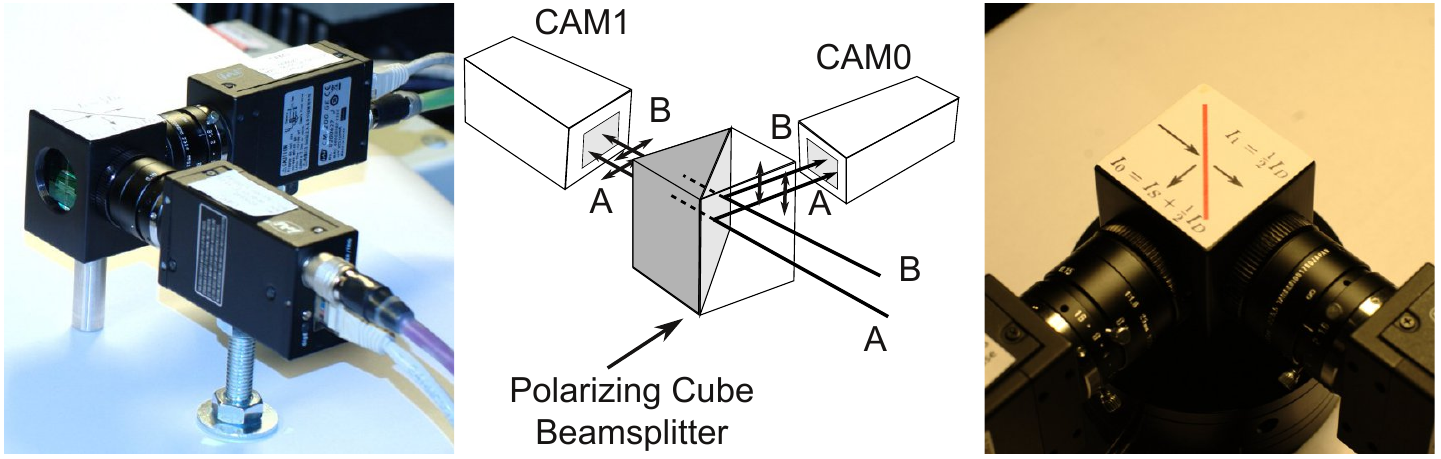}
  \caption{Polarizing cube beam splitter and two camera setup for simultaneous capture of cross polarized images ($I_{0}$ and $I_{1}$)}
  \label{fig:design_and_calib_of_mls_beam_splitter_setup}
\end{figure}
A square linear polariser mounted on a servo motor in front of the camera lens was used by \cite[p46]{ma2008framework} as the analyzer. The servo motor rotated the filter rapidly to allow capture of cross polarized images. The polariser rotation time required by this mechanical setup caused some delay in the capture of the two cross polarized images. In the case of static objects, this does not cause any problem. However, when cross polarized images of a human face are captured by such servo motor based setup, slight motion between the two images cannot be avoided. So the two cross polarized images are not in perfect registration and thus requires some alignment before the diffuse and specular only images can be computed from them. Moreover, the mechanical rotation setup achieved by a servo motor cannot ensure equal amount of rotation in every instance.

To avoid the problems introduced by a mechanical servo motor based system, we used the Techspec\textregistered Polarizing Cube Beam splitter\footnote{\url{http://www.edmundoptics.com/onlinecatalog/displayproduct.cfm?productID=2986}} (25mm, Visible Range) to split the incoming light into $S$ polarized\footnote{In
s polarization, the electric field vector is perpendicular to the plane of incidence.} and $P$ polarized\footnote{In p polarization the electric field vector is parallel to the plane of incidence} components. These two reflection components are recorded by two cameras attached to the two faces of a cube beam splitter as shown in \figurename\ref{fig:design_and_calib_of_mls_beam_splitter_setup}. This setup ensures that both the cameras simultaneously capture the cross polarized images. It is interesting to note that one of the cameras needs to be rotated by $180^{\circ}$ and the captured image be compensated for mirror reflection (using MATLAB \texttt{fliplr()}) to undo the inversion of image caused by splitting of incoming light along two orthogonal axes. In the leftmost image of \figurename\ref{fig:design_and_calib_of_mls_beam_splitter_setup}, the rotation of one of the camera in our setup is evident from the flipped sequence of the ethernet and power cables.

\subsection{Registration of Cross Polarized Images}
The images captured by both the cameras are automatically registered if their principal axes intersect. However, such a setup is not possible to achieve as the connection adapter used to screw in the camera lens to the beam splitter mount introduces offset between the principal axes of both cameras. Hence, to align the two cross polarized images, we need to compute a 2D homography matrix $H$ that transforms one of the cross polarized image in order to align it with the other image. Note that this alignment step is quite different from the registration in Ma \etal \cite{ma2008framework} setup required to compensate for the motion of the subject during the capture process. This alignment is performed to cancel the offset in the principal axis of the two camera so that the images captured by the two camera are in perfect registration.
\begin{figure}[htbp]
  \centering
  \includegraphics{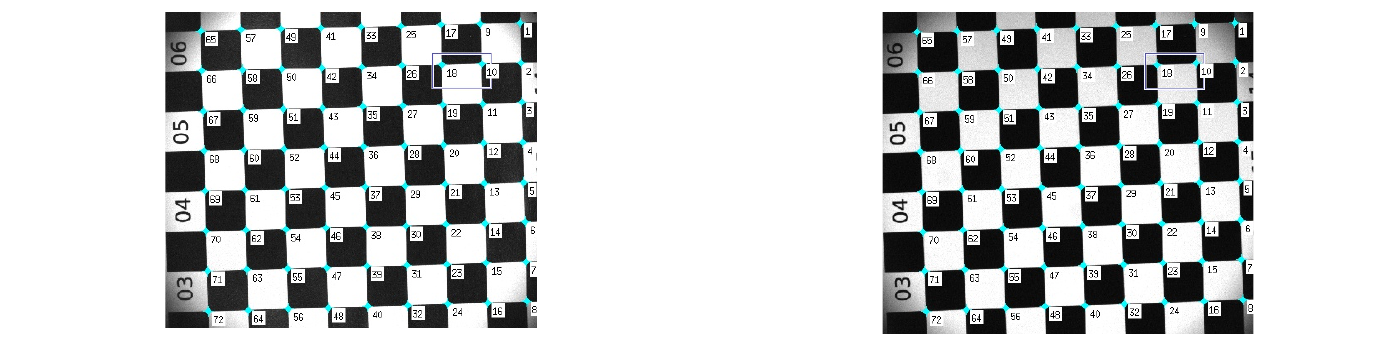}
  \caption{Manually selected correspondence points in CAM0 (left) and CAM1 (right) image for normalized DLT algorithm}
  \label{fig:design_and_calib_of_mls_cam0_cam1_checkerboard_sel_pts}
\end{figure}

We obtain an initial estimate of the 2D homography matrix $H$ using normalized Direct Linear Transform (DLT) \cite[p109]{hartley2004multiple}. The automatic corner detection tool in \cite{bouguet2004camera} is used to detect $65$ corner points in a checkerboard image captured using the two camera and beam splitter setup. These points are corrected manually, as shown in \figurename\ref{fig:design_and_calib_of_mls_cam0_cam1_checkerboard_sel_pts}, for subpixel accuracy and then supplied to the DLT algorithm as ($n=65$) initial 2D to 2D point correspondances. Using the initial estimate of $H$ from DLT, we determine the Maximum Likelihood Estimate (MLE) of $\hat{H}$ that minimizes the Samson's error \cite[p114]{hartley2004multiple}. This homography matrix is applied to all the images captured by CAM0 so that the transformed CAM0 images are in alignment with the CAM1 image.

Lens distortion is another effect that can lead to misalignment of images by the two cameras as described in \cite{zhang1996epipolar}. However, the effect is likely to be very small and therefore we ignore the contribution of lens distortion.

\subsection{Results of Diffuse and Specular Separation}
The CAM0 gradient images ($I_{0}$) can be transformed using 2D homography $\hat{H}$ to obtain images $\hat{I_{0}}$ that are aligned with the CAM1 gradient images ($I_{1}$). Diffuse and specular only images can now easily be obtained using
\[
  I_{S} = \hat{I_{0}} - I_{1} \qquad \textrm{and} \qquad I_{D} = 2I_{1} .
\]
\begin{figure}[htbp]
  \centering
  \includegraphics{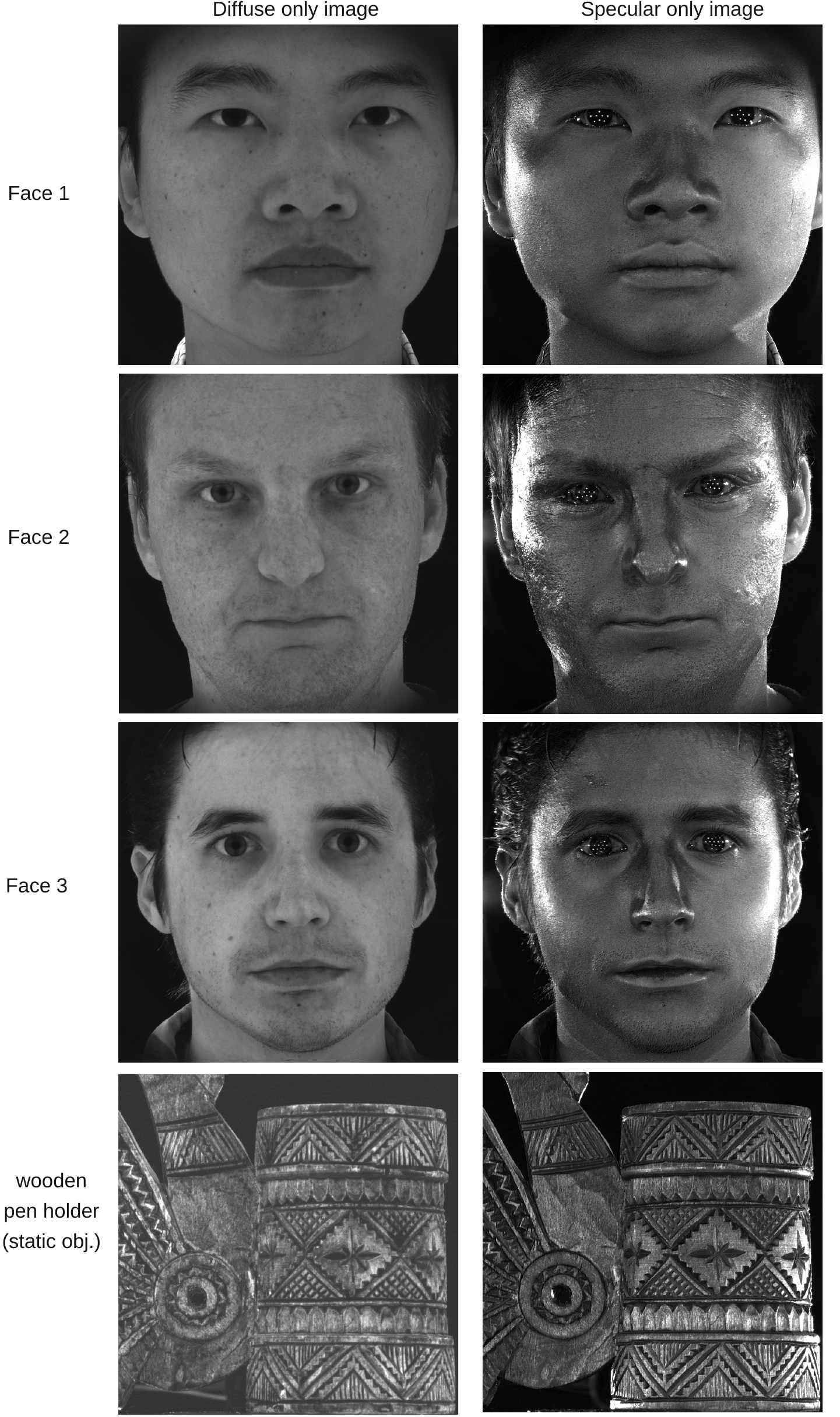}
  \caption{Result of diffuse and specular reflectance separation}
  \label{fig:design_and_calib_of_mls_diffuse_and_spec_only_images}
\end{figure}
The result of specular and diffuse separation for constant spherical illumination of faces and a static object is shown in \figurename\ref{fig:design_and_calib_of_mls_diffuse_and_spec_only_images}.

\section{Extending the Basic Light Stage Design for Multispectral Capture}
\label{ch:design_and_calib_of_mls:sec:extending_ls_for_multispectral_cap}
Most real world objects (human skin, fruits, etc..) are made up of multiple layers having different absorption and reflectance properties. These properties are very useful in Computer Graphics and Computer Vision research because they reveal the reflectance and absorption characteristics of underlying layers. The visible white light consists of radiation in the visible range ($380$nm to $720$nm) and they have differential penetration depth in human skin with the red band ($620-750$ nm) going deepest. Hence, a mutlispectral image set ---
reflectance recorded at sparse set of bands in the visible range --- contains the reflectance information from different layers of an object. For example: the Multispectral images of a fruit can reveal the properties of its inner layers. This information can be used to study the quality of a fruit \cite{kleynen2005development} \cite{unay2005multispectral}. Similarly,
parametric skin reflectance models like \cite{donner2008layered} rely on skin images captured at a set of narrow bands in the visible range. This information helps in creating a model of light interaction in different layers of the skin.
\begin{figure}[htbp]
  \centering
  \includegraphics{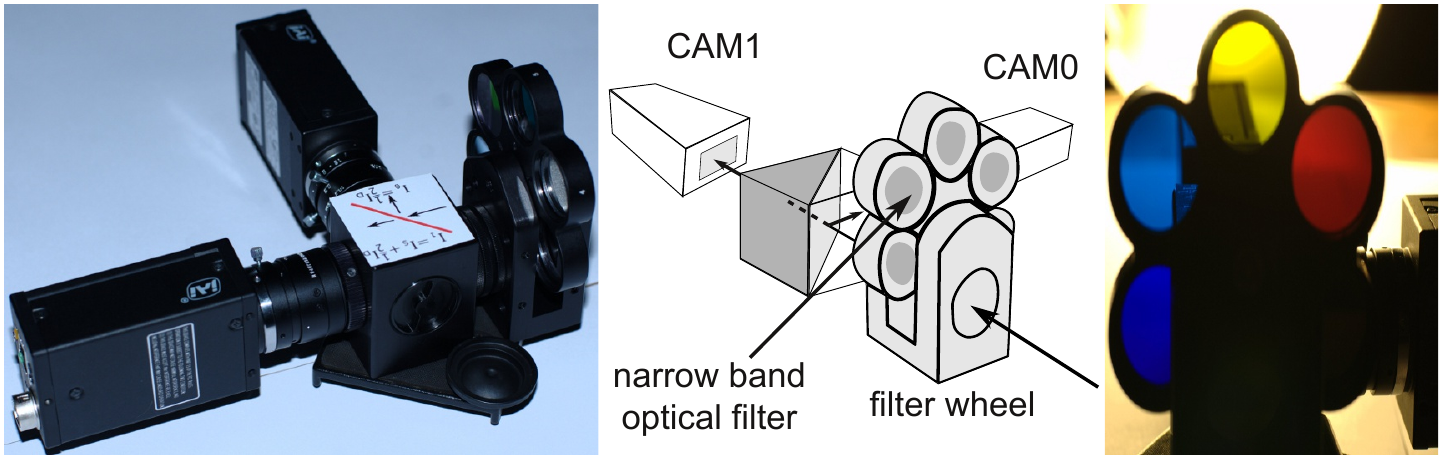}
  \caption{(left and centre) Optical filter wheel attached to existing beam splitter setup for Multispectral acquisition. (right) Optical filters centred at different wavelength of the visible range in a filter wheel}
  \label{fig:design_and_calib_of_mls_multispectral_cap_setup}
\end{figure}

We have made a modification in the light stage capture device proposed by Ma \etal \cite{ma2007rapid} to allow capture of multispectral images. These images are captured in spherical illumination environment and are very useful for analysis of multiple layered objects because they do not contain shading information. We placed BrightLine single bandpass optical filters in front of our existing beam splitter setup (discussed in \ref{ch:design_and_calib_of_mls:sec:simul_cap_cross_pol_img}) to allow simultaneous capture of cross polarized multispectral images. Six filters are mounted to a filter wheel (as shown in \figurename\ref{fig:design_and_calib_of_mls_multispectral_cap_setup}) which snaps into preset position when the filter wheel is rotated. Manual rotation of the filter wheel increases the total capture time to $\sim12$ seconds. Stepper motor driven filter wheel\footnote{\url{http://www.thorlabs.de/NewGroupPage9.cfm?ObjectGroup_ID=988}} (access time of $\sim650$ms) can be used to reduce the capture time to $\sim2$sec. However, these electronic filter wheels are very expensive and hence we choose to use the manual filter wheel.

\begin{figure}[htbp]
  \centering
  \includegraphics{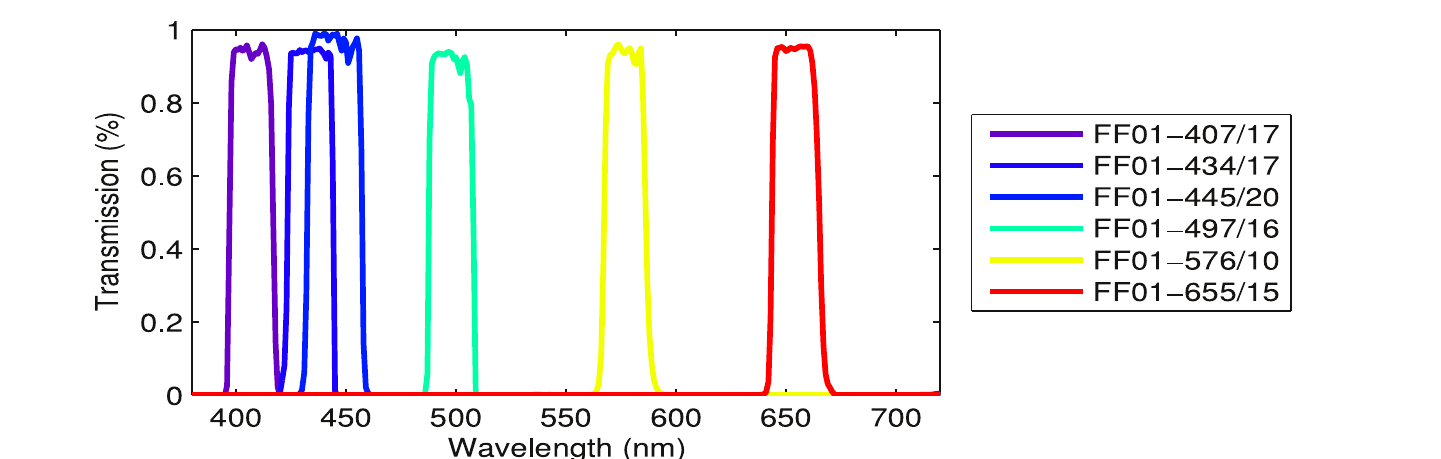}
  \caption{Transmission spectrum of single-bandpass optical filters used for the Multispectral light stage}
  \label{fig:design_and_calib_of_mls_optical_filter_transmission_plot}
\end{figure}

\begin{figure}[htbp]
  \centering
  \includegraphics{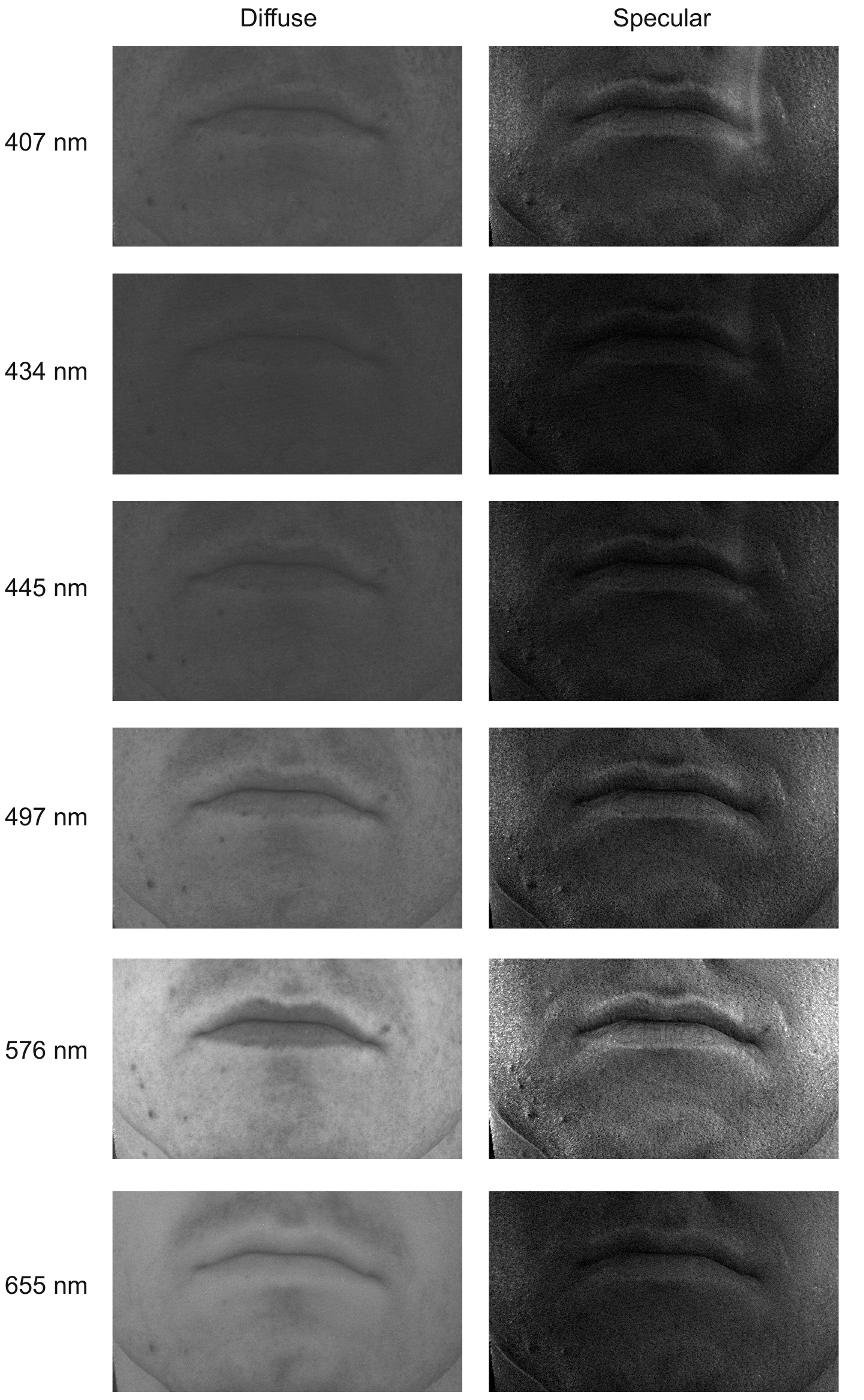}
  \caption{Multispectral diffuse and specular image set for a face region}
  \label{fig:design_and_calib_of_mls_multispectral_image_set_will_lips}
\end{figure}

The specification and transmission spectrum of 6 single bandpass filters used in our Multispectral light stage is given in \tablename\ref{tbl:design_and_calib_of_mls_optical_filter_transmission_plot} and \figurename\ref{fig:design_and_calib_of_mls_optical_filter_transmission_plot} respectively. These optical filters are polarization preserving: a property critical for simultaneous acquisition of cross polarized multispectral images using our existing beam splitter setup. The Semrock Brightline \textregistered \;single bandpass filters\footnote{\url{http://www.semrock.com/Catalog/Category.aspx?CategoryID=27}} have the polarization preserving property with more than $90\%$ transmission in the pass band. Such high transmission property is crucial for our setup because we lose more than $80\%$ of the light source emission due to linear polarizer and the beam splitter. The center wavelength of these filters were chosen to sample the most significant points in the chromophore absorption curve cobtained by \cite[Fig.7]{donner2008layered}. Hence, this filter set targets the subsurface reflectance characteristics of human skin. The result of multispectral capture is shown in \figurename\ref{fig:design_and_calib_of_mls_multispectral_image_set_will_lips}.

The diffuse Multispectral images clearly show the effect of absorption by in multiple skin layers. For example, the 655 nm diffuse image in \figurename\ref{fig:design_and_calib_of_mls_multispectral_image_set_will_lips} do not show freckle and moles which are visible in other bands of the multispectral diffuse image set. It is the subject of future work to use these multispectral images to recover parameters of a skin reflectance model like \cite{donner2008layered}.
\begin{table}[htbp]
  \centering
  \caption{Single bandpass filters used for the Multispectral light stage}
  \begin{tabular}{ |l |c |c| c|}
  \hline
    Filter & Center Wavelength (nm) & Bandwidth (nm) & Average Transmission (\%) \\ \hline
    FF01-407/17-25 & $407$ & $17$ & $>90$\\
    FF01-434/17-25 & $434$ & $17$ & $>90$\\
    FF01-445/20-25 & $445$ & $20$ & $>93$\\
    FF01-497/16-25 & $497$ & $16$ & $>90$\\
    FF01-576/10-25 & $576$ & $10$ & $>90$\\
    FF01-655/15-25 & $655$ & $15$ & $>90$\\  \hline
  \end{tabular}
  \label{tbl:design_and_calib_of_mls_optical_filter_transmission_plot}
\end{table}

\chapter{Multispectral Light Stage Data Processing}
\label{ch:mls_data_processing}
Ma's original Light Stage \cite{ma2007rapid} allows capture of spherical gradient and constant illumination images in full visible spectrum. These images can be used to recover high resolution surface geometry using the spherical gradient photometric stereo technique. Addition of single bandpass optical filters to the existing capture device setup of the Light Stage allows capture of multispectral spherical gradient and constant illumination images at six narrow bands in the visible spectrum. These multispectral images capture the reflectance properties of multi-layered materials like human skin. Such multispectral reflectance maps can be used with parametric skin reflectance models like \cite{donner2008layered}.

In this chapter, we first discuss the theoretical background of the spherical gradient photometric stereo method of Ma \etal \cite{ma2007rapid}. This method assumes perfect registration (or alignment) of all the gradient images being used for computation of photometric normal. For a non-static object like a human face, it is not possible to remain still during capture of all the four (\cite{ma2007rapid}) or six (\cite{wilson2010temporal}) gradient images. To correct for motion during the capture process, we discuss the ``Joint Photometric Alignment'' method proposed by Wilson \etal \cite{wilson2010temporal}. Using our modified radiance equations, we explore a Quadratic Programming (QP) based normal correction algorithm for surface geometry recovered using spherical gradient photometric stereo. Finally, based on our analysis of modified radiance equations, we propose a method to compute photometric normals using minimal four image set consisting of $(X,Y,Z, \{\bar{X}, \bar{Y}, \bar{Z}\})$. We also show that the proposed method has the improved robustness property of \cite{wilson2010temporal} and reduced data capture requirement benifit of \cite{ma2007rapid}.

\section{Spherical Gradient Photometric Stereo using Diffuse Images}
\begin{figure}[htbp]
  \centering
  \includegraphics{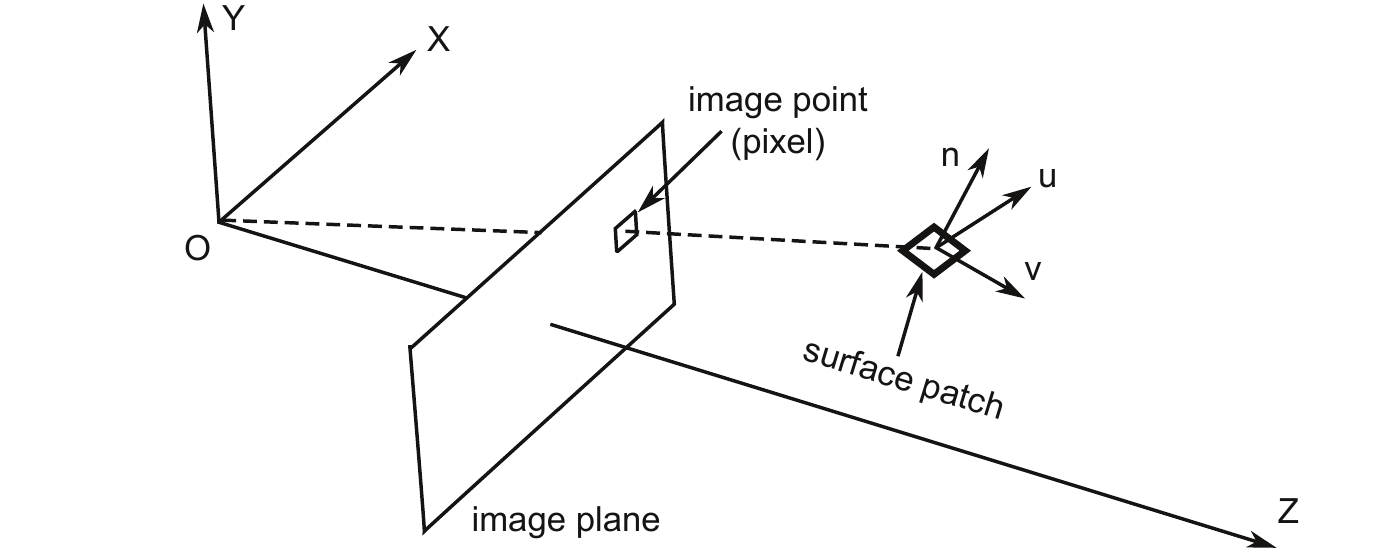}
  \caption{Global $(X,Y,Z)$ and local coordinate $(u,v,n)$ frame for diffuse images}
  \label{fig:mls_data_processing_diffuse_local_and_global_coordinate_frame}
\end{figure}

For every image point (i.e. pixel) in a diffuse image, we define a local coordinate frame $[\vec{u},\vec{v},\vec{n}]$ such that the $\vec{n}$ axis aligns with the surface normal of the surface patch corresponding to that image point as shown in \figurename~\ref{fig:mls_data_processing_diffuse_local_and_global_coordinate_frame}. We use primed symbols, i.e. $\omega'$, to represent vectors in a local coordinate frame.
The axes of local coordinate frame $[\vec{u},\vec{v},\vec{n}]$ can be defined in terms of the global coordinate frame $[O,X,Y,Z]$ as
\begin{eqnarray*}
  \vec{u} & = & (u_{x}i+u_{y}j+u_{z}k), \\
  \vec{v} & = & (v_{x}i+v_{y}j+v_{z}k), \\
  \vec{n} & = & (n_{x}i+n_{y}j+n_{z}k).
\end{eqnarray*}
Let us also define  $\omega'=(\omega'_{u},\omega'_{v},\omega'_{n})$ as the spherical direction in local coordinates such that the corresponding global coordinates are given by
\begin{eqnarray*}
  \omega = (\omega'_{u}u_{x}+\omega'_{v}v_{x}+\omega'_{n}n_{x})i+(\omega'_{u}u_{y}+\omega'_{v}v_{y}+\omega'_{n}n_{y})j+(\omega'_{u}u_{z}+\omega'_{v}v_{z}+\omega'_{n}n_{z})k .
\end{eqnarray*}
For any Lambertian surface, the value of radiance under spherical illumination is given by
\begin{equation}
  r = \int_{\Omega}P(\omega)R(\omega,n)\textrm{d}\omega = \int_{\Omega}P(\omega')R(\omega',[0,0,1])\textrm{d}\omega',
  \label{eq:mls_data_processing_lambertian_radiance_eq}
\end{equation}
where $P(\omega)$ and $P(\omega')$ represent the intensity of light incoming from direction $\omega$ (global coordinate) and $\omega'$ (local coordinate) respectively and $R(\omega,n)$ is the Lambertian Bidirectional Reflectance Distribution Function (BRDF). Recall that both $\omega$ and $\omega'$ represent the same physical direction but in different coordinate frames. In (\ref{eq:mls_data_processing_lambertian_radiance_eq}) we have substituted $[0,0,1]$ vector for the normal because the $n$ axis of local coordinate frame aligns with the surface normal.

In the case of X-gradient spherical illumination, the intensity of light incident from direction $\omega'\in\Omega$ is proportional to the X-component of $\omega\in\Omega$ (the corresponding incident direction represented in global coordinate).
\begin{equation}
P(\omega')=P_{x}(\omega)=(\omega'_{u}u_{x}+\omega'_{v}v_{x}+\omega'_{n}n_{x})\in[-1,1]
\label{eq:mls_data_processing_ideal_gradient_intensity_eq}
\end{equation}
As it is not possible to emit light with negative value of intensity, we cannot realize an X-gradient illumination with $P(\omega')\in[-1,1]$.
Hence, we rescale as follows:
\begin{equation}
P(\omega')=\frac{P_{x}(\omega)+1}{2}=\frac{(\omega'_{u}u_{x}+\omega'_{v}v_{x}+\omega'_{n}n_{x})+1}{2}\in[0,1].
\label{eq:mls_data_processing_practical_gradient_intensity_eq}
\end{equation}

\subsection{Radiance Equation for Gradient Illumination}
Substituting (\ref{eq:mls_data_processing_practical_gradient_intensity_eq}) in (\ref{eq:mls_data_processing_lambertian_radiance_eq}), we can write the radiance equation for X-gradient illumination as:
\begin{equation}
  r_{x} = \int_{\Omega}\left(\frac{\omega'_{u}u_{x}+\omega'_{v}v_{x}+\omega'_{n}n_{x}+1}{2}\right)R(\omega',[0,0,1])\textrm{d}\omega'. \\
\label{eq:mls_data_processing_x_gradient_radiance_eq1}
\end{equation}
Both Ma \etal \cite{ma2007rapid} and Wilson \etal \cite{wilson2010temporal} assumed that the surface is convex and that the diffuse reflectance is symmetric about the surface normal. Hence, the integral over the hemisphere along $u_x$ and $v_x$ axes becomes $0$ and the gradient radiance simplifies to
\begin{eqnarray}
r_{x} & = & \frac{1}{2}\left\{n_{x}\int_{\Omega}\omega'_{n}R(\omega',[0,0,1])\textrm{d}\omega'+\int_{\Omega}R(\omega',[0,0,1])\textrm{d}\omega'\right\} \nonumber, \\
      & = & \frac{\pi\rho_{D}}{2}\left\{n_{x}\int_{0}^{1}\omega'_{n}\omega'_{n}\textrm{d}\omega'+\int_{0}^{1}\omega'_{n}\textrm{d}\omega'\right\} \nonumber, \\
r_{x} & = & \frac{\pi\rho_{D}}{2}\left\{\frac{1}{3}n_{x}+\frac{1}{2}\right\},
\label{eq:mls_data_processing_x_gradient_radiance_eq2}
\end{eqnarray}
where $\rho_{D}$ is the diffuse albedo. In a similar way, we can arrive at the following equations for diffuse radiance in Y and Z gradient illumination:
\begin{eqnarray}
r_{y} & = & \frac{\pi\rho_{D}}{2}\left\{\frac{1}{3}n_{y}+\frac{1}{2}\right\},
\label{eq:mls_data_processing_y_gradient_radiance_eq1}
\end{eqnarray}
\begin{eqnarray}
r_{z} & = & \frac{\pi\rho_{D}}{2}\left\{\frac{1}{3}n_{z}+\frac{1}{2}\right\}.
\label{eq:mls_data_processing_z_gradient_radiance_eq1}
\end{eqnarray}

\subsection{Radiance Equation for Constant Illumination}
For ideal constant spherical illumination, the intensity of light incident from all the possible spherical directions is a constant, i.e.
\begin{eqnarray*}
  P(\omega')=1 \qquad \textrm{for all} \quad \omega'\in\Omega.
\end{eqnarray*}
Thus, the expression for radiance under constant illumination becomes:
\begin{eqnarray}
r_{c} & = & \int_{\Omega}R(\omega',[0,0,1])\textrm{d}\omega'\nonumber \\
      & = & \int_{-1}^{1}(\pi\rho_{D})max(0,\omega'.[0,0,1])\textrm{d}\omega'\nonumber \\
      & = & \pi\rho_{D}\int_{0}^{1}\omega'_{n}\textrm{d}\omega'\nonumber, \\
r_{c} & = & \frac{\pi\rho_{D}}{2}.
\label{eq:mls_data_processing_constant_illum_radiance}
\end{eqnarray}
It is evident from (\ref{eq:mls_data_processing_constant_illum_radiance}) that the constant illumination image is used to recover the diffuse albedo $\rho_{D}$.

\subsection{Surface Normal Estimation}
Ma \etal \cite{ma2007rapid} used the ratio of gradient images to the constant illumination image to recover high resolution surface geometry of the surface visible in the gradient images. Hence, the ratio of (\ref{eq:mls_data_processing_x_gradient_radiance_eq2}),(\ref{eq:mls_data_processing_y_gradient_radiance_eq1}), (\ref{eq:mls_data_processing_z_gradient_radiance_eq1}) to (\ref{eq:mls_data_processing_constant_illum_radiance}) results in:
\begin{eqnarray}
n_x & = & \frac{1}{N_{d}}\left(\frac{r_x}{r_c} - \frac{1}{2}\right) \nonumber, \\
\label{eq:ppalsi_eq9}
n_y & = & \frac{1}{N_{d}}\left(\frac{r_y}{r_c} - \frac{1}{2}\right) \nonumber, \\
\label{eq:ppalsi_eq10}
n_z & = & \frac{1}{N_{d}}\left(\frac{r_z}{r_c} - \frac{1}{2}\right) \nonumber,
\label{eq:mls_data_processing_ma2007_ratio_eq}
\end{eqnarray}
where, $N_{d}$ is a normalizing constant given by
\begin{equation}
N_{d} = \sqrt{\left(\frac{r_x}{r_c} - \frac{1}{2}\right)^2+\left(\frac{r_y}{r_c} - \frac{1}{2}\right)^2+\left(\frac{r_z}{r_c} - \frac{1}{2}\right)^2}.
\label{eq:mls_data_processing_ma2007_norm_const_eq}
\end{equation}

\section{Spherical Gradient Photometric Stereo using Specular Only Images}
\begin{figure}[htbp]
  \centering
  \includegraphics{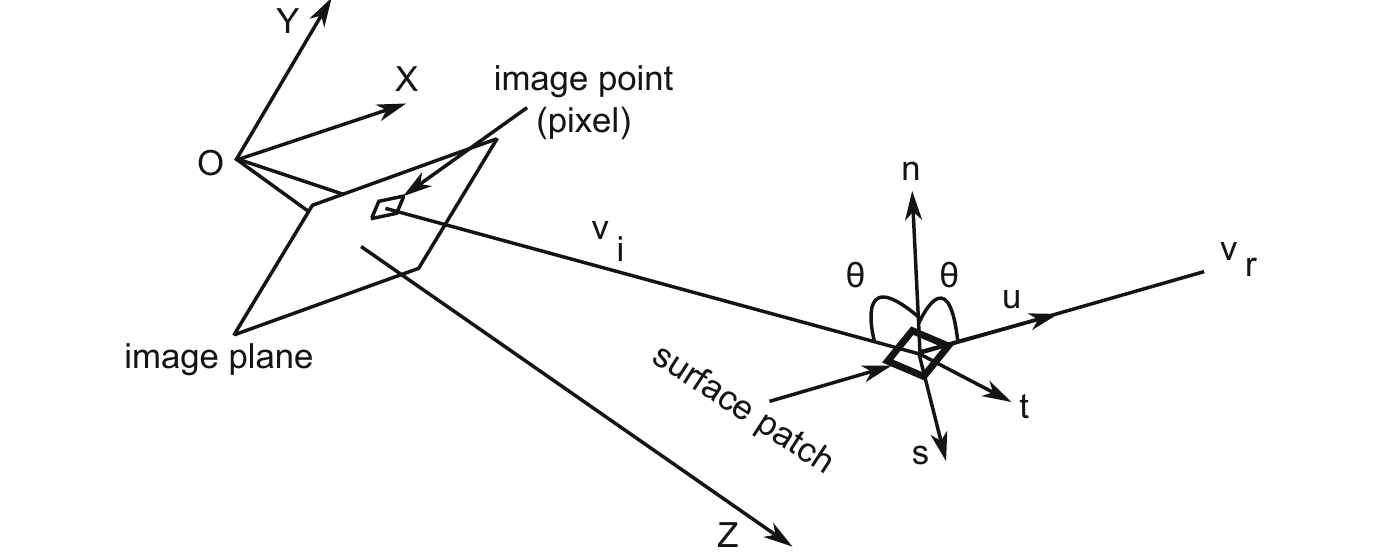}
  \caption{Global $(X,Y,Z)$ and local coordinate $(u,v,n)$ frame for specular images}
  \label{fig:mls_data_processing_specular_local_and_global_coordinate_frame}
\end{figure}

For the analysis of specular radiance, let us define global $[X,Y,Z]$ and local coordinate $[\vec{s},\vec{t},\vec{u}]$ frames as shown in \figurename\ref{fig:mls_data_processing_specular_local_and_global_coordinate_frame}. $v_{i}$ represents the view vector and $v_{r}$ is the reflected direction of view vector which is obtained by $180^{\circ}$ rotation of $v_{i}$ around the surface normal $n$. The local coordinate frame $[\vec{s},\vec{t},\vec{u}]$ for every image point (i.e. pixel) in a specular image is defined such that $u$ axis aligns with the reflected direction of view vector $v_{r}$ and the orthogonal axes $\vec{s},\vec{t}$ are orthogonal to $u$ axis.
The axes of local coordinate frame $[\vec{s},\vec{t},\vec{u}]$ can be defined in terms of the global coordinate frame $[O,X,Y,Z]$ as:
\begin{eqnarray*}
  \vec{s} & = & (s_{x}i+s_{y}j+s_{z}k), \\
  \vec{t} & = & (t_{x}i+t_{y}j+t_{z}k), \\
  \vec{u} & = & (u_{x}i+u_{y}j+u_{z}k).
\end{eqnarray*}
Let us also define $\omega'=(\omega'_{s},\omega'_{t},\omega'_{u})$ as the spherical direction in local coordinate frame such that the corresponding global coordinate frame direction is given by
\begin{eqnarray*}
  \omega = (\omega'_{s}s_{x}+\omega'_{t}t_{x}+\omega'_{u}u_{x})i+(\omega'_{s}s_{y}+\omega'_{t}t_{y}+\omega'_{u}u_{y})j+(\omega'_{s}s_{z}+\omega'_{t}t_{z}+\omega'_{u}u_{z})k.
\end{eqnarray*}
The value of specular radiance under spherical illumination is given by
\begin{equation}
  r = \int_{\Omega}P(\omega)R(\omega,v_{i},n)\textrm{d}\omega = \int_{\Omega}P(\omega')R(\omega',v'_{i},n')\textrm{d}\omega',
  \label{eq:mls_data_processing_specular_radiance_eq}
\end{equation}
where, $P(\omega)$ and $P(\omega')$ represent the intensity of light incoming from direction $\omega$ (global coordinate) and $\omega'$ (local coordinate) respectively and $R(\omega,v_{i},n)$ is the specular Bidirectional Reflectance Distribution Function (BRDF). Recall that both $\omega$ and $\omega'$ represent the same physical direction but in different coordinate frames. The specular BRDF can be expressed as
\begin{equation}
  R(\omega,v_{i},n) = S(r,v_{i},n)\Psi(\omega,n),
  \label{eq:mls_data_processing_specular_brdf_eq}
\end{equation}
where, $r=2(n.\omega)n-\omega$ is the perfect specular reflected direction, $S$ is the specular reflectance lobe which is non-zero around a small solid angle around $r$ and $\Psi$ is the foreshortening factor.

\subsection{Radiance Equation for Gradient Illumination}
In the case of X-gradient spherical illumination, the intensity of light incident from direction $\omega'\in\Omega$ is proportional to the X-component of $\omega\in\Omega$ (the corresponding incident direction represented in global coordinate), i.e.
\begin{equation}
P(\omega')=P_{x}(\omega)=(\omega'_{s}s_{x}+\omega'_{t}t_{x}+\omega'_{u}u_{x})\in[-1,1].
\label{eq:mls_data_processing_stu_frame_ideal_gradient_intensity_eq}
\end{equation}

As it is not possible to emit light with negative value of intensity, we cannot realize an X-gradient illumination with $P(\omega')\in[-1,1]$.
Hence, we rescale as follows:
\begin{equation}
P(\omega')=\frac{P_{x}(\omega)+1}{2}=\frac{(\omega'_{s}s_{x}+\omega'_{t}t_{x}+\omega'_{u}u_{x})+1}{2}\in[0,1].
\label{eq:mls_data_processing_stu_frame_practical_gradient_intensity_eq}
\end{equation}

Substituting (\ref{eq:mls_data_processing_stu_frame_practical_gradient_intensity_eq}) and (\ref{eq:mls_data_processing_specular_brdf_eq}) in (\ref{eq:mls_data_processing_specular_radiance_eq}), we can write the radiance equation for X-gradient illumination as:
\begin{equation}
  r_{x} = \int_{\Omega}\left(\frac{\omega'_{s}s_{x}+\omega'_{t}t_{x}+\omega'_{u}u_{x}+1}{2}\right)S(r',v'_{i},n')\Psi(\omega',n')\textrm{d}\omega'
\label{eq:mls_data_processing_specular_x_gradient_radiance_eq1}
\end{equation}
where the superscript $'$ is added to represent coordinates in local coordinate frame. Ma \etal \cite[p27]{ma2008framework} assumed the foreshortening factor $\Psi$ to be constant (say $c_{f}$). This assumption is not valid for glossy reflections (i.e. the specular lobe $S$ is non-zero around a large solid angle around $r$) and surface patches that lie at grazing angle with respect to the viewer (i.e. $v_{i} \sim 90^{\circ}$).

The ideal specular lobe $S$ is symmetric along the $u$ axis and hence the first two terms involving $s_{x}$ and $t_{x}$ in (\ref{eq:mls_data_processing_specular_x_gradient_radiance_eq1}) become zero resulting in
\begin{equation}
  r_{x} = \frac{1}{2} \left\{ u_{x}\int_{\Omega}\omega'_{u}S(r',v'_{i},n')\Psi(\omega',n')\textrm{d}\omega' + \int_{\Omega}S(r',v'_{i},n')\Psi(\omega',n')\textrm{d}\omega' \right\}.
\label{eq:mls_data_processing_specular_x_gradient_radiance_eq2}
\end{equation}

In a similar way, we can arrive at the following equations for specular radiance in Y and Z gradient illumination:
\begin{equation}
  r_{y} = \frac{1}{2} \left\{ u_{y}\int_{\Omega}\omega'_{u}S(r',v'_{i},n')\Psi(\omega',n')\textrm{d}\omega' + \int_{\Omega}S(r',v'_{i},n')\Psi(\omega',n')\textrm{d}\omega' \right\},
\label{eq:mls_data_processing_specular_y_gradient_radiance_eq2}
\end{equation}
\begin{equation}
  r_{z} = \frac{1}{2} \left\{ u_{z}\int_{\Omega}\omega'_{u}S(r',v'_{i},n')\Psi(\omega',n')\textrm{d}\omega' + \int_{\Omega}S(r',v'_{i},n')\Psi(\omega',n')\textrm{d}\omega' \right\}.
\label{eq:mls_data_processing_specular_z_gradient_radiance_eq2}
\end{equation}

\subsection{Radiance Equation for Constant Illumination}
For ideal constant spherical illumination, the intensity of light incident from all the possible spherical directions is a constant, i.e.
\begin{eqnarray*}
  P(\omega')=1 \qquad \textrm{for all} \quad \omega'\in\Omega.
\end{eqnarray*}
Thus, the expression for specular radiance under constant illumination becomes
\begin{eqnarray}
r_{c} & = & \int_{\Omega}S(r',v'_{i},n')\Psi(\omega',n')\textrm{d}\omega'.
\label{eq:mls_data_processing_specular_constant_illum_radiance}
\end{eqnarray}

\subsection{Surface Normal Estimation}
It is evident from (\ref{eq:mls_data_processing_specular_x_gradient_radiance_eq2}), (\ref{eq:mls_data_processing_specular_y_gradient_radiance_eq2}), (\ref{eq:mls_data_processing_specular_z_gradient_radiance_eq2}) and (\ref{eq:mls_data_processing_specular_constant_illum_radiance}) that we can recover the reflected direction of the view vector $v_{r} = u(u_{x}, u_{y}, u_{z})$ by subtracting the constant illumination specular image from the X, Y and Z gradient illumination specular image followed by normalization. Mathematically,
\begin{eqnarray}
u_{x} & = & \frac{1}{N_{s}} (r_{x} - \frac{1}{2}r_{c}), \nonumber \\
u_{y} & = & \frac{1}{N_{s}} (r_{y} - \frac{1}{2}r_{c}), \nonumber \\
u_{z} & = & \frac{1}{N_{s}} (r_{z} - \frac{1}{2}r_{c}),
\label{eq:mls_data_processing_refl_dir_of_view_vec_eq}
\end{eqnarray}
where, $N_{s}=\sqrt{(r_{x} - \frac{1}{2}r_{c})^{2} + (r_{y} - \frac{1}{2}r_{c})^{2} + (r_{z} - \frac{1}{2}r_{c})^{2}}$ is a normalizing constant. The half way vector between view vector $v_{i} = [0\quad 0\quad -1]^{T}$ and the reflected direction of view vector corresponds to the surface normal and is given by
\begin{eqnarray}
\vec{n} = \frac{1}{\bar{N}}(v_{r} + v_{i}),
\label{eq:mls_data_processing_specular_normal_eq}
\end{eqnarray}
where, $\bar{N}$ is a normalizing constant and $v_{r}=u$.

The specular normal map is able to capture fine surface details because unlike diffuse radiance --- which is a subsurface phenomena --- specular reflection is a surface phenomena. A fine structure due to white paint on the nose tip of a white cement statue is revealed in the specular normal map of \figurename\ref{fig:mls_data_processing_distribution_of_diffuse_spec_norm_const_value} (top right) while the diffuse normal map (top left) does not capture this fine detail. The constant foreshortening factor $F$ assumption of \cite[p27]{ma2008framework} breaks down at grazing angle as revealed by large amount of noise in the boundary of face and both sides of the nose bridge in the specular normal map of \figurename\ref{fig:mls_data_processing_distribution_of_diffuse_spec_norm_const_value} (top right).

\section{Analysis of the Normalizing Constant Value - $N_{d}$ and $N_{s}$}
\begin{figure}[htbp]
  \centering
  \includegraphics{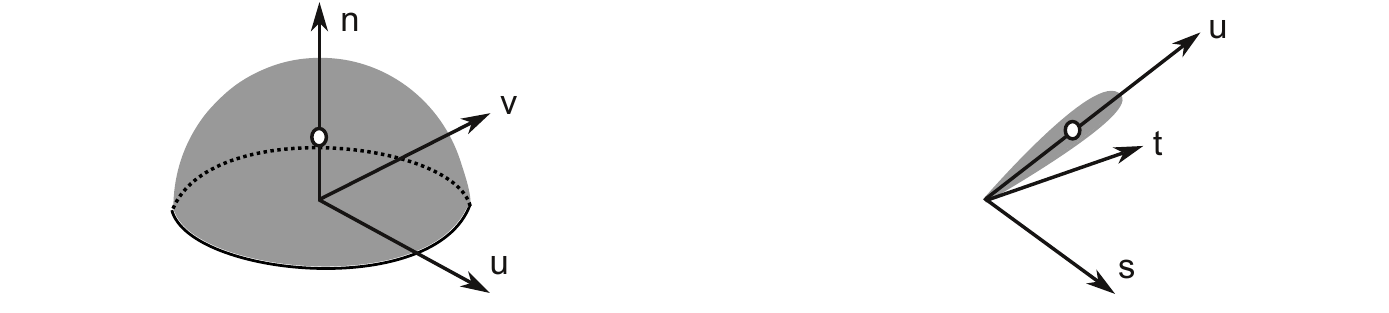}
  \caption{Centroid (depicted by small white circle) of diffuse and specular reflectance lobe}
  \label{fig:mls_data_processing_diffuse_specular_lobe_centroid}
\end{figure}
The vector along the direction of diffuse and specular lobe centroids (as shown by white circle in \figurename\ref{fig:mls_data_processing_diffuse_specular_lobe_centroid}) can be converted to a unit vector by normalization: operation in which a vector is divided by its magnitude (also called normalization constant or $\ell^{2}$-norm ) to obtain a unit vector in its direction. The expression for normalization of diffuse and specular centroid is given by:
\begin{equation*}
N_{d} = \frac{1}{r_{c}^{d}}\sqrt{\left(r_{x}^{d} - \frac{1}{2}r_{c}^{d}\right)^{2} + \left(r_{y}^{d} - \frac{1}{2}r_{c}^{d}\right)^{2} + \left(r_{z}^{d} - \frac{1}{2}r_{c}^{d}\right)^{2}},
\end{equation*}
\begin{equation*}
N_{s}=\sqrt{\left(r_{x}^{s} - \frac{1}{2}r_{c}^{s}\right)^{2} + \left(r_{y}^{s} - \frac{1}{2}r_{c}^{s}\right)^{2} + \left(r_{z}^{s} - \frac{1}{2}r_{c}^{s}\right)^{2}}.
\end{equation*}
Note that these expressions for $N_{d}$ and $N_{s}$ are same as (\ref{eq:mls_data_processing_ma2007_norm_const_eq}) and (\ref{eq:mls_data_processing_refl_dir_of_view_vec_eq}) with only the superscript $^{d}$ and $^{s}$ added to depict the diffuse and specular radiance values. The unit vector along diffuse and specular centroid direction correspond to the surface normal and reflected direction of the view vector respectively.

\begin{figure}[htbp]
  \centering
  \includegraphics{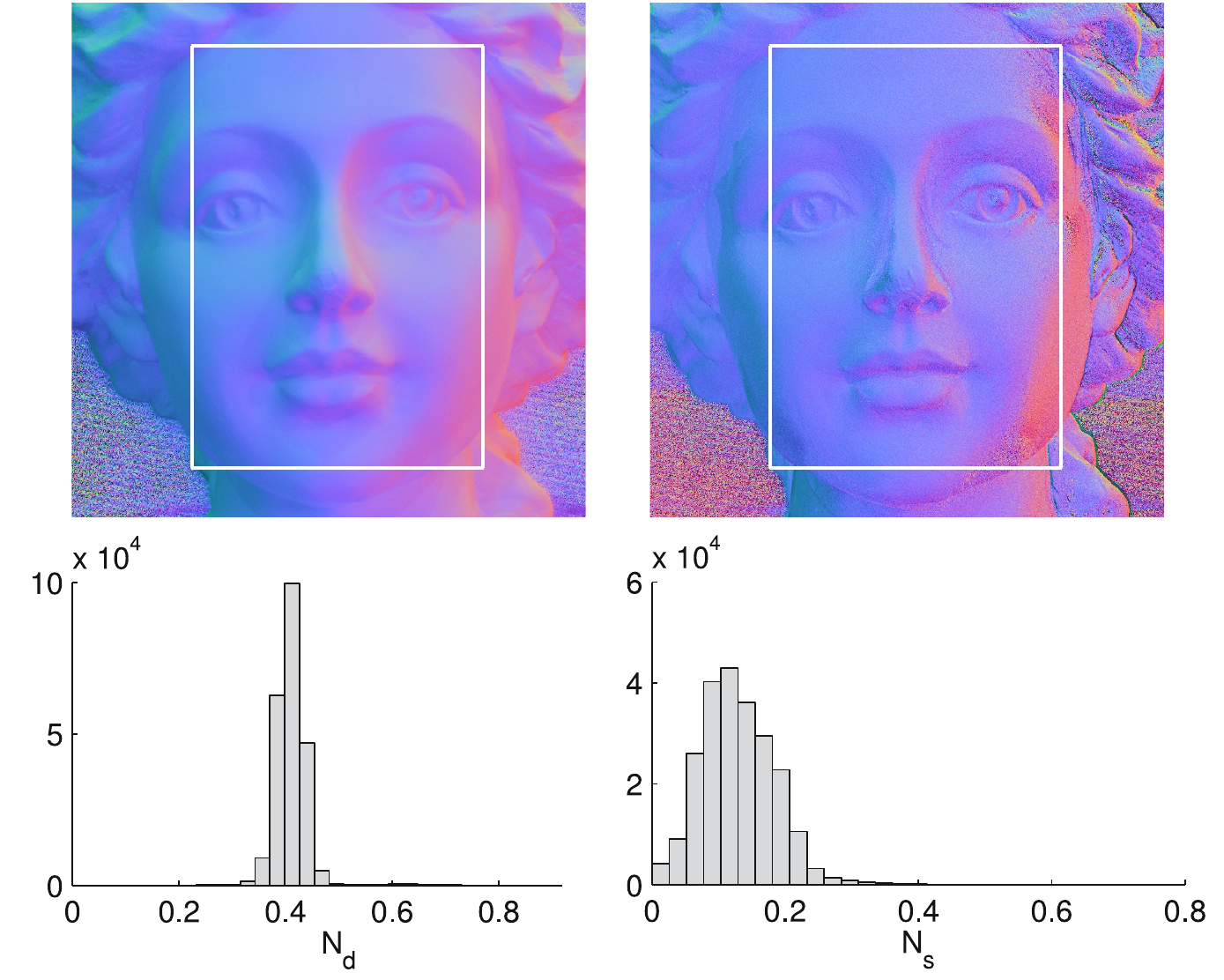}
  \caption{Distribution of diffuse $N_{d}$ (bottom left) and specular $N_{s}$ (bottom right) normalizing constant value for a region (depicted with white rectangle) in the diffuse (top left) and specular(top right) normal map of a white cement plaster statue.}
  \label{fig:mls_data_processing_distribution_of_diffuse_spec_norm_const_value}
\end{figure}

The normalizing constant values $N_{d}$ and $N_{s}$ are proportional to the size of diffuse and specular reflectance lobes respectively. Hence, for a typical diffuse surface, we would expect $N_{s}<N_{d}$ to hold true. The distribution of $N_{d}$ and $N_{s}$ for a white cement plaster statue (diffuse object) shown in \figurename\ref{fig:mls_data_processing_distribution_of_diffuse_spec_norm_const_value} supports this hypothesis.
The distribution of diffuse normalizing constant $N_{d}$ value reveals another interesting fact. Most of the $N_{d}$ values are clustered in $(0.37, 0.41)$ region. We need to investigate deeper into the nature of ideal diffuse reflectance centroid to be able to explain this behaviour. Let us consider an ideal diffuse reflectance lobe symmetric along $n$ axis and stretching $k$ units along this axis of the local coordinate frame $[u,v,n]$ as shown in \figurename\ref{fig:mls_data_processing_diffuse_specular_lobe_centroid}. Such diffuse reflectance lobe can be defined by the solution for $f(u,v,n) = 0$, where the function $f(u,v,n)$ is defined as:
\[
  f(u,v,n) = u^{2} + v^{2} + n^{2} - k^{2} \qquad n \in[0,k].
\]
As the diffuse reflectance function is symmetric about $n$ axis, its centroid is given by
\[
  (u_{0},v_{0},n_{0}) = \left(0, 0, \frac{\int nf(u,v,n)\textrm{dn}}{\int f(u,v,n)\textrm{dn}}\right) = \left(0,0, \frac{3(k^{2}-2k)}{4(k^{2}-3)}\right).
\]
So, for a unit diffuse reflectance lobe (i.e. $k=1$), the centroid lies at $(0,0,0.375)$ and the values of diffuse normalizing constant is $N_{d}=0.375$. Hence, the distribution of $N_{d}$ in \figurename\ref{fig:mls_data_processing_distribution_of_diffuse_spec_norm_const_value} confirms that the white cement plaster surface has reflecting properties that are close to an ideal diffuse surface.

For a wide variety of real world surfaces, the diffuse and specular reflectance lobes get distorted due to inter-reflection, ambient occlusion and coarse approximation of spherical illumination due to light discretization. This causes the diffuse and specular lobe centroids to shift away from its ideal position on the surface normal and reflected direction of the view vector respectively. For these reasons, the normalizing constant values cannot be used to infer the nature of diffuse and specular reflectance lobes. For example, it is possible for a completely distorted diffuse lobe to acquire centroid value of a unit diffuse lobes (i.e. $(0,0,0.375)$). Hence, although the value of normalizing constant is a good measure of the reflecting properties of a surface, it cannot be used to quantify the nature of distortion in the reflectance lobes. Analysis of the normalizing constant values provides a good insight into the basis of spherical gradient photometric stereo technique and its limitations.

\section{Quadratic Programming based Normal Correction}
\label{ch:mls_data_processing:sec:quad_prog_for_norm_improvement}

Quality of surface geometry recovered using spherical gradient photometric stereo \cite{ma2007rapid} is affected by the extent to which the following assumptions are satisfied:
\begin{enumerate}
  \item no shadowing of light sources i.e. object is convex
  \item no inter-reflection i.e. light incident on a surface patch is solely due to light source and not because of reflections from nearby surface patches
  \item light sources closely approximate a continuous illumination environment i.e. effect of \textit{``light discretization''}\footnote{a term used by Ma \etal \cite{ma2007rapid} to refer to coarse approximation of spherical illumination caused by LEDs attached to discrete positions on a twice subdivided icosahedron. It is important to realise that the term ``light discretisation'' does not imply that intensity of light sources is discrete.} is minimal
\end{enumerate}
\begin{figure}[htbp]
  \centering
  \includegraphics{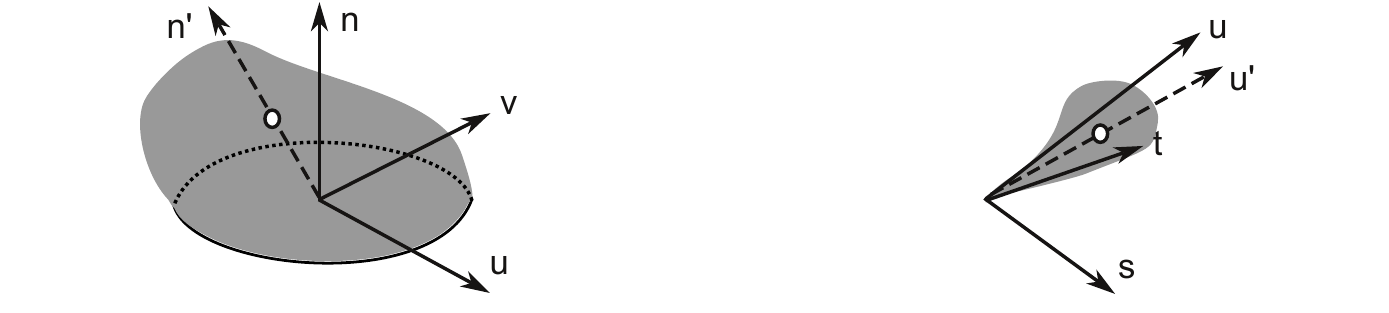}
  \caption{Deformed diffuse (left) and specular (right) lobes due to inter-reflection, ambient occlusion and coarse approximation of spherical illumination}
  \label{fig:mls_data_processing_distorted_diffuse_specular_lobe_centroid}
\end{figure}

In this section, we will introduce new parameters to the original Ma \etal \cite{ma2007rapid} radiance equations in order to quantify the extent of violation of these three assumptions. These modified radiance equations not only helps uncover the limitations of Ma \etal method, but also provide insight into possible modifications of this technique in order to improve the quality of recovered surface geometry. Using these modified equation, we show why the quality of normal estimated by Ma \etal method degrades with deformed diffuse lobe. We also propose a Quadratic Programming (QP) based normal correction technique to compensate for the effects of deformed diffuse lobes and hence improve the quality of recovered surface normals. Finally, based on analysis of our modified radiance equations, we propose a minimal image sets method for spherical gradient photometric stereo which has the improved robustness property of Wilson \etal \cite{wilson2010temporal} and reduced data capture requirement benifit of Ma \etal \cite{ma2007rapid}.

Here, we present an analysis of diffuse lobes deformation only because similar approach can be used to analyse the effects of deformed specular lobes.

\subsection{Modified Radiance Equations for Gradient Illumination}
\label{ch:mls_data_processing:sec:quad_prog_for_norm_improvement:subsec:modified_radiance_eq_grad_illum}
\begin{figure}[htbp]
  \centering
  \includegraphics{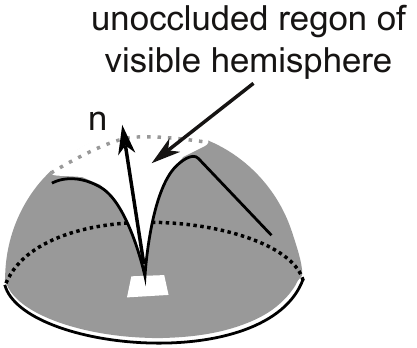}
  \caption{Ambient occlusion in concave surfaces}
  \label{fig:mls_data_processing_ambient_occlusion_illustration}
\end{figure}
Ambient occlusion limits the portion of hemisphere visible to a surface patch as shown in \figurename\ref{fig:mls_data_processing_ambient_occlusion_illustration}. Hence, to quantify the effect of ambient occlusion at an arbitrary surface patch $p$ in spherical illumination images, we introduce the following binary visibility function:
\begin{equation}
  V_{p,\omega'} = \left\{
    \begin{array}{rl}
      1 & \textrm{if direction } \; \omega' \; \textrm{ is unoccluded}, \\
      0 & \textrm{otherwise}.
    \end{array} \right.
\end{equation}
We can now rewrite the X gradient radiance equation of (\ref{eq:mls_data_processing_x_gradient_radiance_eq1}) as:
\begin{equation}
  r_{x} = \int_{\Omega}V_{p,\omega'}\left(\frac{\omega'_{u}u_{x}+\omega'_{v}v_{x}+\omega'_{n}n_{x}+1}{2}\right)R(\omega',[0,0,1])\textrm{d}\omega'.
\label{eq:mls_data_processing_x_gradient_radiance_modified_eq1}
\end{equation}
Inter-reflections and coarse approximation of spherical illumination deforms the diffuse reflectance lobe along the $u_{x}$ and $v_{x}$ axes. Hence, contribution of the integrals along $u_{x}$ and $v_{s}$ axes cannot be ignored in the case of deformed diffuse lobe. In other words, the diffuse reflectance lobe is no more symmetric along the $n_{x}$ axes. Ma \etal \cite{ma2007rapid} and Wilson \etal \cite{wilson2010temporal} assumed a diffuse reflectance lobe symmetric along the $n_{x}$ axis and therefore they were able to ignore the contribution of these integrals in their analysis.

We do not ignore the effect of asymmetry in diffuse reflectance lobe. However, as it is not possible to evaluate the integrals along $u_{x}$ and $v_{x}$, we quantify the extent of distortion in diffuse reflectance lobe using a single scalar $\delta'_{x}$ (distortion coefficient). This parameter scales the diffuse albedo $\pi\rho_{D}$ to quantify the contribution of integrals along $u_{x}$ and $v_{x}$ axes in (\ref{eq:mls_data_processing_x_gradient_radiance_modified_eq1}). In other words, we make the simplifying assumption that overall deformation in the diffuse reflectance lobe for a gradient illumination environment can be quantified using a single parameter  $\delta'_{x}$ (distortion coefficient).

Adding this parameter to (\ref{eq:mls_data_processing_x_gradient_radiance_modified_eq1}) gives:
\begin{equation}
    r_{x} = \frac{\pi\rho_{D}}{2}\left\{ \delta'_{x}+n_{x}\int_{0}^{1}\omega'_{n}V_{p,\omega'}\omega'_{n}\textrm{d}\omega'+\int_{0}^{1}V_{p,\omega'}\omega'_{n}\textrm{d}\omega' \right\}
\label{eq:mls_data_processing_x_gradient_radiance_modified_eq2}
\end{equation}
where,
\[
\delta'_{x}(\pi\rho_{D})=u_{x}\int_{\Omega}\omega'_{u}V_{p,\omega'}R(\omega',[0,0,1])\textrm{d}\omega'+v_{x}\int_{\Omega}\omega'_{v}V_{p,\omega'}R(\omega',[0,0,1])\textrm{d}\omega'
\]
To simplify the evaluation of (\ref{eq:mls_data_processing_x_gradient_radiance_modified_eq2}), we first consider the ideal case value for the visibility function i.e. when complete hemisphere is visible. In this ideal case, $V_{p,\omega'}=1$ for all $\omega'\in\Omega$, (\ref{eq:mls_data_processing_x_gradient_radiance_modified_eq2}) simplifies to:
\begin{eqnarray}
    r_{x} = \frac{\pi\rho_{D}}{2}\left\{\delta'_{x}+n_{x}\int_{0}^{1}\omega'_{n}\omega'_{n}\textrm{d}\omega'+\int_{0}^{1}\omega'_{n}\textrm{d}\omega'\right\}
      & = & \frac{\pi\rho_{D}}{2}\left\{ \delta'_{x}+\frac{1}{3}n_{x}+\frac{1}{2}\right\}.
\label{eq:mls_data_processing_x_gradient_radiance_modified_eq3}
\end{eqnarray}
For real world objects, the ideal case value of visibility function is not valid i.e. $\exists\omega'\in\Omega:V_{p,\omega'}\neq1$. This implies that the actual value of two integrals in (\ref{eq:mls_data_processing_x_gradient_radiance_modified_eq2}) will be less than their ideal case values i.e.
\[
  \int_{0}^{1}\omega'_{n}V_{p,\omega'}\omega'_{n}\textrm{d}\omega'<\frac{1}{3} \qquad \textrm{and} \qquad
  \int_{0}^{1}V_{p,\omega'}\omega'_{n}\textrm{d}\omega'<\frac{1}{2}.
\]
To quantify the overall effect of shadowing, we define the ambient occlusion term $V_{p}\in[0,1]$ such that: $V_{p}=1$ when complete hemisphere is visible and $V_{p}=0$ for completely occluded hemisphere. The intermediate values $0<V_{p}<1$ apply to partial occlusion. Substituting this visibility parameter in (\ref{eq:mls_data_processing_x_gradient_radiance_modified_eq3}), we obtain the following expression for radiance from real world surfaces under X gradient illumination:
\begin{eqnarray}
r_{x} & = & \frac{\pi\rho_{D}V_{p}}{2}\left\{ \delta_{x}+\frac{1}{3}n_{x}+\frac{1}{2}\right\},
\label{eq:mls_data_processing_x_gradient_radiance_modified_final_eq}
\end{eqnarray}
where, $\delta'_{x} = V_{p}\delta_{x}$. In a similar way, we can obtain the expression for radiance in Y and Z gradients illumination
\begin{eqnarray}
r_{y} & = & \frac{\pi\rho_{D}V_{p}}{2}\left\{ \delta_{y}+\frac{1}{3}n_{y}+\frac{1}{2}\right\},
\label{eq:mls_data_processing_y_gradient_radiance_modified_final_eq}
\end{eqnarray}
\begin{eqnarray}
r_{z} & = & \frac{\pi\rho_{D}V_{p}}{2}\left\{ \delta_{z}+\frac{1}{3}n_{z}+\frac{1}{2}\right\}.
\label{eq:mls_data_processing_z_gradient_radiance_modified_final_eq}
\end{eqnarray}

\subsection{Modified Radiance Equation for Constant Illumination}
\label{ch:mls_data_processing:sec:quad_prog_for_norm_improvement:subsec:modified_radiance_eq_const_illum}
For constant spherical illumination, Ma \etal \cite{ma2007rapid} assumed the intensity of light incident from all possible spherical directions to be unity, i.e.
\[
P(\omega')=1 \qquad \textnormal{for} \; \omega'\in\Omega.
\]
This is true for ideal case spherical illumination. However, this assumption ignores: a) light source attenuation effects, which is equivalent to assuming all points on the object lie exactly at the centre of the light stage; b) contribution of inter-reflection and shadowing which can increase or decrease the intensity of light incident from a particular spherical direction. In (\ref{ch:mls_data_processing:sec:quad_prog_for_norm_improvement:subsec:modified_radiance_eq_grad_illum}), we introduced the binary visibility function $V_{p,\omega'}$ which models whether a spherical direction $\omega'$ is visible at any surface patch $p$. For a surface patch, the intensity light incident from a direction $\omega'$ is dependent on the binary visibility function defined for that surface patch. Therefore, we can now define $P(\omega')$ as:
\begin{equation}
  P(\omega') = \left\{
    \begin{array}{rl}
      c_{p,\omega'} & \textnormal{if } \; V_{p,\omega'} = 1 \\
      0 & \textnormal{otherwise},
    \end{array} \right.
\label{eq:constant_illumination_intensity_function_ideal_eq}
\end{equation}
where, $c_{p,\omega'}$ models the angular deviation of intensity under constant illumination for a surface patch $p$. As it is not possible to evaluate radiance integral using this defination of $P(\omega')$, we make the simplifying assumption that the intensity of incident light is unity when a spherical incident direction is visible from a surface patch. In other words, we also use the unit incident intensity assumption of Ma \etal but only for visible spherical directions. Mathematically,
\begin{equation}
  P(\omega') = \left\{
    \begin{array}{rl}
      1 & \textnormal{if} \; V_{p,\omega'} = 1 \\
      0 & \textnormal{otherwise}.
    \end{array} \right.
\label{eq:constant_illumination_intensity_function_simplified_eq}
\end{equation}
This simplifying assumption ignores the light source attenuation effects and contribution of inter-reflection and only includes the contribution of shadowing effects under constant spherical illumination. Using this simplifying assumption, the expression for radiance under constant spherical illumination becomes:
\begin{eqnarray}
r_{c} & = & \int_{\Omega}P(\omega')V_{p,\omega'}R(\omega',[0,0,1])\textnormal{d}\omega'\nonumber \\
  & = & \int_{-1}^{1}V_{p,\omega'}(\pi\rho_{D})max(0,\omega'.[0,0,1])\textnormal{d}\omega'\nonumber \\
  & = & \pi\rho_{D}\int_{0}^{1}V_{p,\omega'}\omega'_{n}\textnormal{d}\omega' =  \frac{\pi\rho_{D}V_{p}}{2}.
\label{eq:mls_data_processing_constant_radiance_modified_final_eq}
\end{eqnarray}

\subsection{Quality of Surface Normal Estimated Using Original Spherical Gradient Photometric Stereo Method}
\label{ch:mls_data_processing:sec:quad_prog_for_norm_improvement:subsec:ma2007_nm_quality_assessment}
\begin{figure}[htbp]
  \centering
  \includegraphics{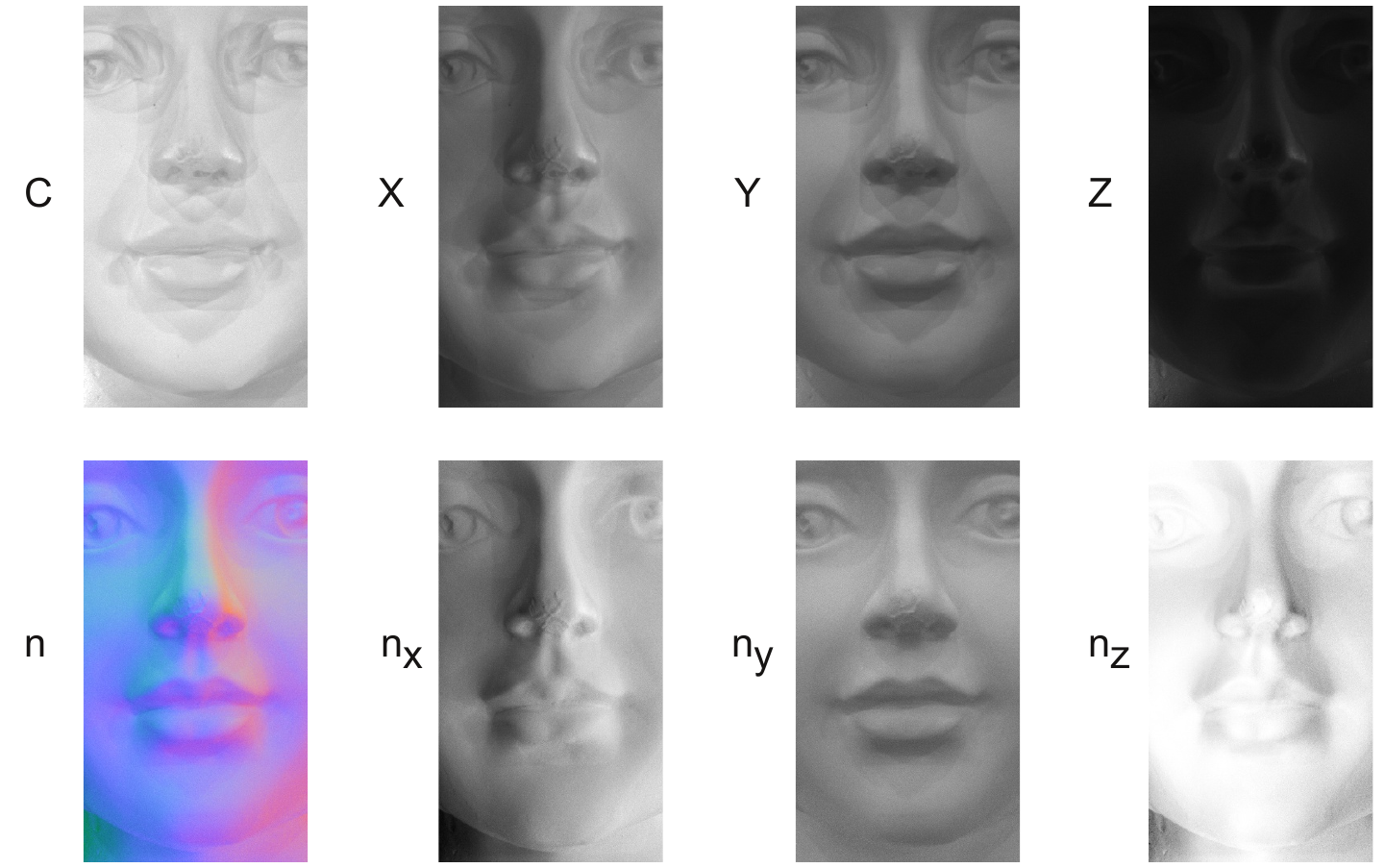}
  \caption{Shadows clearly visible in the constant $C$ and gradient $(X,Y,Z)$ images (top row) of a white cement plaster statue. The normal map (bottom leftmost - normal components mapped to R,G,B) and the X,Y,Z normal components depicted as grayscale image (bottom right three) do not show the effect of shadows.}
  \label{fig:mls_data_processing_ma2007_amb_occl_invar_illus}
\end{figure}

The expression for computing photometric normal using Ma \etal \cite{ma2007rapid} method is:
\begin{equation}
 \vec{n} = \frac{(\frac{r_{x}}{r_{c}} - \frac{1}{2}, \frac{r_{y}}{r_{c}} - \frac{1}{2},\frac{r_{z}}{r_{c}} - \frac{1}{2})}{||(\frac{r_{x}}{r_{c}} - \frac{1}{2}, \frac{r_{y}}{r_{c}} - \frac{1}{2},\frac{r_{z}}{r_{c}} - \frac{1}{2})||}
 \label{eq:ma2007_eq1}
\end{equation}
Now, we use our modified radiance equations to represent the surface normal computed using the above method:
\begin{equation}
 N_{\{x,y,z\}} = \frac{r_{\{x,y,z\}}}{r_{c}} - \frac{1}{2} = \delta{\{x,y,z\}} + \frac{1}{3}n_{\{x,y,z\}},
 \label{eq:ma2007_eq2}
\end{equation}
where, $N_{\{x,y,z\}}$ is the unnormalized surface normal vector. It is evident from above expression that although the occlusion term ($V_{p}$) cancel in this ``ratio method'', the diffuse lobe distortion term $\delta_{\{x,y,z\}}$ does not cancel out. Therefore, we conclude that the quality of surface normals computed using Ma \etal method will degrade with deformation in diffuse lobe. It is important to understand that the cancellation of occlusion term ($V_{p}$) results from the following simplifying assumption used while evaluating the radiance equation for constant illumination: the intensity of incident light is unity when a spherical incident direction is visible from a surface patch i.e. $P(\omega') = 1$ if direction $\omega'$ is unoccluded.

The modified radiance equations of (\ref{eq:mls_data_processing_x_gradient_radiance_modified_final_eq}, \ref{eq:mls_data_processing_y_gradient_radiance_modified_final_eq}, \ref{eq:mls_data_processing_z_gradient_radiance_modified_final_eq} and \ref{eq:mls_data_processing_constant_radiance_modified_final_eq}) form an underdetermined system with $3$ equations and $6$ unknowns. In the next section, we explore the concept of complement image constraint in order to obtain additional constraints for this underdetermined system. This analysis will form the basis for our Quadratic Programming (QP) based normal correction.

\subsection{Modified Radiance Equations for Complement Gradient Illumination}
\label{ch:mls_data_processing:sec:quad_prog_for_norm_improvement:subsec:complement_constraints}
\begin{figure}[htbp]
  \centering
  \includegraphics{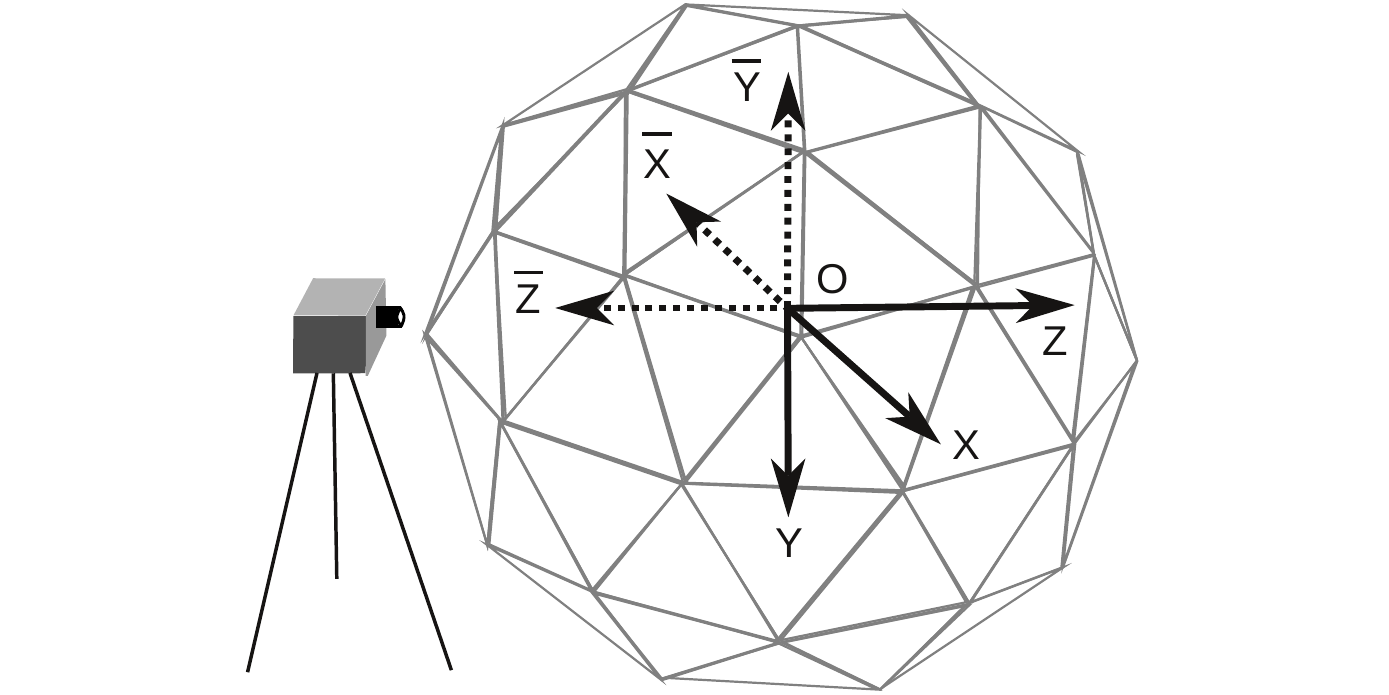}
  \caption{Complement coordinate frames in a Light Stage}
  \label{fig:mls_data_processing_complement_coord_frame_illus}
\end{figure}

Light Stage uses a reference coordinate frame $[O,X,Y,Z]$ to setup gradient illumination. In addition to this gradient condition, complementary coordinate frame $[O,\bar{X},\bar{Y},\bar{Z}]$ can also be used to setup complement gradient illumination environment. Here, $O$ is the center of light stage and $[\bar{X},\bar{Y},\bar{Z}]$ are the coordinate axes obtained by flipping $[X,Y,Z]$ as shown in \figurename\ref{fig:mls_data_processing_complement_coord_frame_illus}.
Although the true surface normal $n$ remains same in both coordinate frames, the distortion of diffuse reflectance lobe may not be identical. In other words, the distortion of diffuse lobe in complement gradient illumination contains some asymmetry with respect to the distortion observed in the gradient illumination. In the case of ideal spherical illumination and absence of ambient occlusion and inter-reflection, diffuse lobe distortion is symmetric in gradient and complement gradient illumination. To model the asymmetry present in complement gradient illumination, we represent the diffuse lobe distortion as the sum of symmetric distortion observed under gradient illumination ($\delta_{x}$) and an asymmetric component ($\delta_{\bar{x}}$). Therefore, we rewrite (\ref{eq:mls_data_processing_x_gradient_radiance_modified_final_eq}) in order to include asymmetry is diffuse lobe under the complement gradient illumination:
\begin{eqnarray}
r_{\bar{x}} & = & \frac{\pi\rho_{D}V_{p}}{2}\left\{ \delta_{x} + \delta_{\bar{x}}+\frac{1}{3}n_{\bar{x}}+\frac{1}{2}\right\},
\label{eq:mls_data_processing_x_complement_gradient_radiance_modified_eq1}
\end{eqnarray}
where, $\delta{\bar{x}}$ is a scalar quantifying the amount of asymmetry (with respect to distortion in gradient illumination) in the distortion of diffuse lobe. Flipping the reference coordinate frame does not alter the true surface normal and therefore (\ref{eq:mls_data_processing_x_complement_gradient_radiance_modified_eq1}) can be rewritten as:
\begin{eqnarray}
r_{\bar{x}} & = & \frac{\pi\rho_{D}V_{p}}{2}\left\{ \delta_{x} + \delta_{\bar{x}} - \frac{1}{3}n_{x} + \frac{1}{2}\right\},
\label{eq:mls_data_processing_x_complement_gradient_radiance_modified_final_eq}
\end{eqnarray}
In a similar way, we can obtain the following expression for radiance under Y and Z complement gradient illumination:
\begin{eqnarray}
r_{\bar{y}} & = & \frac{\pi\rho_{D}V_{p}}{2}\left\{ \delta_{y} + \delta_{\bar{y}} - \frac{1}{3}n_{y} + \frac{1}{2}\right\},
\label{eq:mls_data_processing_y_complement_gradient_radiance_modified_final_eq}
\end{eqnarray}
\begin{eqnarray}
r_{\bar{z}} & = & \frac{\pi\rho_{D}V_{p}}{2}\left\{ \delta_{z} + \delta_{\bar{z}} - \frac{1}{3}n_{z} + \frac{1}{2}\right\}.
\label{eq:mls_data_processing_z_complement_gradient_radiance_modified_final_eq}
\end{eqnarray}

The complement gradient images have also been used by Wilson \etal \cite{wilson2010temporal} to formulate an iterative algorithm (Joint Photometric Alignment) for estimation of optical flow of subject's motion during performance capture in a Light Stage.

\subsection{Correcting Recovered Surface Normals Using Quadratic Programming}
In this section, we will explore a Quadratic Programming (QP) based correction of surface geometry recovered using spherical gradient photometric stereo method. From the analysis so far (section \ref{ch:mls_data_processing:sec:quad_prog_for_norm_improvement:subsec:modified_radiance_eq_grad_illum}, \ref{ch:mls_data_processing:sec:quad_prog_for_norm_improvement:subsec:modified_radiance_eq_const_illum} and \ref{ch:mls_data_processing:sec:quad_prog_for_norm_improvement:subsec:complement_constraints}), we have the following expressions for radiance under gradient and complement gradient spherical illumination:
\begin{equation}
r_{\{x,y,z\}} = \delta_{\{x,y,z\}}+\frac{1}{3}n_{\{x,y,z\}}+\frac{1}{2},
\label{eq:mls_data_processing_quadratic_prog_nm_corr_eq_r_xyz}
\end{equation}
\begin{equation}
r_{\{\bar{x},\bar{y},\bar{z}\}} = \delta_{\{x,y,z\}} + \delta_{\{\bar{x},\bar{y},\bar{z}\}}-\frac{1}{3}n_{\{x,y,z\}}+\frac{1}{2}.
\label{eq:mls_data_processing_quadratic_prog_nm_corr_eq_r_xyz_bar}
\end{equation}

From (\ref{eq:mls_data_processing_quadratic_prog_nm_corr_eq_r_xyz}) and (\ref{eq:mls_data_processing_quadratic_prog_nm_corr_eq_r_xyz_bar}), we have $6$ linear equations resulting in an underdetermined system in $9$ unknowns ${{\bf x}=\left(\delta_{x},\delta_{y},\delta_{z},\delta_{\bar{x}},\delta_{\bar{y}},\delta_{\bar{z}},n_{x},n_{y},n_{z}\right)}$ which can be expressed in matrix form as:
\begin{equation}
  \textbf{Ax} = \textbf{b} \qquad \textnormal{where} \quad \textbf{A} \in \mathbf{R}^{6\times 9},
  \label{eq:mls_data_processing_quadratic_prog_linear_system_eq}
\end{equation}
\[
 A = 
 \begin{bmatrix}
	  1 & 0 & 0 & 0 & 0 & 0 & \frac{1}{3} &       0     &      0      \\
	  0 & 1 & 0 & 0 & 0 & 0 &      0      & \frac{1}{3} &      0      \\
	  0 & 0 & 1 & 0 & 0 & 0 &      0      &       0     & \frac{1}{3} \\
	  0 &  0 &  0 & -1 &  0 &  0 & \frac{2}{3} &       0     &      0      \\
	  0 &  0 &  0 &  0 & -1 &  0 &       0      & \frac{2}{3}&      0      \\
	  0 &  0 &  0 &  0 &  0 & -1 &       0      &       0     & \frac{2}{3} \\
 \end{bmatrix},
 \;\textnormal{and} \;
 b =
 \begin{bmatrix}
 \frac{r_{x}}{r_{c}} - \frac{1}{2} \\
 \frac{r_{y}}{r_{c}} - \frac{1}{2} \\
 \frac{r_{z}}{r_{c}} - \frac{1}{2} \\
 \frac{r_{x} - r_{\bar{x}}}{r_{c}} \\
 \frac{r_{y} - r_{\bar{y}}}{r_{c}} \\
 \frac{r_{z} - r_{\bar{z}}}{r_{c}}
 \end{bmatrix}.
\]

We apply a Quadratic Programming (QP) approach to perform correction to the surface normals computed using our minimal image sets method (or that computed using \cite{wilson2010temporal} or \cite{ma2007rapid}). We regularize the problem such that QP computes new surface normals and estimates for distortion coefficients such that our linear system is satisfied and the new surface normals are closest to an initial solution. For example, if we take the diffuse centroid $(n_{x}^{\textnormal{Wil}}, n_{y}^{\textnormal{Wil}}, n_{z}^{\textnormal{Wil}})$ estimated by the method of Wilson \etal \cite{wilson2010temporal} and define ${\bf x}_{0}=(0,0,0,0,0,0,n_{x}^{\textnormal{Wil}},n_{y}^{\textnormal{Wil}},n_{z}^{\textnormal{Wil}})$, then we can correct for deformed diffuse lobes by solving the following quadratic programming problem
\begin{equation}
  \textnormal{minimise} \; \|{\bf x}-{\bf x}_{0}\|^{2} \qquad \textnormal{subject to} \quad{\bf Ax}={\bf b}.
  \label{eq:mls_data_processing_quadratic_prog_problem_eq}
\end{equation}

\subsubsection{Results from QP based Surface Normal Correction}
First, we analyse the results of QP based surface normal correction for a simple static object (white cylinder) because the captured gradient images are perfectly aligned and its ground truth data is known. Moreover, the simple convex surface of this object allows us to evaluate the performance of our QP based normal correction strategy.
\begin{figure}[htbp]
  \centering
  \includegraphics{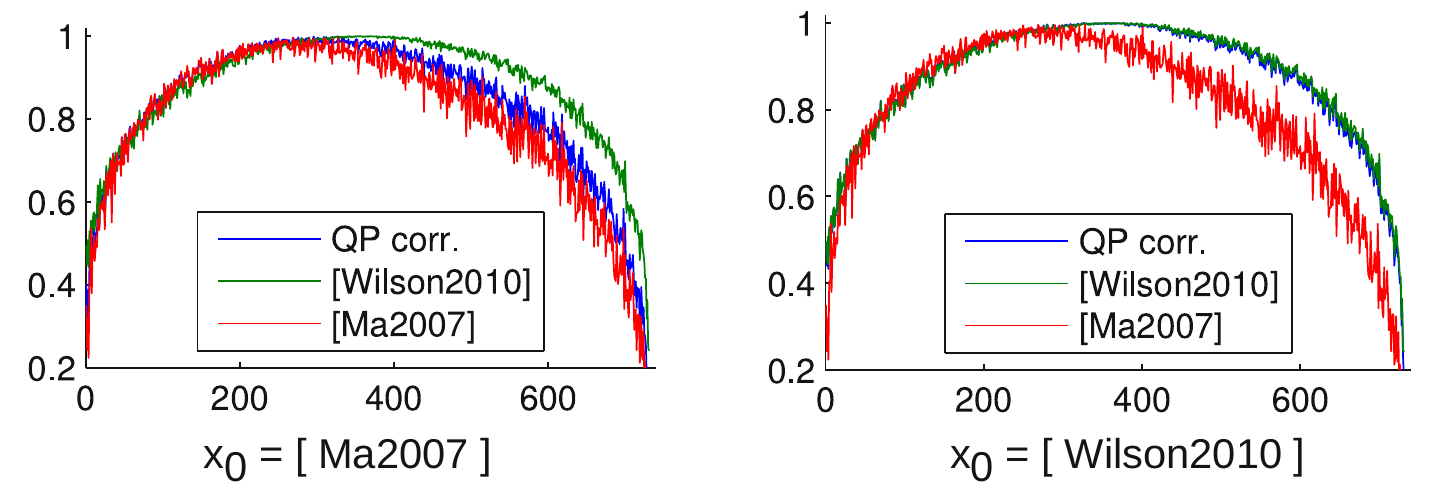}
  \caption{Result of QP based normal correction applied to surface normals of a white cylinder when initial solution is (left) ${\bf x}_{0}=(0,0,0,0,0,0,n_{x}^{\textnormal{Ma}},n_{y}^{\textnormal{Ma}},n_{z}^{\textnormal{Ma}})$ and (right) ${\bf x}_{0}=(0,0,0,0,0,0,n_{x}^{\textnormal{Wil}},n_{y}^{\textnormal{Wil}},n_{z}^{\textnormal{Wil}})$.}
  \label{fig:mls_data_processing_qp_results_for_white_cylinder_nz}
\end{figure}

When the initial solution to the QP based normal correction is the surface normal recovered using Ma \etal \cite{ma2007rapid} method, the corrected normals tend to move towards the true surface normal as shown in \figurename~\ref{fig:mls_data_processing_qp_results_for_white_cylinder_nz} (left). The correction algorithm cannot recover true normals because we seek the corrections that are closest to the initial solution given by the Ma \etal \cite{ma2007rapid} method. When the initial solution to the QP based normal correction is the surface normal recovered using Wilson \etal \cite{wilson2010temporal} method, the corrected normals tend to remain close to the initial solution as shown in \figurename~\ref{fig:mls_data_processing_qp_results_for_white_cylinder_nz} (right). This indicates that the surface normals recovered using Wilson \etal \cite{wilson2010temporal} are already close to the true surface normals.

\begin{figure}[htbp]
  \centering
  \includegraphics{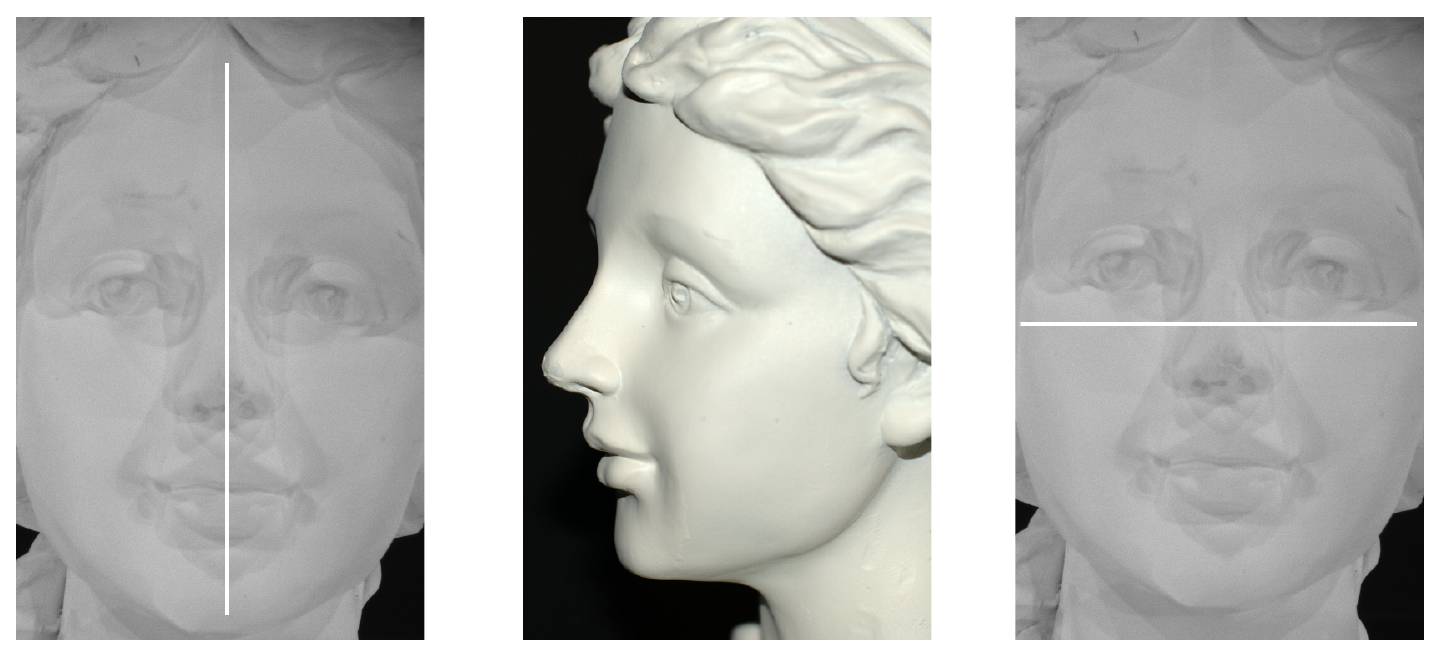}
  \caption{(left) $1$ pixel wide verticle region and (right) $1$ pixel wide horizontal region in gradient images of the face region of a statue selected for analysis of QP based normal correction. (center) Side view photograph of the statue's face region.}
  \label{fig:mls_data_processing_qp_lady_statue_side_view_photo}
\end{figure}

\begin{figure}[htbp]
  \centering
  \includegraphics{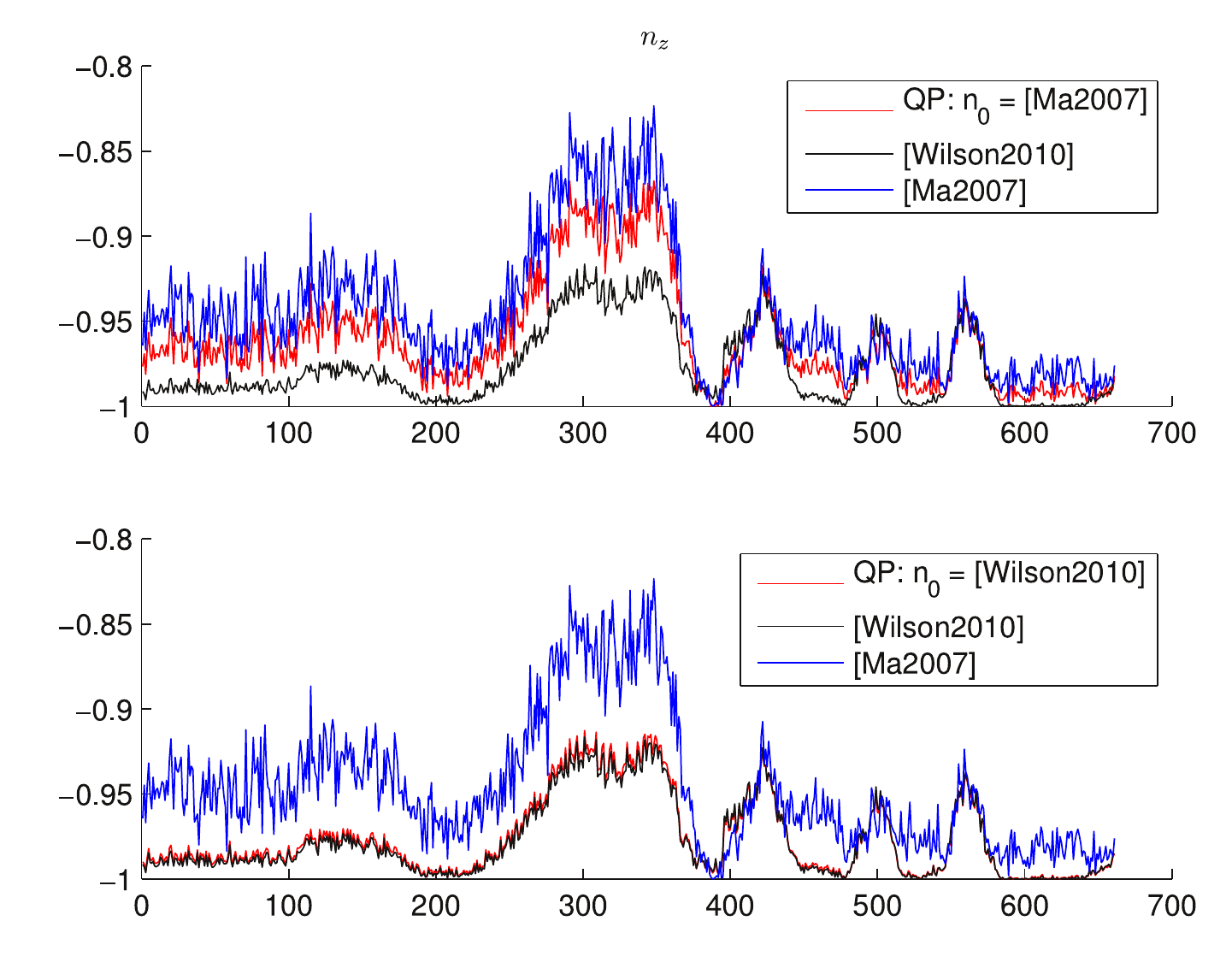}
  \caption{$n_{z}$ component of surface normals in 1 pixel wide selected vertical region (\figurename~\ref{fig:mls_data_processing_qp_lady_statue_side_view_photo} - left) obtained after applying QP based normal correction with initial estimate of surface normals from (top) Ma \etal method i.e $n_{0} = n^{[Ma2007]}$ and (bottom) Wilson \etal method i.e. $n_{0} = n^{[Wilson]}$.}
  \label{fig:mls_data_processing_vert_stripe_qp_result}
\end{figure}

\begin{figure}[htbp]
  \centering
  \includegraphics{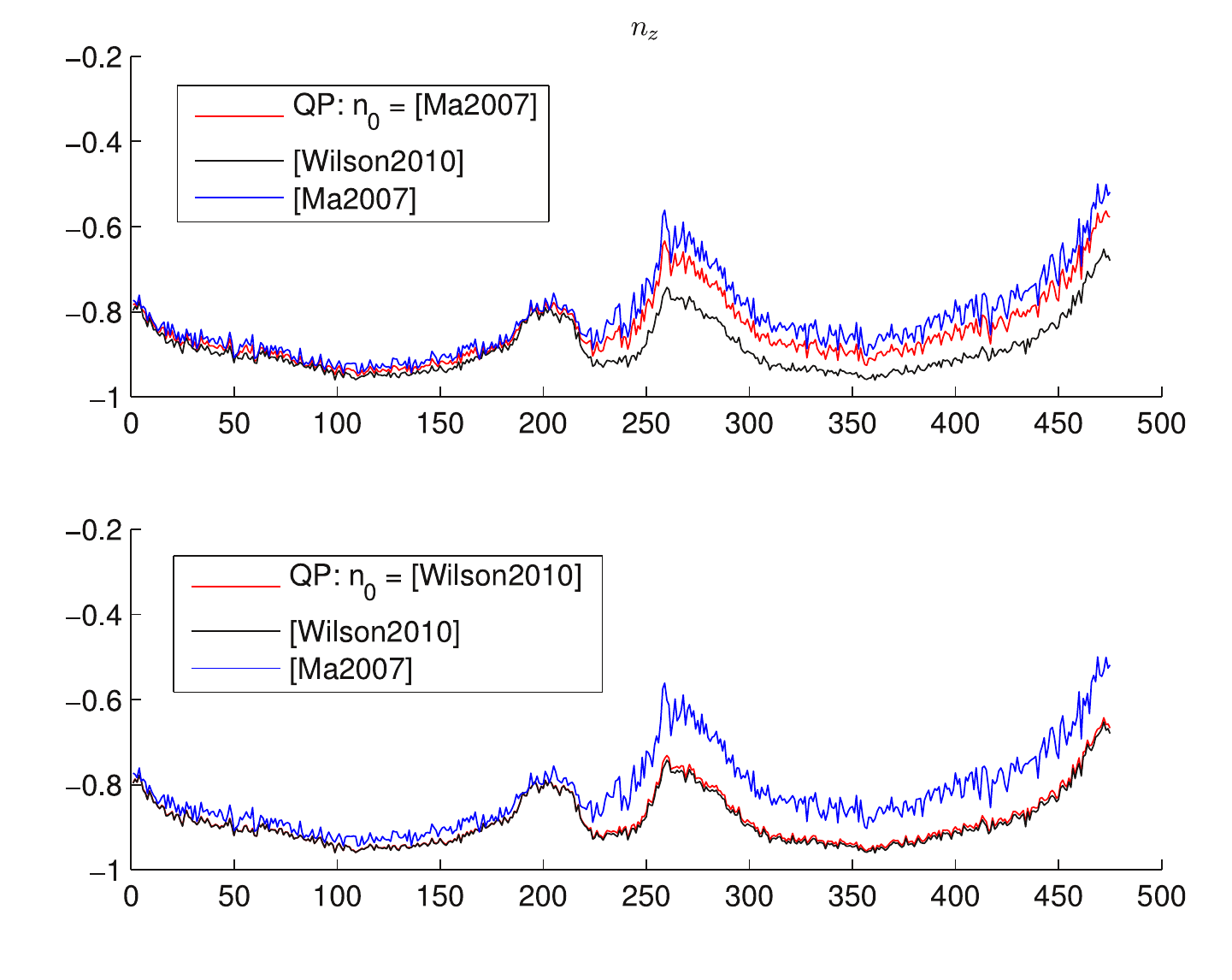}
  \caption{$n_{z}$ component of surface normals in 1 pixel wide selected horizontal region (\figurename~\ref{fig:mls_data_processing_qp_lady_statue_side_view_photo} - right) obtained after applying QP based normal correction with initial estimate of surface normals from (top) Ma \etal method i.e $n_{0} = n^{[Ma2007]}$ and (bottom) Wilson \etal method i.e. $n_{0} = n^{[Wilson]}$.}
  \label{fig:mls_data_processing_horz_stripe_qp_result}
\end{figure}

Now, we analyse the results of QP based surface normal correction applied to the face region of a statue made of white cement: a material that exhibits property very close to an ideal diffuse surface. First, let us consider the surface normals in a $1$ pixel wide vertical region (\figurename~\ref{fig:mls_data_processing_qp_lady_statue_side_view_photo} - left) as shown in \figurename~\ref{fig:mls_data_processing_vert_stripe_qp_result}. When the normals computed by Ma \etal method is the initial solution, the corrected normals tend to move closer towards the surface normals computed using Wilson \etal method. On the other hand, when the normal computed by Wilson \etal method is the initial solution, the corrections computed by QP is insignificant. This suggests that the normals computed using Wilson \etal method is already very close to satisfying the constraints i.e. ${\bf Ax}={\bf b}$. Moreover, \figurename~\ref{fig:mls_data_processing_vert_stripe_qp_result} clearly shows that the corrected normals retain the noise characteristics of the initial solution, irrespective of the choice of initial solution. This behaviour could be attributed to the fact that we apply QP based correction to each image pixel independently and therefore the noise characteristics is propagated to corrected normals. We obtain similar results for the  $1$ pixel wide horizontal region (\figurename~\ref{fig:mls_data_processing_qp_lady_statue_side_view_photo} - right) as shown in \figurename~\ref{fig:mls_data_processing_horz_stripe_qp_result}. From \figurename~\ref{fig:mls_data_processing_horz_stripe_qp_result} (top), it is evident that QP based correction is significant in the region ($250$ to $450$ pixel region) where the initial solution had large deviation from the surface normals computed by Wilson \etal method.

We used MATLAB 7.9 (R2009b) implementation of Quadratic Programming, \textit{qprog()}, running on Slackware 13.1-2-12 on 3 GHz Intel\textregistered Core2 Duo CPU for testing this normal correction algorithm. It takes around $1.78$ hours to perform normal correction on a photometric normal map of size $1624\times  1236$.

\subsection{Discussion}
QP based normal correction algorithm provides insignificant improvement in the recovered surface geometry. However, it will be evident in the next section that this analysis is pivotal to the development of minimal image sets method for robust spherical gradient photometric stereo. The modified radiance equations, which resulted in QP based correction algorithm, not only reveal the limitations of original Ma \etal \cite{ma2007rapid} method but also provide an explanation for the improvement in quality of surface normals recovered by the complement gradient method of Wilson \etal \cite{wilson2010temporal}. Furthermore, in the next section, we use this analysis to show that our proposed minimal image sets method combines the advantage of the original method of Ma \etal (reduced data capture requirement) with that of Wilson \etal (improved robustness). Hence, although the QP based normal correction algorithm did not result in significant improvement over existing methods, it provided us with valuable insight into the limitations and strength of the spherical gradient photometric stereo technique.

\section{Minimal Image Sets for Robust Spherical Gradient Photometric Stereo}
\label{ch:mls_data_processing:sec:minimal_image_sets}
In section \ref{ch:mls_data_processing:sec:quad_prog_for_norm_improvement:subsec:ma2007_nm_quality_assessment}, we used our modified radiance equations to show that the quality of surface normals computed using Ma \etal \cite{ma2007rapid} method will degrade with deformation in diffuse lobe. In this section, using the same modified radiance equations, we show how the method of Wilson \etal \cite{wilson2010temporal} uses a set of $6$ gradient and complement gradient images to cancel out the effects of deformed diffuse lobe. Finally, based on the analysis of spherical gradient photometric stereo using our modified radiance equations, we propose a minimal $4$ image set method that combines the advantage of the original method of Ma et al. (reduced data capture requirement) with that of Wilson et al. (improved robustness).

Recently, Wilson \etal \cite{wilson2010temporal} proposed the use of complement gradient images, in addition to gradient images, to improve the quality of recovered surface normals. This method used the difference of gradient and complement gradient images to recover surface normals. Mathematically, the ``difference method'' of \cite{wilson2010temporal} is given by:
\begin{equation}
 \vec{n} = \frac{[r_{x}-r_{\bar{x}}, r_{y}-r_{\bar{y}}, r_{z}-r_{\bar{z}}]^{T}}{||[r_{x}-r_{\bar{x}}, r_{y}-r_{\bar{y}}, r_{z}-r_{\bar{z}}]||}.
 \label{eq:mls_data_processing_wilson2010_normal_eq1}
\end{equation}
They claimed that this method improves the quality of the normal estimates over estimates from Ma \etal \cite{ma2007rapid}, since ``the pixels that are dark under one gradient illumination condition are most likely well exposed under the complement gradient illumination condition'' \cite[p17:5]{wilson2010temporal}. Indeed, the validity of this claim is easily demonstrated by our modified radiance equations. Once again, we use our modified radiance equations to represent the surface normal computed using Wilson \etal method:
\begin{eqnarray}
 N_{\{x,y,z\}} & = & r_{\{x,y,z\}} - r_{\{\bar{x},\bar{y},\bar{z}\}} \nonumber \\
 & = & \frac{\pi\rho_{D}V_{p}}{2} \left\{ \delta_{\{\bar{x},\bar{y},\bar{z}\}} + \frac{2}{3}n_{\{x,y,z\}} \right\}.
 \label{eq:mls_data_processing_wilson2010_normal_eq2}
\end{eqnarray}
As the reader considers equation (\ref{eq:mls_data_processing_wilson2010_normal_eq2}), it is critical to understand that we arrived at this expression using the modified radiance equations that are based on the following simplifying assumption described in section \ref{ch:mls_data_processing:sec:quad_prog_for_norm_improvement:subsec:modified_radiance_eq_grad_illum} and \ref{ch:mls_data_processing:sec:quad_prog_for_norm_improvement:subsec:complement_constraints}: overall deformation in the diffuse reflectance lobe for gradient and complement gradient illumination environment can be quantified using a single scalar parameter $\delta_{\{x,y,z\}}$ and $\delta_{\{x,y,z\}} + \delta_{\{\bar{x},\bar{y},\bar{z}\}}$ respectively.

The interesting observation in (\ref{eq:mls_data_processing_wilson2010_normal_eq2}) is that the symmetric distortion of diffuse lobe cancel out and the only component contributing to error in the recovery of surface normal is the asymmetric distortion parameter $\delta_{\{\bar{x},\bar{y}, \bar{z}\}}$. In other words, symmetric deformations in the reflection lobe are averaged out and therefore the surface normals recovered by Wilson \etal \cite{wilson2010temporal} method are less affected by deformation to the diffuse reflectance lobe caused by shadowing, inter-reflection and coarse approximation of spherical illumination due to light discretization. Therefore, we conclude that the method of Wilson \etal \cite{wilson2010temporal} recovers surface geometry closer to the true surface geometry because its ``difference method'' involves cancellation of symmetric deformation in diffuse reflectance lobes. This ``symmetric deformation'' cancellation property is not present in the ``ratio image'' method proposed by Ma \etal \cite{ma2007rapid}. Note that the remaining constant cancel out during the vector normalization step.

Using our modified radiance equations and building upon the ``difference method'' proposed by \cite{wilson2010temporal}, we derive a minimal four image solution in which symmetric deformations of diffuse lobes still cancel. We exploit the following complement image contraint to arrive at the minimal four image solution:
\begin{equation}
 r_{x}+r_{\bar{x}}=r_{y}+r_{\bar{y}}=r_{z}+r_{\bar{z}}=r_{c}.
 \label{eq:complement_image_constraint}
\end{equation}
Using this X complement image constraint, we rewrite (\ref{eq:mls_data_processing_wilson2010_normal_eq1}) as:
\begin{equation}
{\bf n_{(x,y,z,\bar{x})}} = \frac{[r_{x}-r_{\bar{x}}, 2r_{y}-(r_{x}+r_{\bar{x}}), 2r_{z}-(r_{x}+r_{\bar{x}})]^{T}}{\|[r_{x}-r_{\bar{x}}, 2r_{y}-(r_{x}+r_{\bar{x}}), 2r_{z}-(r_{x}+r_{\bar{x}})]^{T}\|}
\label{eq:mls_data_processing_minimal_img_set_eq_n_xyz_xbar}
\end{equation}
Similarly, Y and Z base complement pairs can also be used to obtain ${\bf n_{(x,y,z,\bar{y})}}$ and ${\bf n_{(x,y,z,\bar{z})}}$ as follows:
\begin{equation}
{\bf n_{(x,y,z,\bar{y})}} = \frac{[2r_{x}-(r_{y}+r_{\bar{y}}), r_{y}-r_{\bar{y}}, 2r_{z}-(r_{y}+r_{\bar{y}})]^{T}}{\|[2r_{x}-(r_{y}+r_{\bar{y}}), r_{y}-r_{\bar{y}}, 2r_{z}-(r_{y}+r_{\bar{y}})]^{T}\|}
\label{eq:mls_data_processing_minimal_img_set_eq_n_xyz_ybar}
\end{equation}
\begin{equation}
{\bf n_{(x,y,z,\bar{z})}} = \frac{[2r_{x}-(r_{z}+r_{\bar{z}}), 2r_{y} - (r_{z}+r_{\bar{z}}), r_{z}-r_{\bar{z}}]^{T}}{\|[2r_{x}-(r_{z}+r_{\bar{z}}), 2r_{y} - (r_{z}+r_{\bar{z}}), r_{z}-r_{\bar{z}}]^{T}\|}
\label{eq:mls_data_processing_minimal_img_set_eq_n_xyz_zbar}
\end{equation}
In a similar way, we can also derive expressions for complement minimal image sets: ${\bf n_{(\bar{x},\bar{y},\bar{z},x)}}$, ${\bf n_{(\bar{x},\bar{y},\bar{z},y)}}$ and ${\bf n_{(\bar{x},\bar{y},\bar{z},z)}}$. Therefore, we have total six image sets in our minimal image sets formulation: ${\bf n_{(x,y,z,\{\bar{x},\bar{y},\bar{z}\})}}$ and ${\bf n_{(\bar{x},\bar{y},\bar{z},\{x,y,z\})}}$.

In the next section, we show that there is very small angular deviation ($\sim3.9^{\circ}$) between the normals computed using Wilson \etal method and our method. This observation supports our claim that above substitution indeed preserves the ``symmetric deformation cancellation property''.

\begin{figure}[htbp]
  \centering
  \includegraphics{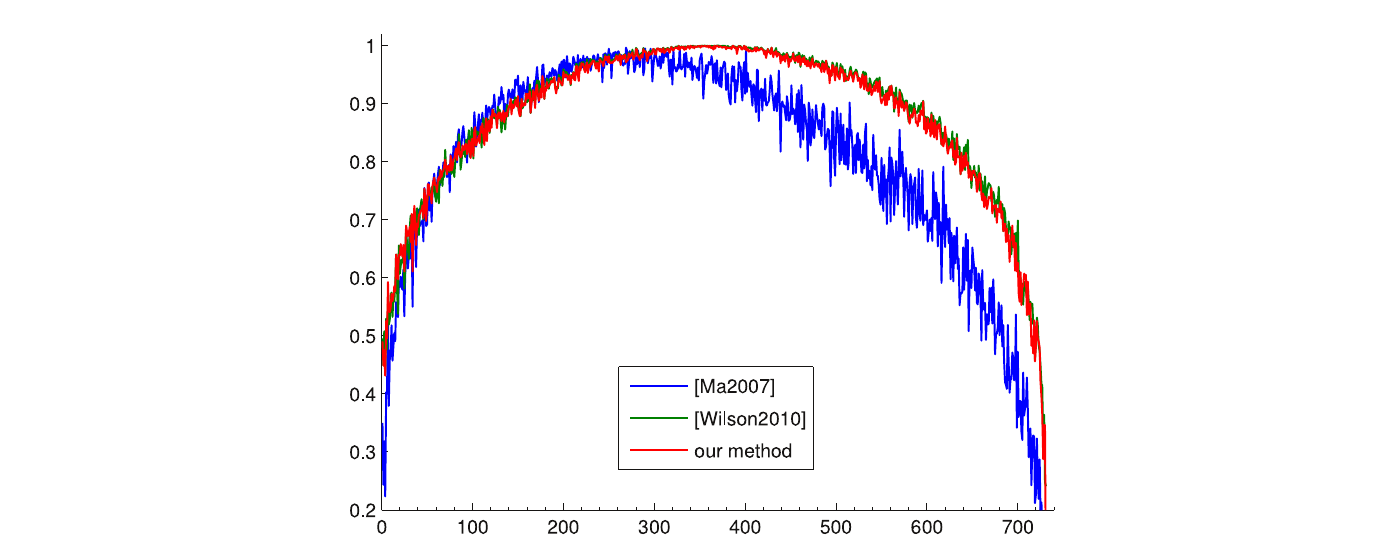}
  \caption{z-component ($n_{z}$) of estimated surface normals of a cylinder}
  \label{fig:mls_data_processing_cylinder_nz_ma2007_wilson2010_dutta2010_comparision}
\end{figure}

The non-symmetric deformation $\delta_{\{\bar{x},\bar{y},\bar{z}\}}$ do not cancel and still contribute to error in the recovered surface normals. To analyze the influence of non-symmetric deformation, we computed surface normal of a cylinder as shown in \figurename\ref{fig:mls_data_processing_cylinder_nz_ma2007_wilson2010_dutta2010_comparision} using Wilson \etal \cite{wilson2010temporal} and our method. If there were significant contribution of non-symmetric deformation, the normals computed using Wilson \etal method and our method would have deviated strongly from the ground truth (not shown in the plot as it aligns with surface recovered using Wilson \etal method). Therefore, we conclude that in practice the contribution of non-symmetric deformation is very small. Unavailability of ground truth data prevented us from verifying this claim for other more complex surfaces like a human face. 

\subsection{Results}
From our analysis in previous section, we concluded that the method of Wilson \etal \cite{wilson2010temporal} recovers optimal surface geometry as it involves cancellation of symmetric deformation in diffuse reflectance lobes. Hence, we use the normal map recovered using \cite{wilson2010temporal} to assess the quality of normals recovered using our minimal image sets method and that obtained from Ma \etal \cite{ma2007rapid}. First let us analyse the results for a static object (a statue). This object is made up of white cement plaster and hence its reflectance properties are very close to an ideal diffuse surface. Moreover, the static nature of this object ensures that the captured gradient images are perfectly aligned\footnote{misalignment can be caused by motion of the subject during the capture process}.

\begin{figure}[htbp]
  \centering
  \includegraphics{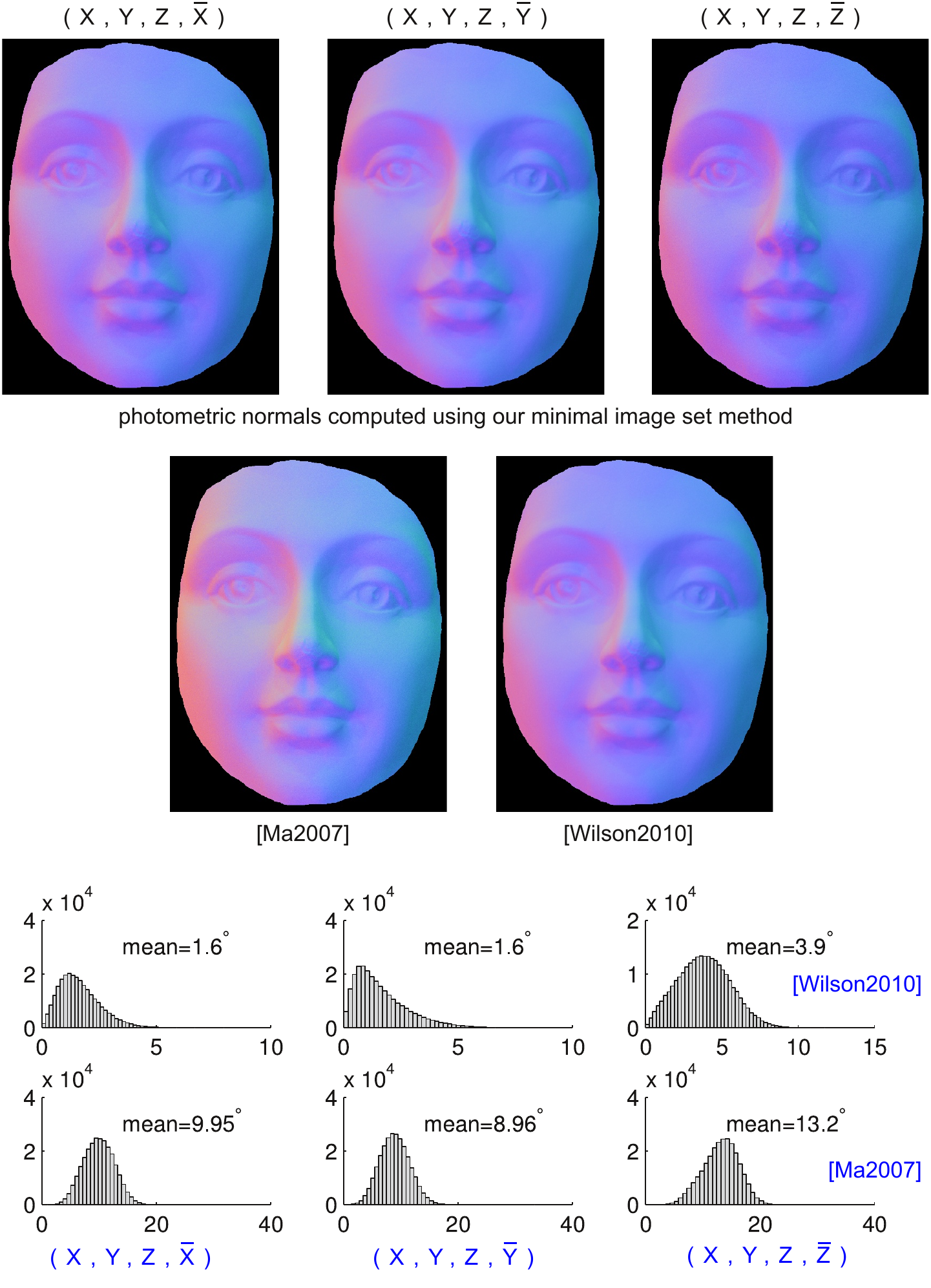}
  \caption{Photometric normals of a statue (static) computed using our minimal image set method (top) and that computed using Ma \etal \cite{ma2007rapid} (middle left) and Wilson \etal \cite{wilson2010temporal} (middle right). All the three complement base pairs --- $(X,\bar{X}), (Y, \bar{Y}) \textrm{and} (Z,\bar{Z})$ --- possible in our minimal image set method was used to generate similar photometric normals. (Bottom) Distribution of angular difference between normal maps computed using our minimal image set method and that computed using \cite{ma2007rapid} and \cite{wilson2010temporal}.}
  \label{fig:mls_data_processing_minimal_img_set_ladystatue_nm_comp_wilson2010_ma2007}
\end{figure}

In \figurename\ref{fig:mls_data_processing_minimal_img_set_ladystatue_nm_comp_wilson2010_ma2007} (top row), we show the normal maps computed using the three possible minimal image sets : $(X,Y,Z,\bar{X})$, $(X,Y,Z,\bar{Y})$ and $(X,Y,Z,\bar{Z})$. The normal maps computed for same gradient images using Wilson \etal and Ma \etal is shown in the middle row of \figurename\ref{fig:mls_data_processing_minimal_img_set_ladystatue_nm_comp_wilson2010_ma2007}. The bottom row in this figure shows the distribution of angular error between normal map computed using our minimal image set method and that computed using \cite{wilson2010temporal} and \cite{ma2007rapid}. It is evident from these histogram that the normal map estimated using our minimal image sets method (requiring just $4$ images) is very close to that estimated by \cite{wilson2010temporal} (requiring $6$ images). Also, the angular difference is not very large for the normal map estimate by \cite{ma2007rapid}. Hence, analysis of normal map estimates of a static object suggests that there is not much significant difference in the normal maps computed using these three methods.

\begin{figure}[htbp]
  \centering
  \includegraphics{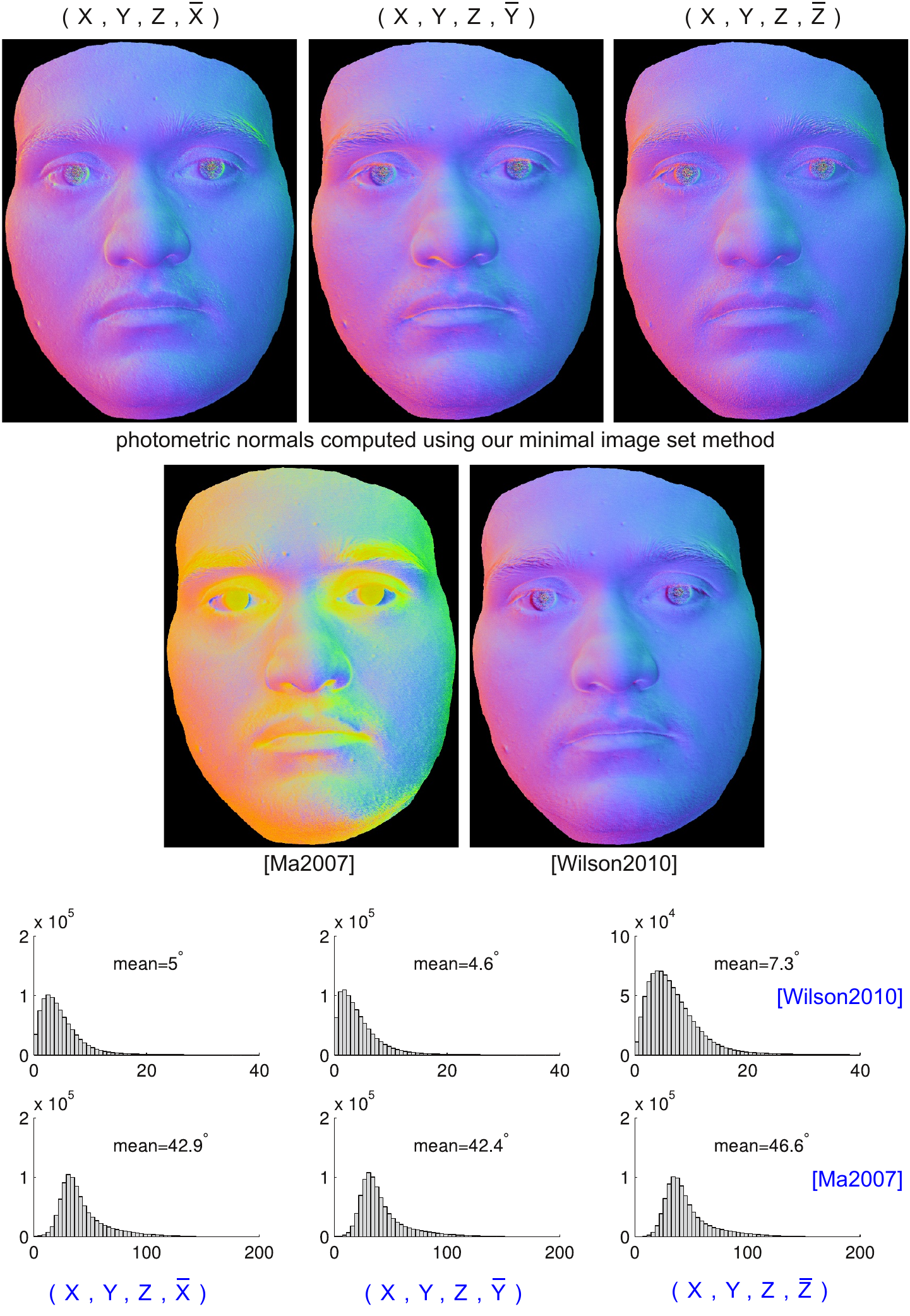}
  \caption{Photometric normals of a face (non-static) computed using our minimal image set method (top) and that computed using Ma \etal \cite{ma2007rapid} (middle left) and Wilson \etal \cite{wilson2010temporal} (middle right). All the three complement base pairs --- $(X,\bar{X}), (Y, \bar{Y}) \textrm{and} (Z,\bar{Z})$ --- possible in our minimal image set method was used to generate similar photometric normals. (Bottom) Distribution of angular difference between normal maps computed using our minimal image set method and that computed using \cite{ma2007rapid} and \cite{wilson2010temporal}.}
  \label{fig:mls_data_processing_minimal_img_set_abhishek_nm_comp_wilson2010_ma2007}
\end{figure}

Now let us perform similar analysis for the normal map of a human face. Small motion between gradient images of non-static objects, like a human face, is unavoidable. This results in misalignment of gradient images and therefore causes surface normal deviations that cannot be modelled using our ``diffuse reflectance lobe distortion'' framework. Our Light Stage has significant \textit{``light discretization''} as we use only $41$ LED ($74\%$ less than the Light Stage of \cite{ma2007rapid} and \cite{wilson2010temporal}). This contributes significantly to the deformation of diffuse lobes. The normal estimation technique of Ma \etal is unable to cope with distortion in the diffuse lobes. This causes the recovered surface normal to have large deviation from the true surface normal. From our analysis in \ref{ch:mls_data_processing:sec:minimal_image_sets}, we know that if the deformation in diffuse lobes is symmetric in the complement images, the normal estimation technique of Wilson \etal results in cancellation of these deformations. Our minimal image sets formulation preserves this ``deformation cancellation'' property and hence there is very small angular difference ($\sim7.3^{\circ}$) with the normal map computed using Wilson \etal as shown in \figurename\ref{fig:mls_data_processing_minimal_img_set_abhishek_nm_comp_wilson2010_ma2007}. The inability of Ma \etal method to cope with deformation in diffuse lobe is also evident from the distribution of angular deviation shown in \figurename\ref{fig:mls_data_processing_minimal_img_set_abhishek_nm_comp_wilson2010_ma2007} (bottom). It exhibits large angular deviation ($>42^{\circ}$) with our minimal image set normal map and that of Wilson \etal \cite{wilson2010temporal}. 

\subsection{Discussion}
The method of Ma \etal also used 4 images, $r_{x}, r_{y}, r_{z}, r_{c}$, but it does not use the additional information about deformation in reflectance lobes obtained from complement gradient images. On the other hand, Wilson \etal method requires 6 images, $r_{x}, r_{y}, r_{z}, r_{\bar{x}}, r_{\bar{y}}, r_{\bar{z}}$, to compensate for deformation of reflectance lobes. Our method requires only 4 images, $r_{x}, r_{y}, r_{z}, r_{\bar{x}}$, because it exploits the information obtained from X and complement X gradient condition in the estimation of Y and Z complements. In other words, this new method combines the advantage of the original method of Ma \etal (reduced data capture requirement) with that of Wilson \etal (improved robustness). This new formulation is able to reduce the data requirements which is extremely important if the spherical gradient photometric stereo is to be used for real time performance capture as discussed in \ref{ch:app_of_light_stage_data:sec:real_time_perf_cap}.

It is important to understand that our analysis is based on the following simplifying assumption described in section \ref{ch:mls_data_processing:sec:quad_prog_for_norm_improvement:subsec:modified_radiance_eq_grad_illum} and \ref{ch:mls_data_processing:sec:quad_prog_for_norm_improvement:subsec:complement_constraints}: overall deformation in the diffuse reflectance lobe for a gradient and complement gradient illumination environment can be quantified using a single scalar parameter $\delta_{\{x,y,z\}}$ and $\delta_{\{x,y,z\}} + \delta_{\{\bar{x},\bar{y},\bar{z}\}}$ respectively.

\section{Registration of Spherical Illumination Images}
\label{ch:mls_data_processing:sec:reg_of_sph_illum_images}
Spherical gradient photometric stereo technique requires capture of $4$ spherical illumination images ($X,Y,Z,C$) with the assumption that the imaged object remains at the same position during the capture process. In other words, a pixel position in all the gradient images should correspond to the same surface patch. However, for non-static objects like a human face, it is difficult to remain at same position during the capture of these $4$ images. Even at high capture frame rate, apparant motion between $1^{st}$ and $4^{th}$ image is unavoidable which causes some inaccuracy in the photometric normals computed using misaligned gradient images. Hence, in order to recover accurate photometric normals, we must align these gradient images to the constant illumination image. This task is achieved by the Joint Photometric Alignment method proposed by Wilson \etal \cite{wilson2010temporal}.

Traditional optical flow techniques have been successfully applied for alignment of images consisting of small motion of the imaged object. Such techniques estimate the apparent motion of object in a sequence of images by exploiting the brightness constancy assumption i.e. corresponding image points maintain their brightness level despite apparent motion. Mathematically, this assumption can be expressed as:
\[
 I(x,t) = I(x+u, t+1)
\]
where, $I(x,t)$ is the image pixel value at a 2D spatial location $\vec{x} = [x\;y]^{T}$ and time $t$. Optical flow based alignment techniques estimate the 2D warp function $u$ (flow field) to minimise
\[
 u \leftarrow argmin_{u} \;\varepsilon \left(I(x+u,t+1), I(x,t)\right)
\]
where, $\varepsilon(.)$ is the error function which quantifies the extent of misalignment between the source $I(x,t)$ and target $I(x+u,t+1)$ images. The $4$ illumination conditions $(X,Y,Z,C)$ in spherical gradient photometric stereo are delibrately designed to dramatically change the pixel brightness of each image point in order to reveal the corresponding surface geometry. This causes violation of the brightness constancy assumption in the $4$ images and hence traditional optical flow based tecniques cannot be directly applied to align the gradient images.

The Joint Photometric Alignment technique of Wilson \etal \cite{wilson2010temporal} can align these gradient images at the expense of capturing additional 3 images called the complement gradient images\footnote{introduced in section \ref{ch:mls_data_processing:sec:quad_prog_for_norm_improvement:subsec:complement_constraints}} - $(\bar{X}, \bar{Y}, \bar{Z})$.  They exploit the complement image constraint to align the gradient $(X,Y,Z)$ and complement gradient images $(\bar{X}, \bar{Y}, \bar{Z})$ to the constant illumination image $C$ (also called tracking frame). Mathematically, the complement image constraint can be expressed as:
\begin{equation}
 r_{\{x,y,z\}} + r_{\{\bar{x},\bar{y},\bar{z}\}} = r_{c}
\end{equation}
where, $r_{\{x,y,z\}}$, $r_{\{\bar{x},\bar{y},\bar{z}\}}$ and $r_{c}$ represent gradient, complement gradient and constant illumination image respectively. The Joint Photometric Alignment method is an iterative algorithm that estimates optimal 2D warp functions $u$ and $v$ (flow fields) for the gradient and complement gradient images such that extent of complement constraint violation is minimized. Bootstrapping both flows ($u$ and $v$) initialized to zero, the iterative algorithm proceeds to minimise the following error in each iteration:
\begin{eqnarray*}
  u^{(i+1)} & \leftarrow & \textrm{argmin}_{u} \; \varepsilon\left( r_{\{x,y,z\}}(u), c - r_{\{\bar{x},\bar{y},\bar{z}\}}(v^{(i)}) \right) \\
  v^{(i+1)} & \leftarrow & \textrm{argmin}_{v} \; \varepsilon\left( r_{\{\bar{x},\bar{y},\bar{z}\}}(v), c -  r_{\{x,y,z\}}(u^{(i+1)}) \right)
\end{eqnarray*}

\subsection{Result from alignment of gradient images}
\begin{figure}[htbp]
  \centering
  \includegraphics{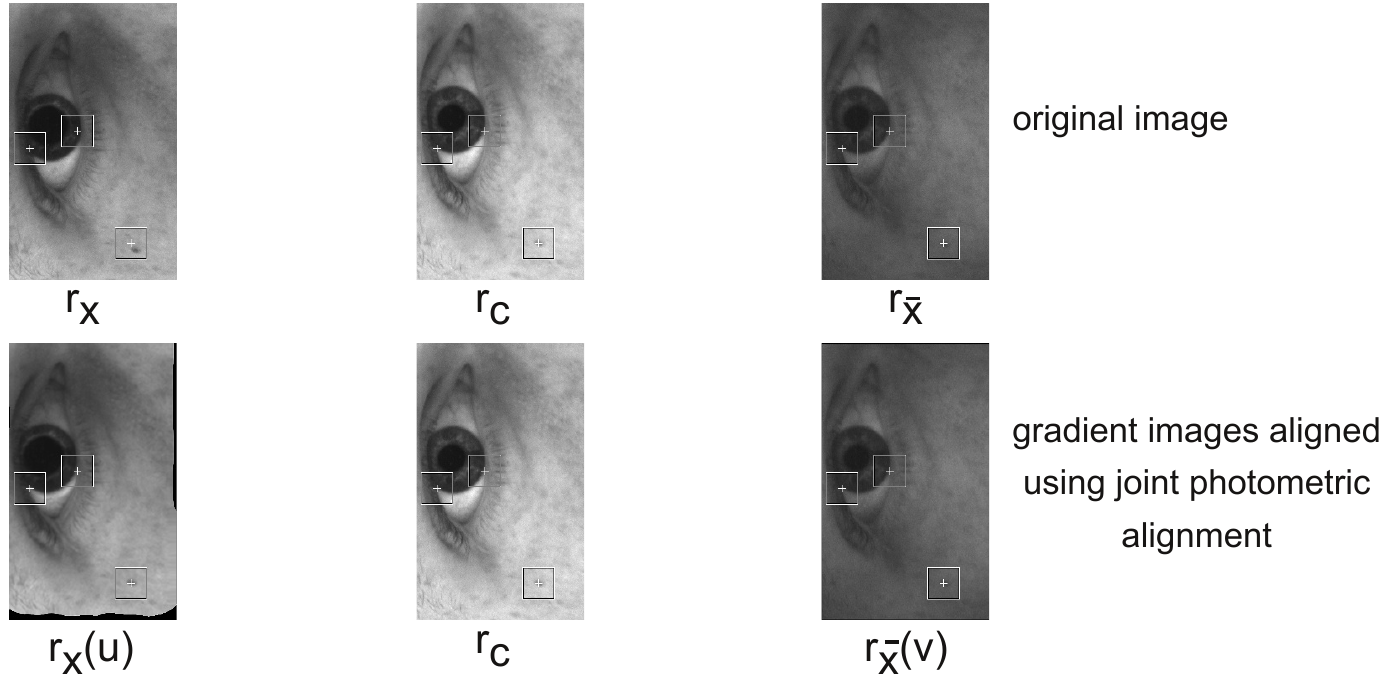}
  \caption{Alignment of spherical gradient images used Joint Photometric Alignment\cite{wilson2010temporal}. For illustration purpose, all the intensity values were scaled by $2$ except the warped $r_{\bar{x}}(v)$ which was scaled by $3$ because it falls on the dark side of spherical gradient illumination.}
  \label{fig:mls_data_processing_x_gradient_alignment_illustration}
\end{figure}
To illustrate the result of joint photometric alignment, we marked $3$ feature points (cross hair inside a bounding rectangle) in the gradient (top left), constant (top center) and complement gradient (top right) illumination images as shown in \figurename\ref{fig:mls_data_processing_x_gradient_alignment_illustration}. We used the following image capture sequence: $X,Z,Y,C,\bar{X}, \bar{Z}, \bar{Y}$. As $C$ and $\bar{X}$ are consecutive frames in the capture sequence, the apparant motion of feature points (clearly visible due to bounding rectangle) is negligible and hence requires no warping. However, $X$ and $C$ are two frames apart in the capture sequence and hence there is significant displacement of the feature points. After the application of joint photometric alignment technique, the marked feature points get aligned in the warped $X$ gradient image (bottom left) as shown in \figurename\ref{fig:mls_data_processing_x_gradient_alignment_illustration}. Large value of flow field $u$ for $X$ gradient image is evident from dark regions in the boundary of the corresponding warped image.

\begin{figure}[htbp]
  \centering
  \includegraphics{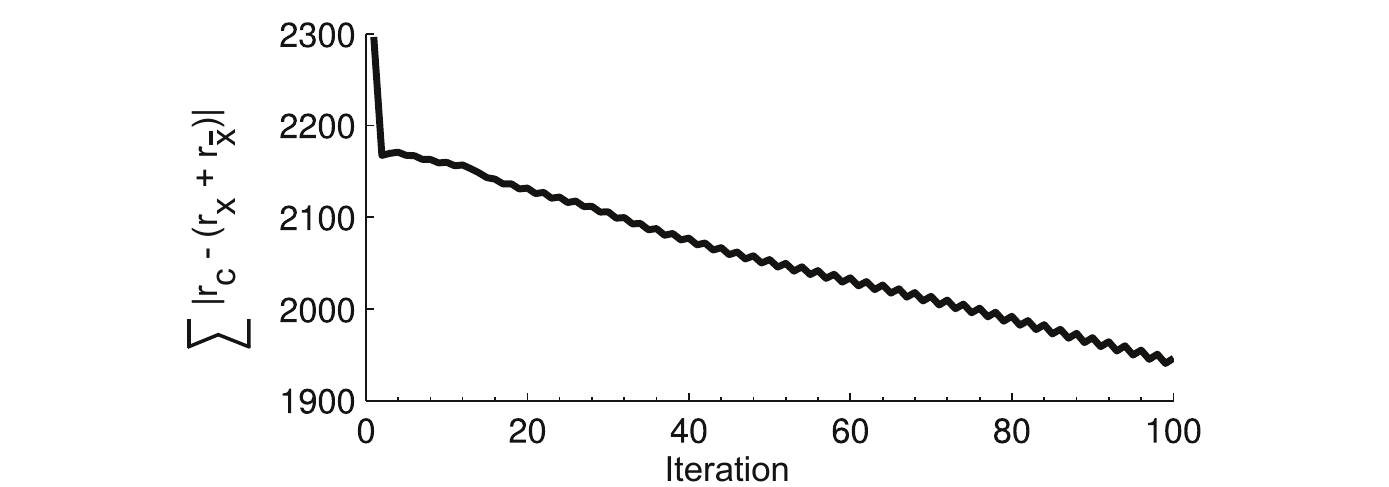}
  \caption{Complement constrain residual for $100$ iterations of the joint photometric alignment technique applied to a $298\times182$ spherical X gradient image}
  \label{fig:mls_data_processing_joint_photometric_alignment_error_plot}
\end{figure}

As reported in \cite{wilson2010temporal}, the iterative nature of this alignment technique requires considerable amount of time to arrive at acceptable level of alignment. For a $298\times182$ grayscale image, it took $595.30$ sec ($\sim10$ min.) to complete $10$ iterations\footnote{each iteration involves two execution of Brox \etal \cite{brox2004high} optical flow technique (C implementation provided by the authors) running in Slackware 13.1-2-12 on 3 GHz Intel\textregistered Core2 Duo CPU}. The plot of residual\footnote{this residual, $\sum|r_{c} - (r_{x}+r_{\bar{x}})|$, quantifies the extent of complement constraint violation} at each iteration is shown in \figurename\ref{fig:mls_data_processing_joint_photometric_alignment_error_plot}. This plot depicts that at each iteration there is significant reduction in residual and hence the number of iterations should be large for the joint photometric alignment technique to converge at the optimal flow field value.

\chapter{Applications of the Light Stage}
\label{ch:app_of_light_stage_data}
A Light Stage provides a rich source of geometric and photometric information that is useful for many research avenues in, but not limited to, Computer Vision and Computer Graphics. In this chapter, we will discuss two such applications of the Light Stage which we have explored.

\section{Real Time Performance Capture}
\label{ch:app_of_light_stage_data:sec:real_time_perf_cap}
The real time facial geometry of a dynamic performance can be captured using the spherical gradient photometric stereo based performance capture and photometric alignment method proposed by Wilson \etal \cite{wilson2010temporal}. We modify the capture sequence proposed by Wilson \etal based on our minimal image sets for robust spherical gradient photometric stereo (discussed in section \ref{ch:mls_data_processing:sec:minimal_image_sets}). This modified performance capture sequence results in:
\begin{itemize}
 \item Reduced data capture requirement for real time performance capture without compromising the quality of recovered photometric normals. In other words, we show that only $5$ spherical illumination images, instead of $7$, is sufficient for estimation of tracking frame photometric normal and corresponding warped normals.
 \item Lower post processing overhead because the modified capture sequence requires joint photometric alignment of only one pair, instead of three, of gradient and complement gradient images. In other words, the post processing time required for alignment of gradient images is significantly reduced because only one pair of gradient images, farthest from the tracking frame, require alignment.
\end{itemize}

\begin{sidewaysfigure}
  \centering
  \includegraphics{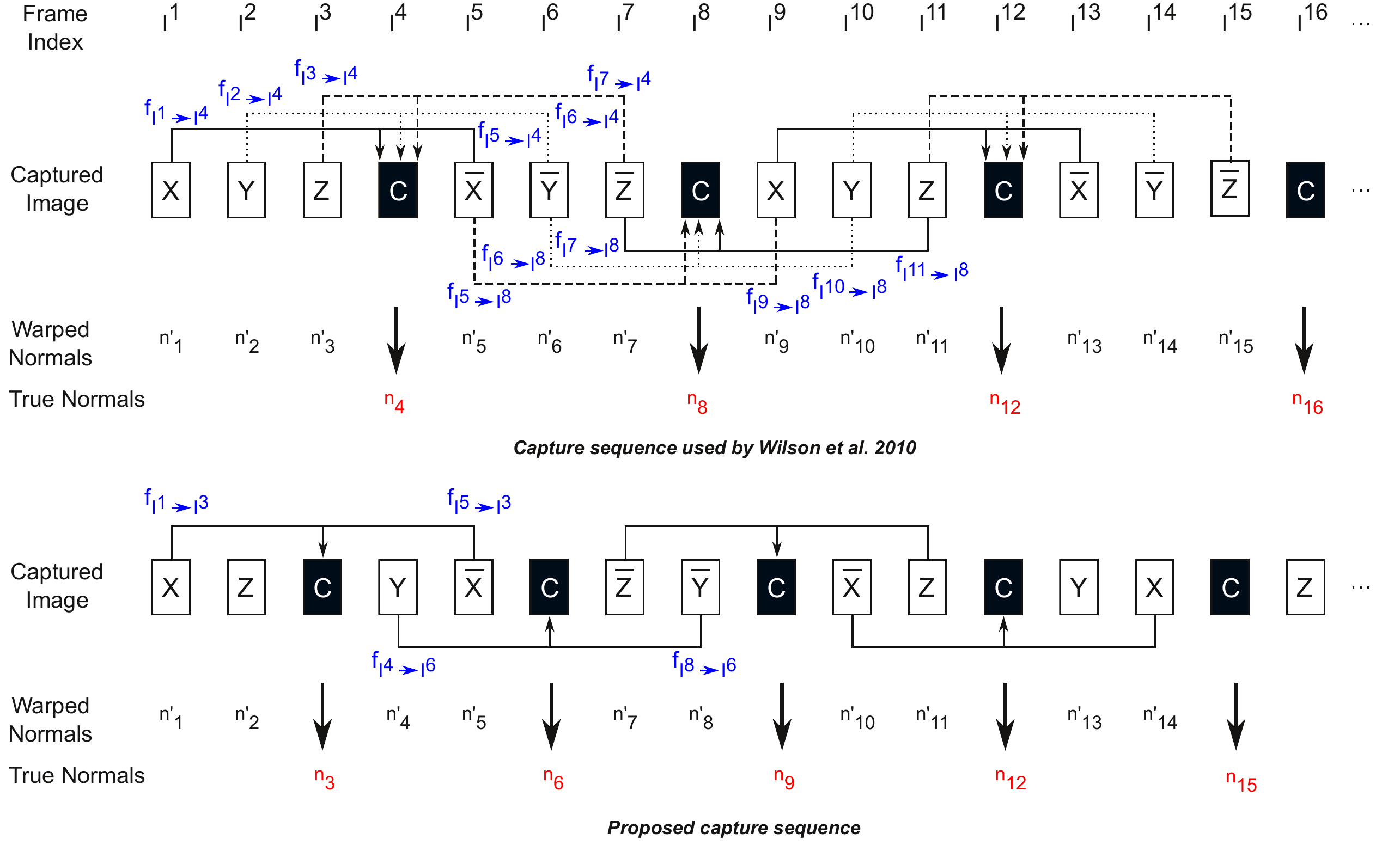}
  \caption{Image capture sequence used by Wilson \etal \cite{wilson2010temporal} (top). The proposed image capture sequence (bottom) based on our minimal image sets method.}
	\label{fig:app_of_light_stage_data:sec:real_time_perf_cap_perf_cap_seq_wilson2010_dutta2010}
\end{sidewaysfigure}

\subsection{Original Performance Geometry Capture Method}
In this section, we briefly describe the real time performance geometry capture method proposed by Wilson \etal \cite{wilson2010temporal}. The performance capture sequence developed by Wilson \etal is depicted in \figurename\ref{fig:app_of_light_stage_data:sec:real_time_perf_cap_perf_cap_seq_wilson2010_dutta2010} (top). The frame index $(\textrm{I}^{1}, \textrm{I}^{2}, \textrm{I}^{3}, ...)$ is used to indicate the temporal sequence of each illumination condition in the capture sequence. Wilson \etal developed the Joint Photometric Alignment method (discussed in section \ref{ch:mls_data_processing:sec:reg_of_sph_illum_images}), for the alignment of gradient $\{X,Y,Z\}$ and complement gradient $\{\bar{X}, \bar{Y}, \bar{Z}\}$ images to the flanked constant illumination $C$ image i.e. the tracking frame. For example: the Joint Photometric Alignment applied to frames $(\textrm{I}^{1}, \textrm{I}^{4}, \textrm{I}^{5})$ results in two optical flow fields, $f_{\textrm{I}^{1} \rightarrow \textrm{I}^{4}}$ and $f_{\textrm{I}^{5} \rightarrow \textrm{I}^{4}}$, that align the gradient $X$ and complement gradient $\bar{X}$ images to the tracking frame $C$ respectively. In other words, warping $X$ and $\bar{X}$ according to $f_{\textrm{I}^{1} \rightarrow \textrm{I}^{4}}$ and $f_{\textrm{I}^{5} \rightarrow \textrm{I}^{4}}$ respectively, aligns both $X$ and $\bar{X}$ to the tracking frame $C$. We represent the warping operation of an image $I$ by the flow field $f$ as $\textbf{W}(I,f)$ where $I \in \mathbb{R}^{N \times M}$. Note that warping of a vector field $N$, involves warping of each normal vector component followed by renormalisation which can be represented as $\textbf{W}(N,f)$ where $N \in \mathbb{R}^{N \times M \times 3}$.

\subsubsection{Tracking Frame Photometric Normals}
The photometric normal computed at each tracking frame represents the true\footnote{referring to non-warped normal and not the ground truth normal} normal. At each tracking frame, we have access to $3$ pairs of aligned gradient and complement gradient images. These aligned gradient images are used for computation of photometric normal at each tracking frame. For example, in \figurename\ref{fig:app_of_light_stage_data:sec:real_time_perf_cap_perf_cap_seq_wilson2010_dutta2010}, photometric normal at the tracking frame $\textrm{I}^{4}$ is given by:

\begin{equation}
 n_{4} = \frac{\left[X_{w} - \bar{X}_{w}, Y_{w} - \bar{Y}_{w}, Z_{w} - \bar{Z}_{w}\right]}{||\left[X_{w} - \bar{X}_{w}, Y_{w} - \bar{Y}_{w}, Z_{w} - \bar{Z}_{w}\right]||}
\end{equation}
where,
\begin{eqnarray*}
 X_{w} = \textbf{W} \left(X,f_{\textrm{I}^{1} \rightarrow \textrm{I}^{4}} \right) & , & \bar{X}_{w} = \textbf{W} \left(\bar{X},f_{\textrm{I}^{5} \rightarrow \textrm{I}^{4}} \right) \\
 Y_{w} = \textbf{W} \left(Y,f_{\textrm{I}^{2} \rightarrow \textrm{I}^{4}} \right) & , & \bar{Y}_{w} = \textbf{W} \left(\bar{Y},f_{\textrm{I}^{6} \rightarrow \textrm{I}^{4}} \right) \\
 Z_{w} = \textbf{W} \left(Z,f_{\textrm{I}^{3} \rightarrow \textrm{I}^{4}} \right) & , & \bar{Z}_{w} = \textbf{W} \left(\bar{Z},f_{\textrm{I}^{7} \rightarrow \textrm{I}^{4}} \right)
\end{eqnarray*}

In a similar way, all the tracking frame photometric normals can also be computed. Note that, we represent tracking frame photometric normals by the symbol $n_{\{x\}}$ while the warped photometric normals are depicted as $n^{'}_{\{x\}}$.

\subsubsection{Warped Normals at Intermediate Gradient Frames}
With the flow field from each gradient and complement gradient to a common tracking frame at hand, Wilson \etal warped the tracking frame photometric normals to obtain normals corresponding to the temporal location of gradient and complement gradient images. Wilson \etal used the term ``Temporal Up-sampling'' to refer to this operation of estimating photometric normal at non-tracking frames.

The warped normal at frames $\textrm{I}^{1}, \textrm{I}^{2}, \textrm{I}^{3}$ is given by:

\[
 n^{'}_{1} = \textbf{W} \left(n_{4},-f_{\textrm{I}^{1} \rightarrow \textrm{I}^{4}} \right)  \quad,\quad
 n^{'}_{2} = \textbf{W} \left(n_{4},-f_{\textrm{I}^{2} \rightarrow \textrm{I}^{4}} \right)  \quad,\quad
 n^{'}_{3} = \textbf{W} \left(n_{4},-f_{\textrm{I}^{3} \rightarrow \textrm{I}^{4}} \right)
\]
Recall that we use the symbol $n'_{\{x\}}$ to represent the warped photometric normals.

For each subsequent frames, every gradient frame is flanked by two tracking frames. Therefore, two flow fields exist for each gradient frame and hence, there are two versions of warped photometric normal corresponding to each gradient frame. For example, if we consider the gradient image $\bar{X}$ at frame location $\textrm{I}^{5}$, we have the following two warped normals for this frame location:
\[
 n^{''}_{5} = \textbf{W} \left(n_{4},-f_{\textrm{I}^{5} \rightarrow \textrm{I}^{4}} \right)  \quad,\quad
 n^{'''}_{5} = \textbf{W} \left(n_{8},-f_{\textrm{I}^{5} \rightarrow \textrm{I}^{8}} \right)
\]
Wilson \etal used the weighted average (weighted according to the temporal distance) of these two warped normals as the photometric normal for intermediate gradient frames.
\[
 n^{'}_{5} = \frac{3n^{''}_{5}+n^{'''}_{5}}{||3n^{''}_{5}+n^{'''}_{5}||}
\]
In a similar way, we can compute warped normals at all the remaining intermediate gradient frames.

\section{Performance Capture Sequence based on Minimal Image Sets}
We can modify the original performance capture sequence of Wilson \etal in order to exploit the minimal image set method of computing photometric normal. Based on the analysis presented in \ref{ch:mls_data_processing:sec:minimal_image_sets}, we can say that a set of $3$ gradient images $(X, Y, Z)$ and any $1$ of the three complement gradient images $(\bar{X}, \bar{Y}, \bar{Z})$ is sufficient to compute photometric normals. So we can mathematically represent our minimal $4$ image set as $\left[r_{x}, r_{y}, r_{z}, \{ r_{\bar{x}}, r_{\bar{y}}, r_{\bar{z}} \}\right]$. For example, if our minimal $4$ image set is $[r_{x}, r_{y}, r_{z}, r_{\bar{x}}]$, then we have the following expression for the corresponding photometric normal:
\begin{equation}
  \vec{n} = \frac{[r_{x}-r_{\bar{x}}, 2r_{y}-(r_{x} + r_{\bar{x}}), 2r_{z}-(r_{x} + r_{\bar{x}})]^{T}}{||[r_{x}-r_{\bar{x}}, 2r_{y}-(r_{x} + r_{\bar{x}}), 2r_{z}-(r_{x} + r_{\bar{x}})]||}.
  \label{eq:real_time_perf_cap_dutta2010_normal_eq1}
\end{equation}
It is imperative to recall that minimal $4$ image set and the expression for surface normal is valid only when the gradient and complement gradient image satisfy the complement image constraint. Mathematically, if $[r_{x}, r_{y}, r_{z}, r_{\bar{x}}]$ is the $4$ image set, then the following complement image constraint must hold true.
\[
  r_{x} + r_{\bar{x}} = r_{c}
\]
where, $r_{c}$ is the constant illumination image and $r_{x}$ and $r_{\bar{x}}$ form the base complement pair.

The dual of this proposition also exists. The dual minimal $4$ image set can be represented as $\left[r_{\bar{x}}, r_{\bar{y}}, r_{\bar{z}}, \{r_{x}, r_{y}, r_{z}\}\right]$. For example, if our minimal $4$ image set is $(r_{\bar{x}}, r_{\bar{y}}, r_{\bar{z}}, r_{x})$, then the expression for photometric normal is given by:
\begin{equation}
  \vec{n} = \frac{[r_{x}-r_{\bar{x}}, -2r_{\bar{y}}+(r_{x} + r_{\bar{x}}), -2r_{\bar{z}}+(r_{x} + r_{\bar{x}})]^{T}}{||[r_{x}-r_{\bar{x}}, -2r_{\bar{y}}+(r_{x} + r_{\bar{x}}), -2r_{\bar{z}}+(r_{x} + r_{\bar{x}})]^{T}||}.
  \label{eq:real_time_perf_cap_dutta2010_normal_eq2}
\end{equation}
We can exploit this flexibility in computation of photometric normal to develop a new image capture sequence compatible with the realtime facial geometry capture framework developed by Wilson \etal.

\subsubsection{The New Image Capture Sequence}
\begin{figure}[htbp]
  \centering
  \includegraphics{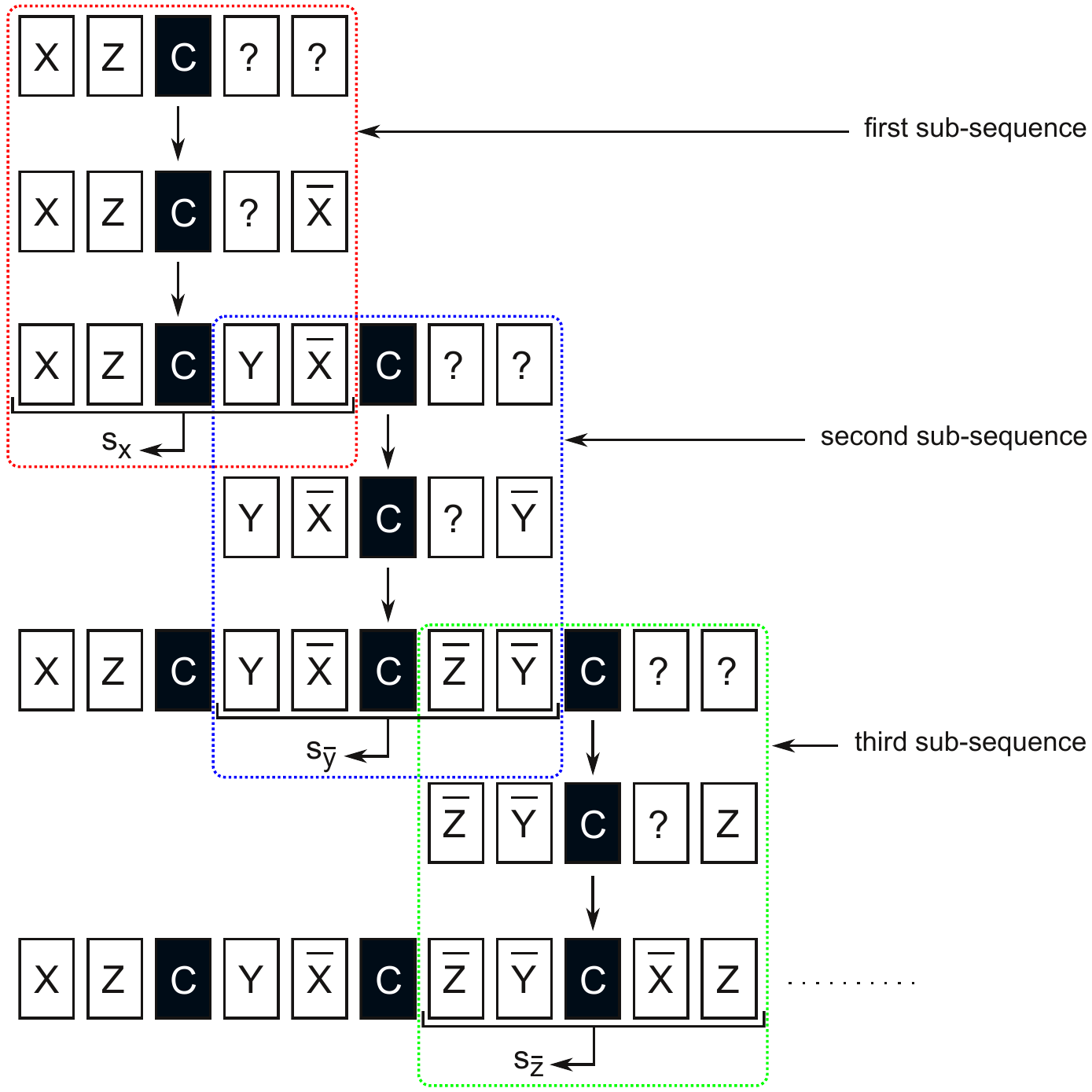}
  \caption{Development of the modified capture sequence}
  \label{fig:real_time_perf_cap_dutta2010_mod_capture_sequence_illus_1}
\end{figure}
Using the flexibility provided by minimal image sets in computation of photometric normal, we develop an image capture sequence containing gradient and complement gradient images interleaved in such a way that either a minimal $\left[r_{x}, r_{y}, r_{z}, \{ r_{\bar{x}}, r_{\bar{y}}, r_{\bar{z}} \}\right]$ or dual minimal image set $\left[r_{\bar{x}}, r_{\bar{y}}, r_{\bar{z}}, \{r_{x}, r_{y}, r_{z}\}\right]$ always flanks the tracking frame. Such a sequence can be created with the help of following rules:
\begin{enumerate}
  \item base complement pair (i.e. $r_{x}$ and $r_{\bar{x}}$ in $[r_{x}, r_{y}, r_{z}, r_{\bar{x}}]$) should be placed farthest from the tracking frame (i.e. the constant illumination image)
  \item exactly two gradient images should lie between any two tracking frames in a sequence
  \item tracking frame should be flanked by two gradient images and these frames should have linear subject motion with respect to the tracking frame
\end{enumerate}
The third rule is based on the assumption that, at high frame rate of capture, three consecutive frames do not have significant subject motion and hence the subject motion can be assumed to be linear in these frames. Even for exaggerated facial motion, this assumption is reasonable given that the gradient image capture frame rate is large (for example 60 fps). Based on this assumption, we align the gradient images adjacent to a tracking frame by half of the flow field from base gradient image pairs in that subsequence.

To illustrate the development of image capture sequence based on minimal image sets, let us consider an example in which we start the sequence with any two arbitrary gradient images and a constant illumination image(tracking frame) $[X \rightarrow Z \rightarrow C \cdots]$ as shown in \figurename\ref{fig:real_time_perf_cap_dutta2010_mod_capture_sequence_illus_1}. The constant illumination image $C$ is preceeded by two gradient images in accordance to Rule 2. According to Rule 1, the base complement pair must be placed farthest i.e. at the two ends of a sub-sequence. As the first position is occupied by $X$, the other base complement pair $\bar{X}$ must appear at the other end as illustrated in the next stage of first sub-sequence shown in \figurename\ref{fig:real_time_perf_cap_dutta2010_mod_capture_sequence_illus_1} (second row from top). We have now partial minimal $4$ image set of $[X,?,Z,\bar{X}]$. It is evident that the unknown image in the set can only be filled by $Y$ gradient. Hence, the final minimal image set corresponding to this sub-sequence is $[X,Y,Z,\bar{X}]$. We name this sub-sequence as $s_{x}$ because $X$ is the base complement pair and gradient images (not the complement gradient images) form the first three members of the minimal set. At this stage, we have the following subsequence: $[X \rightarrow Z \rightarrow C  \rightarrow Y \rightarrow \bar{X} \cdots]$. This subsequence is sufficient to compute photometric normals. Moreover, the warp from $X$ to $C$ and $\bar{X}$ to $C$ can be computed using the Joint Photometric Alignment.

Using the same three rule, we are now ready to further grow the first sub-sequence to second sub-sequence. According to Rule 2, we first place the tracking frame $C$. The other end of this sub-sequence will be filled by $\bar{Y}$ according to Rule 1. The partial minimal image set is $[\bar{X},\bar{Y},?,Y]$. From the dual of our image set, it is evident that the blank space will be occupied by $\bar{Z}$. We name this sub-sequence as $s_{\bar{y}}$ because $Y$ is the base complement pair and complement gradient images form the first three members of the minimal set. Now the capture sequence becomes: $[X \rightarrow Z \rightarrow C  \rightarrow Y \rightarrow \bar{X} \rightarrow C \rightarrow \bar{Z} \rightarrow \bar{Y} \cdots]$.

In a similar way of growing the sequence, we obtain the minimal image set for third sub-sequence as $[?,\bar{Y}, \bar{Z}, ? ,Z]$ and we name this sub-sequence as $s_{\bar{z}}$. At this stage, the combination of three sub-sequences has resulted in the unit sequence $(s_{x},s_{\bar{y}},s_{\bar{z}})$ whose expanded form is given by:
\[
[X \rightarrow Z \rightarrow C  \rightarrow Y \rightarrow \bar{X} \rightarrow C \rightarrow \bar{Z} \rightarrow \bar{Y} \rightarrow C \rightarrow X \rightarrow Z \rightarrow C \cdots]
\]
The end of this capture unit sequence can be combined with the unit sequence generated similarly by $(s_{x},s_{y},s_{\bar{z}})$ which in turn can be combined with $(s_{\bar{x}},s_{y},s_{z})$ and so on. There are total $6$ possible combinations to form sub-sequences: $\left(s_{\{x,\bar{x}\}}, s_{\{y,\bar{y}\}}, s_{\{z,\bar{z}\}}\right)$. Out of these, two unit sequences are not possible: $(s_{\bar{x}},s_{\bar{y}},s_{\bar{z}})$ and $(s_{x},s_{y},s_{z})$. Hence, it is only possible to have the following $4$ unique sub-sequences: $(s_{x},s_{\bar{y}},s_{\bar{z}})$, $(s_{x},s_{y},s_{\bar{z}})$, $(s_{\bar{x}},s_{y},s_{z})$, $(s_{\bar{x}},s_{\bar{y}},s_{z})$. Thus, the final capture sequence for real time performance capture is:
\[
(s_{x},s_{\bar{y}},s_{\bar{z}}) \rightarrow (s_{x},s_{y},s_{\bar{z}}) \rightarrow (s_{\bar{x}},s_{y},s_{z}) \rightarrow (s_{\bar{x}},s_{\bar{y}},s_{z}) \rightarrow (s_{x},s_{\bar{y}},s_{\bar{z}}) \rightarrow \cdots \nonumber
\]
as depicted in \figurename\ref{fig:app_of_light_stage_data:sec:real_time_perf_cap_perf_cap_seq_wilson2010_dutta2010} (bottom).

\subsection{Performance Geometry Capture using the New Image Capture Sequence}
\begin{figure}[htbp]
  \centering
  \includegraphics{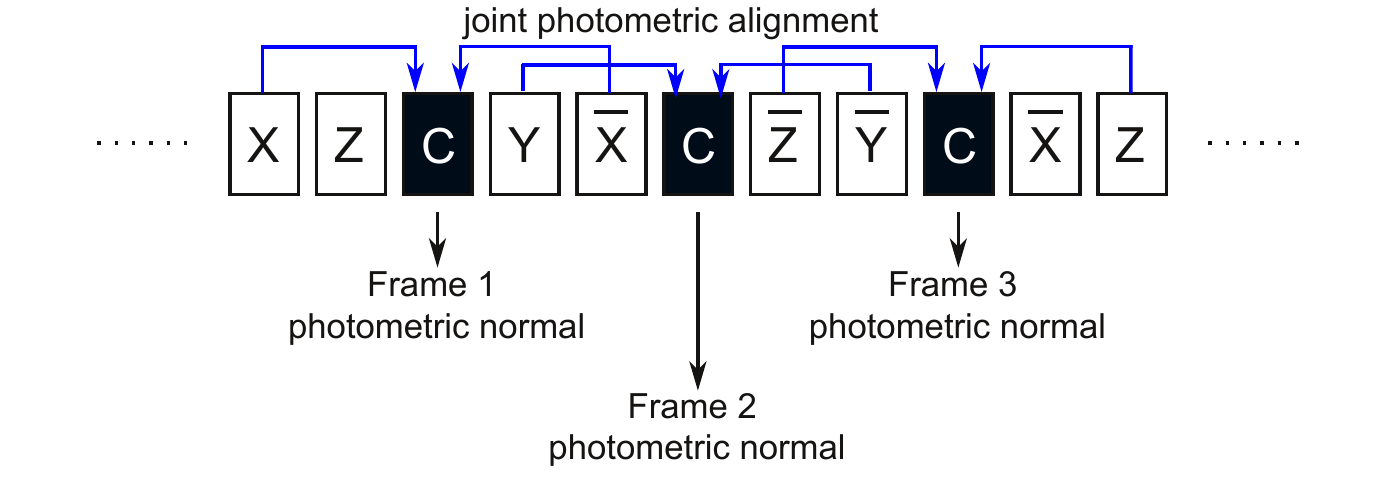}
  \caption{Development of the modified capture sequence}
  \label{fig:real_time_perf_cap_dutta2010_alignment_perf_cap_illustration}
\end{figure}
From the discussion in previous section, we now have a image capture sequence as shown in \figurename\ref{fig:real_time_perf_cap_dutta2010_alignment_perf_cap_illustration}. The images captured in this sequence is sufficient to recover true photometric normals at each tracking frame (frame ${\bf C}$ in \figurename\ref{fig:real_time_perf_cap_dutta2010_alignment_perf_cap_illustration}) using the equations from our minimal image sets analysis discussed in section \ref{ch:mls_data_processing:sec:minimal_image_sets}. 

In \figurename\ref{fig:real_time_perf_cap_dutta2010_alignment_perf_cap_illustration}, the alignment of $X$ and $\bar{X}$ to tracking frame $C$ can be achieved using the Joint Photometric Alignment method of Wilson \etal. However, the photometric normal at tracking frame $C$ cannot be computed from the images of this capture sequence ($[X \rightarrow Z \rightarrow C  \rightarrow Y \rightarrow \bar{X}$) because the two images ($Z$ and $Y$) flanking the tracking frame remain misaligned.

At high frame rate of image capture, three consecutive frames do not have significant non-linear subject motion and hence the subject motion can be assumed to be linear in these frames. Even for exaggerated facial motion, this assumption is reasonable given that the gradient image capture frame rate is large (for example 60 fps). Therefore, for $3$ images captured in a sequence, the flow between $2^{\textnormal{nd}}$ and $3^{\textnormal{rd}}$ frames can be approximated by half of the flow between $1^{\textnormal{st}}$ and $3^{\textnormal{rd}}$ frames. For example: the optical flow field from $Z$ to $C$ in \figurename\ref{fig:real_time_perf_cap_dutta2010_alignment_perf_cap_illustration} can be approximated as half of the flow field from $X$ to $C$ i.e. $f_{Z \rightarrow C}=\frac{f_{X \rightarrow C}}{2}$ and similarly, $f_{Y \rightarrow C}=\frac{f_{\bar{X} \rightarrow C}}{2}$. Hence, the optical flow field for base gradient image pair ($X$ and $\bar{X}$) obtained using Joint Photometric Alignment method of Wilson \etal can be used to approximate the flow field of intermediate gradient frames ($Z$ and $Y$) that flank the tracking frame ($C$).

We will illustrate the procedure of alignment and computation of tracking frame and gradient frame photometric normals using the capture sequence given in \figurename\ref{fig:app_of_light_stage_data:sec:real_time_perf_cap_perf_cap_seq_wilson2010_dutta2010} (bottom). The Joint Photometric Alignment applied to $(\textnormal{I}^{1}, \textnormal{I}^{3}, \textnormal{I}^{5})$ results in two optical flow fields, $f_{\textnormal{I}^{1} \rightarrow \textnormal{I}^{3}}$ and $f_{\textnormal{I}^{5} \rightarrow \textnormal{I}^{3}}$, that align the gradient $X$ and complement gradient $\bar{X}$ images to the flanked tracking frame $C$ respectively. In other words, warping $X$ and $\bar{X}$ according to $f_{\textnormal{I}^{1} \rightarrow \textnormal{I}^{4}}$ and $f_{\textnormal{I}^{5} \rightarrow \textnormal{I}^{4}}$ respectively, aligns both $X$ and $\bar{X}$ to the tracking frame $C$.

\subsubsection{Tracking Frame Photometric Normals}
At each tracking frame, we have access to $1$ pair of aligned gradient and complement gradient images. Assuming that the two gradient images flanking the tracking frame ($Z$ and $Y$ in this sequence) have linear subject motion, photometric normal at the tracking frame $\textrm{I}^{3}$ is given by:

\begin{equation}
 n_{3} = \frac{\left[X_{w} - \bar{X}_{w}, 2Y_{w} - (X_{w}+\bar{X_{w}}), 2Z_{w} - (X_{w}+\bar{X_{w}})\right]}{||\left[X_{w} - \bar{X}_{w}, 2Y_{w} - (X_{w}+\bar{X_{w}}), 2Z_{w} - (X_{w}+\bar{X_{w}})\right]||},
\end{equation}
where,
\begin{eqnarray*}
 X_{w} = \textbf{W} \left(X,f_{\textrm{I}^{1} \rightarrow \textrm{I}^{3}} \right) & , & \bar{X}_{w} = \textbf{W} \left(\bar{X},f_{\textrm{I}^{5} \rightarrow \textrm{I}^{4}} \right),
\end{eqnarray*}
\begin{eqnarray*}
 Y_{w} = \textbf{W} \left(Y,\frac{f_{\textrm{I}^{5} \rightarrow \textrm{I}^{4}}}{2} \right) & , &  Z_{w} = \textbf{W} \left(Z,\frac{f_{\textrm{I}^{1} \rightarrow \textrm{I}^{3}}}{2} \right)
\end{eqnarray*}

Recall that we have used our minimal image set method to compute photometric normal using just $4$ spherical illumination images. Also, note that we have warped $Y$ and $Z$ according to the average flow of the gradient and complement gradient images with respect to the tracking frame.

Similarly, the normal at frame $\textrm{I}^{6}$ is given by:
\begin{equation}
 n_{6} = \frac{\left[ -2\bar{X_{w}} + (Y_{w} + \bar{Y}_{w}), Y_{w} - \bar{Y}_{w}, -2\bar{Z_{w}} + (Y_{w} + \bar{Y}_{w}) \right]}{||\left[ -2\bar{X_{w}} + (Y_{w} + \bar{Y}_{w}), Y_{w} - \bar{Y}_{w}, -2\bar{Z_{w}} + (Y_{w} + \bar{Y}_{w}) \right]||},
\end{equation}
where,
\begin{eqnarray*}
 Y_{w} = \textbf{W} \left(Y,f_{\textrm{I}^{4} \rightarrow \textrm{I}^{6}} \right) & , & \bar{Y}_{w} = \textbf{W} \left(\bar{Y},f_{\textrm{I}^{8} \rightarrow \textrm{I}^{6}} \right).
\end{eqnarray*}
\begin{eqnarray*}
 \bar{X}_{w} = \textbf{W} \left(X,\frac{f_{\textrm{I}^{4} \rightarrow \textrm{I}^{6}}}{2} \right) & , & \bar{Z}_{w} = \textbf{W} \left(Z,\frac{f_{\textrm{I}^{8} \rightarrow \textrm{I}^{6}}}{2} \right).
\end{eqnarray*}
Note the slight change in photometric normal formula caused by dual minimal image set $(\bar{X}, \bar{Y}, \bar{Z}, Y)$.

For frame $\textrm{I}^{9}$, the tracking frame photometric normal is given by:
\begin{equation}
 n_{9} = \frac{\left[ -2\bar{X_{w}} + (Z_{w} + \bar{Z_{w}}), -2\bar{Y_{w}} + (Z_{w} + \bar{Z_{w}}), Z_{w} - \bar{Z_{w}} \right]}{||\left[ -2\bar{X_{w}} + (Z_{w} + \bar{Z_{w}}), -2\bar{Y_{w}} + (Z_{w} + \bar{Z_{w}}), Z - \bar{Z_{w}} \right]||},
\end{equation}
where,
\begin{eqnarray*}
 Z_{w} = \textbf{W} \left(Z,f_{\textrm{I}^{7} \rightarrow \textrm{I}^{9}} \right) & , & \bar{Z}_{w} = \textbf{W} \left(\bar{Y},f_{\textrm{I}^{11} \rightarrow \textrm{I}^{9}} \right).
\end{eqnarray*}
\begin{eqnarray*}
 \bar{X}_{w} = \textbf{W} \left(X,\frac{f_{\textrm{I}^{11} \rightarrow \textrm{I}^{9}}}{2} \right) & , & \bar{Y}_{w} = \textbf{W} \left(Y,\frac{f_{\textrm{I}^{7} \rightarrow \textrm{I}^{9}}}{2} \right).
\end{eqnarray*}
In a similar way, all the other tracking frame photometric normals can also be computed.

\subsubsection{Warped Normals at Intermediate Gradient Frames}
With the flow field from gradient and complement gradient frames to a common tracking frame at hand, we can compute warped normals corresponding to temporal location of gradient and complement gradient images. The warped normal at frames $\textrm{I}^{1}, \textrm{I}^{5}$ is given by:

\[
 n^{'}_{1} = \textbf{W} \left(n_{3},-f_{\textrm{I}^{1} \rightarrow \textrm{I}^{3}} \right) \quad,\quad
 n^{'}_{5} = \textbf{W} \left(n_{3},-f_{\textrm{I}^{5} \rightarrow \textrm{I}^{3}} \right)
\]
Assuming linear subject motion between frame $\textrm{I}^{1}$ and $\textrm{I}^{3}$, the warped normal at frame $\textrm{I}^{2}$ is given by:
\[
 n^{'}_{2} = \textbf{W} \left(n_{3},-\frac{f_{\textrm{I}^{1} \rightarrow \textrm{I}^{3}}}{2} \right)
\]
For each subsequent frames, each gradient frame is flanked by two tracking frame. Therefore, two flow fields exist for each gradient frame and hence, there are two version of warped photometric normal corresponding to each gradient frame. For example, if we consider the gradient image $\bar{X}$ at frame location $\textrm{I}^{5}$, we have the following two warped normals for this frame location:
\[
 n^{''}_{5} = \textbf{W} \left(n_{3},-f_{\textrm{I}^{5} \rightarrow \textrm{I}^{3}} \right) \quad,\quad
 n^{'''}_{5} = \textbf{W} \left(n_{6},-\frac{f_{\textrm{I}^{4} \rightarrow \textrm{I}^{6}}}{2} \right)
\]
Note that, computation of the warped normal $n^{'''}_{5}$ is based on linear subject motion assumption. Based on the temporal distace of $\textrm{I}^{5}$ with respect to the two flanking tracking frames $\textrm{I}^{3}$ and $\textrm{I}^{6}$, the warped normal at frame $\textrm{I}^{5}$ is given by:
\[
 n^{'}_{5} = \frac{2n^{''}_{5}+n^{'''}_{5}}{||2n^{''}_{5}+n^{'''}_{5}||}
\]

In a similar way, we can compute warped normals at all the remaining intermediate gradient frames.

\subsection{Results}
\begin{sidewaysfigure}
  \centering
  \includegraphics{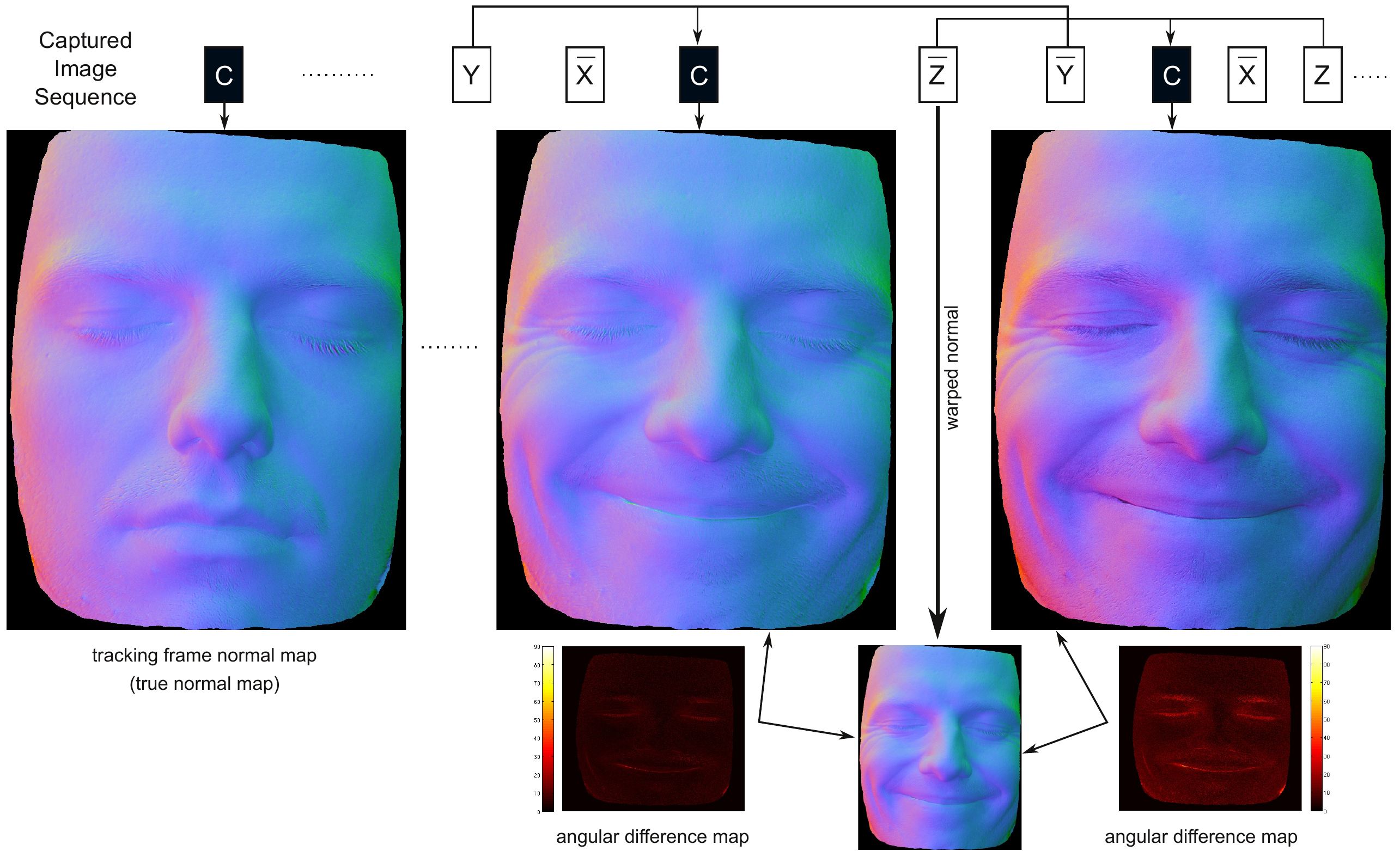}
  \caption{Photometric normals obtained from proposed image capture sequence based on minimal image sets}	\label{fig:app_of_light_stage_data:sec:real_time_perf_cap_app_of_light_stage_data_dutta2010_perf_geom_results}
\end{sidewaysfigure}
Before discussing the results, we describe our capture device setup and its limitations. We used a monochrome JAICM200GE GigE camera. We could not capture gradient images at $60$ fps (capture rate of Wilson \etal) for the following two reasons:
\begin{itemize}
 \item The computer that received the captured image packets via ethernet only supported a maximum ethernet packet size of $1428$ bytes. Although, our camera is capable of using jumbo ethernet packet ($9000$ bytes) for a very high frame rate capture, we could not use this feature due to the limitation of our receiving network node.
 \item Our Light Stage uses only 41 LED. Hence, based on the sensitivity of our camera, we observed that a minimum exposure time of $50$ ms is required to capture well exposed face images.
\end{itemize}
We observed that when quick exaggerated facial motion is performed, there occurs drastic change in the $1^{\textrm{st}}$ and $5^{\textrm{th}}$ frames of a subsequence block i.e. $(X,Y,Z,C,\bar{X})$. Hence, to address the capture rate limitation of our device, we asked our subject to change facial expression slowly while we captured gradient frames at the rate of 20 fps. At higher frame rate, we believe that our proposed sequence can resolve the facial performance geometry more finely.

\figurename\ref{fig:app_of_light_stage_data:sec:real_time_perf_cap_app_of_light_stage_data_dutta2010_perf_geom_results} shows the photometric normals computed using images from the modified image capture sequence based on minimal image sets. The tracking frame photometric normals accurately captures the facial geometry during facial motion. Warped photometric normal is computed as weighted average of the tracking frame normals. The weighting of tracking frame normal is performed according to the temporal distance of warped normal from these tracking frames. This weighting strategy is evident from the angular difference map shown in \figurename\ref{fig:app_of_light_stage_data:sec:real_time_perf_cap_app_of_light_stage_data_dutta2010_perf_geom_results} (bottom). These angular difference maps also depict very small motion in the lips and eyes region.

\subsection{Discussion}
\begin{figure}[htbp]
  \centering
  \includegraphics{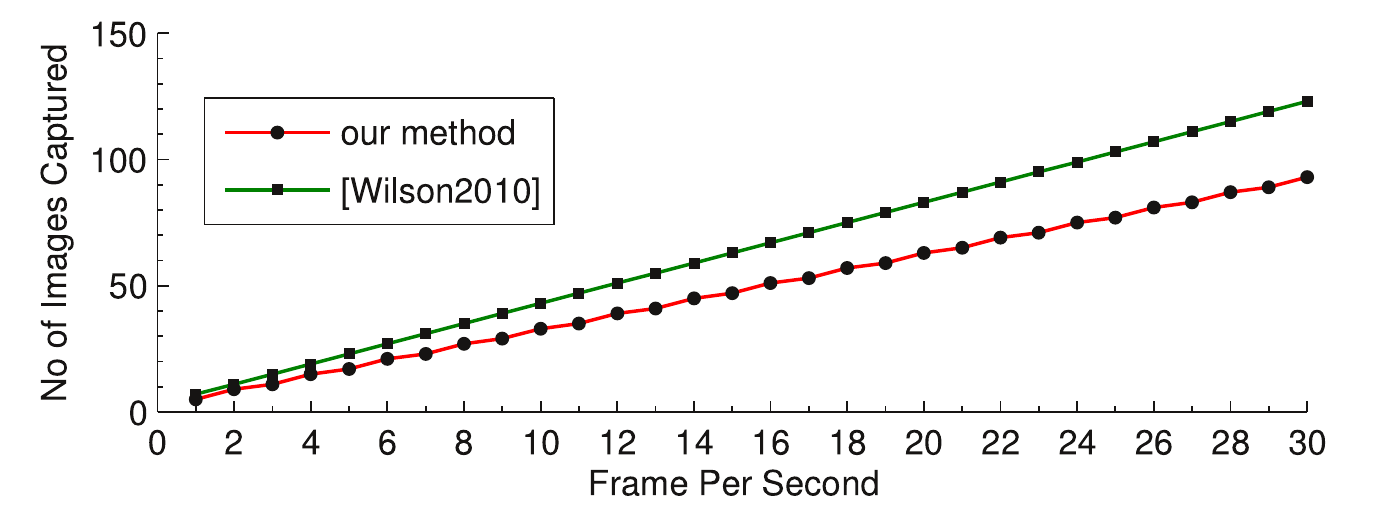}
  \caption{Image capture requirement analysis for performance capture using Wilson \etal \cite{wilson2010temporal} and our 4 image method}
  \label{fig:real_time_perf_cap_wilson2010_dutta2010_data_cap_req_analysis}
\end{figure}

We have shown that minimal image sets can be exploited to form a capture sequence that can not only reduce the data capture requirement of a realtime performance geometry capture but also reduces the computational cost involved in alignment of the captured images. \tablename\ref{tbl:real_time_perf_cap_wilson2010_dutta2010_data_cap_requirement} shows the relationship between tracking frame capture rate and required number of images to be captured for Wilson \etal and our 4 image method. The impact of reduction in image capture requirement for real time performance capture is pronounced for higher frame rate as shown in  \figurename\ref{fig:real_time_perf_cap_wilson2010_dutta2010_data_cap_req_analysis}.
\begin{table}
  \centering
  \caption{Image capture requirements for performance capture using Wilson \etal and our 4 image method}
  \begin{tabular}{ |l |c |c| }
  \hline
    & \multicolumn{2}{|c|}{Total Number of Images Captured} \\
    \hline
    Tracking Frame Count ($n$) & Wilson \etal \cite{wilson2010temporal} & our method \\ \hline
    $1$ & $7$ & $5$ \\
    $2$ & $11$ & $9$ \\
    $3$ & $15$ & $11$ \\
    $4$ & $19$ & $15$ \\
    $5$ & $23$ & $17$ \\
    $6$ & $27$ & $21$ \\
    $\cdots$ & $\cdots$ & $\cdots$ \\
    \multirow{2}{*}{$n$} & \multirow{2}{*}{$4n+3$} & \multirow{2}{*}{
      $\left\{
      \begin{array}{rl}
        6\left(\lfloor \frac{n}{2}\rfloor +1 \right)-1 & \textrm{odd} \; n \\
        3n + 3 & \textrm{even} \; n
      \end{array} \right. $
    } \\
    & & \\ \hline
  \end{tabular}
  \label{tbl:real_time_perf_cap_wilson2010_dutta2010_data_cap_requirement}
\end{table}

Using our proposed real time performance capture sequence, we can compute true photometric normal map after every two image capture. This allows us to densely sample the complete dynamic performance even at lower frame rate.

\subsubsection{Limitations}
The proposed capture sequence and method of photometric normal computation assumes high capture frame rate $(\sim 60 \textrm{fps})$ for gradient images. For lower frame rates $(\sim 20 \textrm{fps})$, the geometry of quick exaggerated facial motion cannot be correctly recovered because the alignment algorithm cannot handle non-linear changes in facial geometry. Moreover, the assumption that three consecutive frames do not have significant subject motion and that the subject motion can be assumed to be linear in these frames become invalid for such quick exaggerated facial motion.

\section{Stimuli Image Dataset for Psychology Experiment}
\label{ch:app_of_light_stage_data:sec:psych_exp_dataset}
The overall appearance of a human face is due to its 3D shape and 2D skin reflectance (skin texture) property. Hence, these two parameters are believed to play a critical role in face processing and recognition carried out by the human brain. Knowledge of how these two sources of information are represented and processed in the neural level is the key to understanding the face recognition mechanism of the human brain.

The face adaptation paradigm is commonly used to study the representation and processing of these two information i.e. 3D shape and 2D skin reflectance information. Face adaptation refers to the decay in neuronal response of face processing regions in human brain when a human observer is exposed repeatedly to same stimulus (e.g. face image). Original neuronal response can be recovered by altering some properties of the stimulus. Face adaptation paradigm is based on the assumption that the changes in stimuli that causes recovery of the neuronal response relate to the functional properties of cortical neurons \cite{grillspector2001fmr}.

Application of adaptation paradigm requires the ability to control specific properties of the stimuli. Caharel \etal \cite{caharel2009recognizing} used 3D morphable model to control the 3D shape and 2D reflectance information of stimuli images. They examined the time course (i.e. temporal sequence) for the processing of 3D shape and 2D skin reflectance information using the Event Related Brain Potential - ERP\footnote{Electroencephalography (EEG) recording during an epoch (time slot in which stimulus is shown) constitute ERP} adaptation paradigm. They discovered that 3D shape information caused early sensitivity $(\sim 160-250\textrm{ms})$ to human faces. Furthermore, they also found that both 3D shape and 2D skin reflectance information (skin texture) contributed equally to ERP on the later time window $(\sim 250-350\textrm{ms})$.

We collaborated with Jones \etal \cite{jones2010how} to study the neural representation of face's 3D shape and 2D skin reflectance information in face selective regions of the human brain. Using fMR adaptation paradigm \cite{grillspector2001fmr}, Jones \etal analyzed the adaptation of face selective regions in the Fusiform Face Area (FFA), Occipital Face Area (OFA) and Superior Temporal Sulcus (STS). Participants were shown stimuli face images that contained :
\begin{enumerate}
 \item 3D shape information (shape only)
 \item 2D skin reflectance information (texture only)
 \item both shape and texture information
\end{enumerate}

\subsection{Stimuli Image Dataset}
\begin{figure}[htbp]
  \centering
  \includegraphics{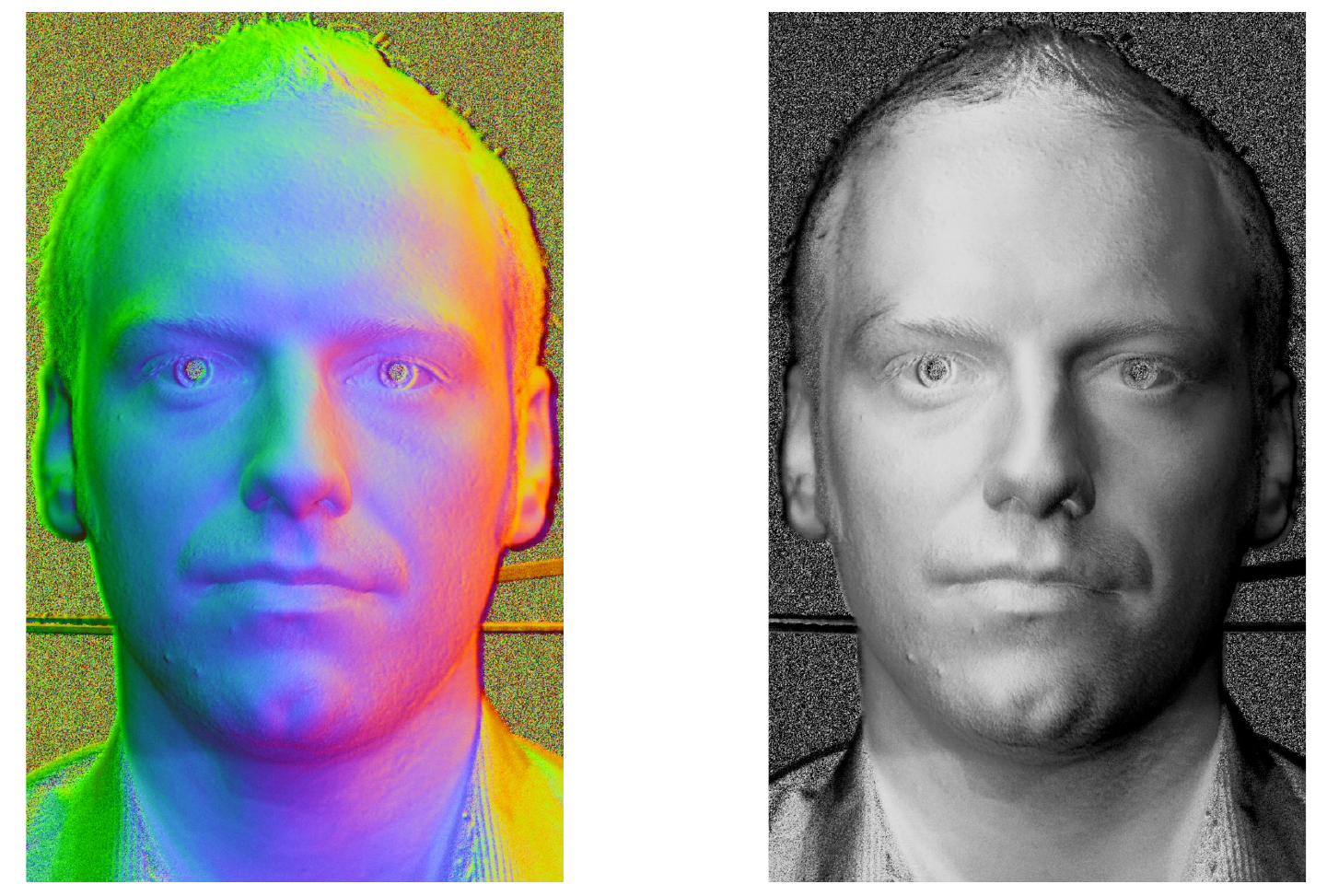}
  \caption{(left) normal map obtained using spherical gradient photometric stereo, and corresponding (right) \textit{shape only} stimuli image}
  \label{fig:app_of_light_stage_data_shape_only_img}
\end{figure}
Using our Light Stage, we created the stimuli images dataset required for this study aiming to investigate the neural representation of 3D shape and 2D skin reflectance in the visual cortex. The experiment was conducted by Jones \etal and a detailed account of the experimental procedure and discussion of the results is available in \cite{jones2010how}. Here, we discuss the method that was used to create stimuli image dataset consisting of (a) texture only (b) shape only and (c) texture and shape images of human faces. The frontal view spherical gradient $(X,Y,Z)$ and constant illumination $C$ images of participants in neutral expression was captured and the corresponding \textit{shape-only} and \textit{texture-only} images were generated as follows:

\subsubsection{Computing the 3D shape only image}
We can acquire highly detailed (down to the level of skin pore detail) photometric normal map (left - \figurename\ref{fig:app_of_light_stage_data_shape_only_img}) of a human face using the spherical gradient photometric stereo technique discussed in \chaptername~\ref{ch:mls_data_processing}. This normal map recovers facial geometry in the form of surface normal vector at each surface point covered by individual pixels of an imaging device. Using this normal map, we can generate a front lit Lambertian rendering as follows:
\[
 \textrm{I}_{\textrm{shape-only}} = n.l_{1} + n.l_{2}
\]
where, $l_{1}$ and $l_{2}$ are the two front lighting direction vectors (chosen manually to create realistic shape only image) and $n$ is the facial normal map computed using the spherical gradient photometric stereo technique. The resulting \textit{shape-only} rendered image is shown in \figurename\ref{fig:app_of_light_stage_data_shape_only_img} (right).

\subsubsection{Computing the texture only image}
Skin texture is the result of light reflected after subsurface scattering. In other words, the portion of incident light reflected after entering the skin surface constitutes the characteristic skin colour. Using cross polarization, we separate the facial reflectance into the diffuse and specular components as described in section \ref{ch:design_and_calib_of_mls:sec:diff_spec_separation}. The diffuse only image captured under constant full spherical illumination records the reflectance component responsible for skin texture as shown in \figurename\ref{fig:app_of_light_stage_data_shape_and_texture_img} (left). We use this image as the \textit{texture-only} stimuli.

\subsubsection{Combined shape and texture image}
\begin{figure}[htbp]
  \centering
  \includegraphics{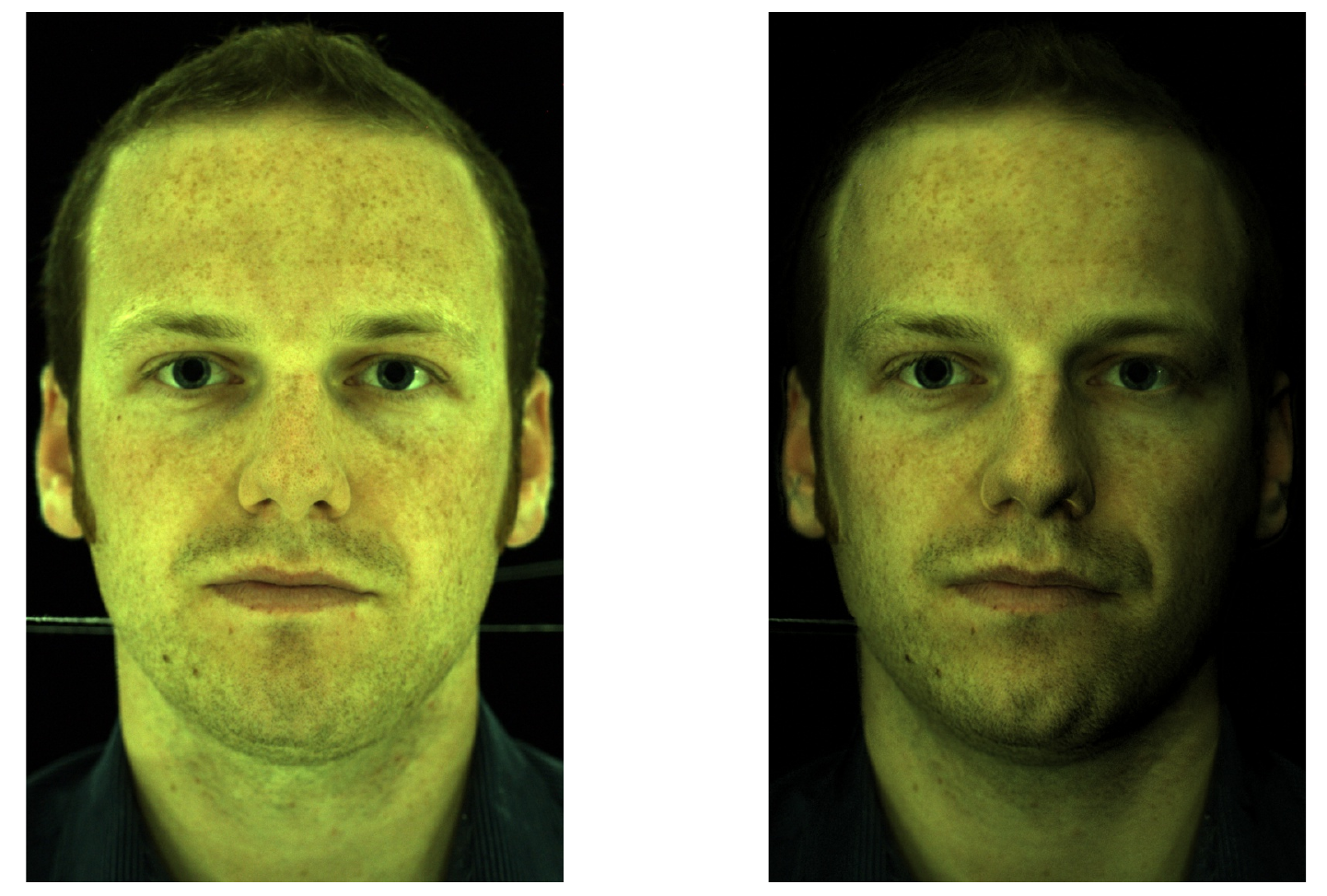}
  \caption{(left) \textit{texture-only} stimuli image, and (right) combined \textit{shape and texture} stimuli image}
  \label{fig:app_of_light_stage_data_shape_and_texture_img}
\end{figure}

Combining the \textit{shape-only} image with the \textit{texture-only} image (i.e. diffuse albedo) results in a combined shape and texture image as shown in \figurename\ref{fig:app_of_light_stage_data_shape_and_texture_img} (right). This stimuli image represents the facial images as captured by a real world camera.

\subsection{Results and Discussion}
A complete explanation of experimental procedure and results are available in Jones \etal \cite{jones2010how}. Here, we present a brief summary of results discussed in \cite{jones2010how}. Both FFA and OFA regions exibited significant adaptation to all the image types in the stimuli image dataset. Moreover, there was no significant difference in the activation of the hemisperes. Based on the adaptation paradigm, Jones \etal concluded that the 3D shape and 2D skin reflectance information are represented equally in the face selective regions of the brain. Furthermore, they also found that there was no significant effect of familiarity in the activation of FFA region. This indicated that the FFA region is largely involved in general face processing task rather than in dealing with facial identity.

A Light Stage with $41$ LEDs was used to generate the stimuli images for this experiment. It required capture of just $4$ images (capture time $\sim1$ sec) and almost no post processing to compute the \textit{shape-only} and \textit{texture-only} stimuli images. Electronic control of each LED brightness and data capture in a dark room ensured that the level of illumination remained consistent across different face images.

The spherical gradient photometric stereo technique of Ma \etal \cite{ma2007rapid} was used to compute the facial normal map which in turn allowed rendering of \textit{shape-only} images. The quality of photometric normals computed using \cite{ma2007rapid} is known to degrade with light discretization i.e. coarse approximation of spherical illumination (see \ref{ch:mls_data_processing:sec:quad_prog_for_norm_improvement} for details). Our light stage used only $41$ LED: $74\%$ less light sources as compared to $151$ LED used by \cite{ma2007rapid}. At the time of the stimuli dataset creation, we had not discovered our minimal image sets method discussed in section \ref{ch:mls_data_processing:sec:minimal_image_sets}. We were also not aware of the normal map computation technique proposed by Wilson \etal \cite{wilson2010temporal} which required capture of $6$ gradient images $(X,Y,Z,\bar{X},\bar{Y},\bar{Z})$. Hence, increasing the number of light source in our Light Stage was the only possible but expensive route to improve the quality of normal map computed using \cite{ma2007rapid}. However, with the minimal image sets method (see section \ref{ch:mls_data_processing:sec:minimal_image_sets}) in hand, we can now use the same 41 LED light stage to compute very accurate photometric normals without incuring the cost of capturing extra images as required by \cite{wilson2010temporal}.

Concave regions of a human face (like corner of the eyes) do not receive full hemispherical illumination. In other words, non-convex regions of a face are affected by ambient occlusion. Although, ambient occlusion helps add realism in 3D Computer Graphics, this effect is not desirable for \textit{texture only} stimulus image because it adds shading information to the non-convex regions. Hence, the \textit{texture only} stimulus images of a human face have some shading effect in the non-convex facial regions.

\chapter{Conclusion}
\label{ch:conclusion}
In this thesis, we have presented a detailed analysis of design and calibration (geometric and radiometric) of a novel shape and reflectance acquisition device called the Multispectral Light Stage. Using the spherical gradient photometric stereo method, we capture highly detailed facial geometry (down to the level of skin pores detail). We used a beam splitter based capture device setup to simultaneously capture both the parallel and cross polarised reflectance components. Therefore, the image alignment procedure is not required to compute specular and diffuse images from the captured parallel and cross polarised images. To record Multispectral skin reflectance map, we added a set of narrow bandpass optical filters to our image capture device. These reflectance maps can be used to estimate biophysical skin parameters such as the distribution of pigmentation and blood beneath the surface of the skin.

We have extended the analysis of original spherical gradient photometric stereo method to consider the effect of diffuse lobes distortion on the quality of recovered surface geometry. Using our modified radiance equations, we show that the symmetric deformation in diffuse reflectance lobe under gradient and complement gradient illumination cancel when computing surface normal using Wilson \etal \cite{wilson2010temporal} $6$ image method. In addition, we also show that the method of Ma \etal \cite{ma2008framework}, which requires $4$ images, is highly affected by deformed diffuse lobes. We propose a minimal image set method, requiring just $4$ images, that combines the advantage of the original method of Ma \etal (reduced data capture requirement) with that of Wilson \etal (improved robustness). We show that our method maintains the quality of Wilson \etal while requiring fewer gradient images. Using our modified radiance equations, we also explore a Quadratic Programming (QP) based normal correction algorithm for surface normals recovered using spherical gradient photometric stereo.

Based on our minimal image sets method, we have proposed a modification to the original performance geometry capture sequence of Wilson \etal \cite{wilson2010temporal}. Minimal image sets method provides the flexibility of computing accurate photometric normals from all the possible combinations in minimal image set $(X,Y,Z,\{\bar{X},\bar{Y},\bar{Z}\})$ or the dual minimal image set $(\bar{X},\bar{Y},\bar{Z},\{X,Y,Z\})$. We exploit this flexibility to create a performance capture sequence which contain gradient and complement gradient images interleaved in such a way that it always becomes possible to compute aligned photometric normals at the tracking frame (i.e. constant illumination image). This new capture sequence not only reduces the data capture requirement but also reduces the postprocessing computation cost of existing photometric stereo based performance geometry capture methods like \cite{wilson2010temporal}.

We have also explored the use of Light Stage data for creating stimulus image dataset for a psychology experiment investigating the neural representation of 3D shape and 2D skin reflectance (texture) of a human face. For a given face, we generate three stimulus images: the first contains only the 3D shape information, the second contains only 2D skin reflectance (texture) information and the third contains both shape and texture information. This image dataset has been used by Jones \etal \cite{jones2010how} for studying the neural representation of 3D shape and texture of a human face. The high quality photometric normal map obtained from spherical gradient images is used to create a front lit Lambertian rendering of that face. This \textit{shape only} rendered image contains only the 3D shape information. The constant spherical illumination image represents the \textit{texture only} because no shading cues are present due to spherical illumination.

\section{Future Work}
The present design of Multispectral Light Stage discussed in \chaptername~\ref{ch:design_and_calib_of_mls} can be improved in many ways. First, finer approximation of spherical gradient illumination can be achieved by increasing the number of light sources to $162$. Present version of our Light Stage consists of only $41$ LEDs attached to the vertices of a twice subdivided icosahedron. The light reaching the camera sensor is attenuated by the light source polarizer($<50\%$ transmission), optical filter($<90\%$ transmission) and the polarizing beam splitter($<50\%$ transmission). Hence the camera sensor receives only $\sim22\%$ of the total emitted light even if we image a perfect reflector. As a result, capture of multispectral reflectance map requires longer exposure which increases the overall capture time. Adding LEDs to the edges of the twice subdivided icosahedron will not only result in finer approximation of spherical illumination but also ensure that more light is reflected off the object present at the center of the Light Stage.

The second improvement in the design of Multispectral Light Stage can be accomplished by using a stepper motor driven filter wheel. This would help reduce the capture time of Multispectral skin reflectance map. Adding electronic control to the filter wheel would also help automate the whole capture process. For geometric calibration, using a sphere instead of planar checker board would result in more accurate model of image formation.

\chaptername~\ref{ch:mls_data_processing} describes the minimal image sets method that not only reduces the data capture requirement of spherical gradient photometric stereo but also improves the quality of recovered surface geometry when diffuse lobes are distorted. Future work in this area can explore such correction mechanism for specular reflectance lobes as well. Moreover, we also investigated a Quadratic Programming (QP) based approach for correction of deformed diffuse lobe. However, as the resulting system was sevearly underconstrained ($6$ equations and $9$ unknowns), our optimization based correction strategy did not result in significant improvement. Future research can also explore improved modeling of specular and diffuse reflectance lobe deformation and search for more constraints.

We used the Joint Photometric Alignment method proposed by Wilson \etal \cite{wilson2010temporal} to align gradient and complement gradient images to a common tracking frame. However, this alignment technique is not applicable to multispectral reflectance maps. Hence, future research in this area could investigate into alignment methods for multispectral reflectance maps. One interesting observation regarding this future work is that as the specular reflectance is a surface phenomena, the value of specular radiance should remain constant throughout all the multispectral images. This relationship between multispectral specular reflectance maps can be used to align the specular images and in turn, also the diffuse images.

\chaptername~\ref{ch:app_of_light_stage_data} also invites future work. The alignment method used in the realtime performance geometry capture cannot handle changes in facial geometry. Moreover, it is based on the assumption that the subject motion is linear. Future work in this area can investigate to overcome these limitations of the photometric alignment method. Additionally, capture of performance geometry does not allow tracking of facial feature points. This prevents transfer of facial performance geometry to 3D face models (obtained from range scannner) because no correspondance between image point and its corresponding 3D vertex can be established. Furthermore, the application of Light Stage in creating stimulus image dataset for different types of psychology experiment can also be explored.

Further application of Light Stage data can also explore multi view photometric stereo. This involves capturing facial geometry and reflectance map using two or more camera capturing different view of a face. This would allow reconstruction of high quality 3D geometry based on photometric normals from multiple views.

\bibliographystyle{plain}
\bibliography{dutta2010_msc_thesis}
\nocite{*}

\end{document}